\documentclass[11pt]{article}

\usepackage[preprint]{acl}

\usepackage{times}
\usepackage{latexsym}
\usepackage[T1]{fontenc}
\usepackage[utf8]{inputenc}
\usepackage{microtype}
\usepackage{inconsolata}
\usepackage{graphicx}

\usepackage{amsmath}
\usepackage{amsfonts}
\usepackage{amssymb}
\usepackage{booktabs}
\usepackage{nicefrac}
\usepackage{xcolor}
\usepackage{hyperref}

\usepackage{enumitem}
\usepackage{placeins}
\usepackage{float}
\usepackage{subcaption}
\usepackage{afterpage}
\usepackage[textsize=tiny]{todonotes}

\title{Bayesian Control for Coding Agents}

\author{Theodore Papamarkou\textsuperscript{1,2 $\diamondsuit$} \,
Vladislav Smirnov\textsuperscript{3 $\diamondsuit$} \,
Viktor Mazanov\textsuperscript{3 $\diamondsuit$} \,
Artem Vazhentsev\textsuperscript{3}\\
\textbf{Preslav Nakov}\textsuperscript{3} \,
\textbf{Timothy Baldwin}\textsuperscript{3} \,
\textbf{Artem Shelmanov}\textsuperscript{3}\\
\textsuperscript{1}PolyShape \,
\textsuperscript{2}National Technical University of Athens \,
\textsuperscript{3}MBZUAI \\
$^{\diamondsuit}$Equal contribution}

\begin{document}
\maketitle

\begin{abstract}
Modern coding agents pair LLM generators with various tools, including cheap diagnostics and expensive verifiers. The tool-use decisions are typically governed by orchestrators that often use fixed rules and ignore uncertainty. We formulate orchestration as cost-sensitive sequential hypothesis testing: a Bayesian controller maintains a belief over candidate correctness and dynamically decides whether to gather more evidence, refine the candidate, verify it, or stop. Across six generators and nine coding benchmarks, Bayesian control proves to be most valuable when verification is costly and critics are informative but imperfect. Beyond control, the belief state yields an interpretable correctness score that outperforms token-probability and raw tool-success baselines for uncertainty quantification.
\end{abstract}

\section{Introduction}
\label{sec:intro}

Large Language Models (LLMs) are increasingly capable of generating sophisticated code, but their practical impact is most fully realized through coding agents that orchestrate generation, diagnosis, refinement, and verification. A modern coding-agent harness typically wraps an LLM-based generator with auxiliary tools, such as syntax checkers, public tests, LLM-based reviewers, refinement prompts, and a high-fidelity verifier, also known as an oracle, that approximates or establishes ground truth~\citep{yang2024sweagent,wang2025openhands}. Because these tools vary in cost and reliability, the central engineering problem is to produce correct code while efficiently allocating limited resources by strategically deciding when to refine, discard, or verify a candidate solution.

\begin{figure}
    \centering
    \includegraphics[width=1\linewidth,trim=0mm 0mm 0mm 0mm,clip]{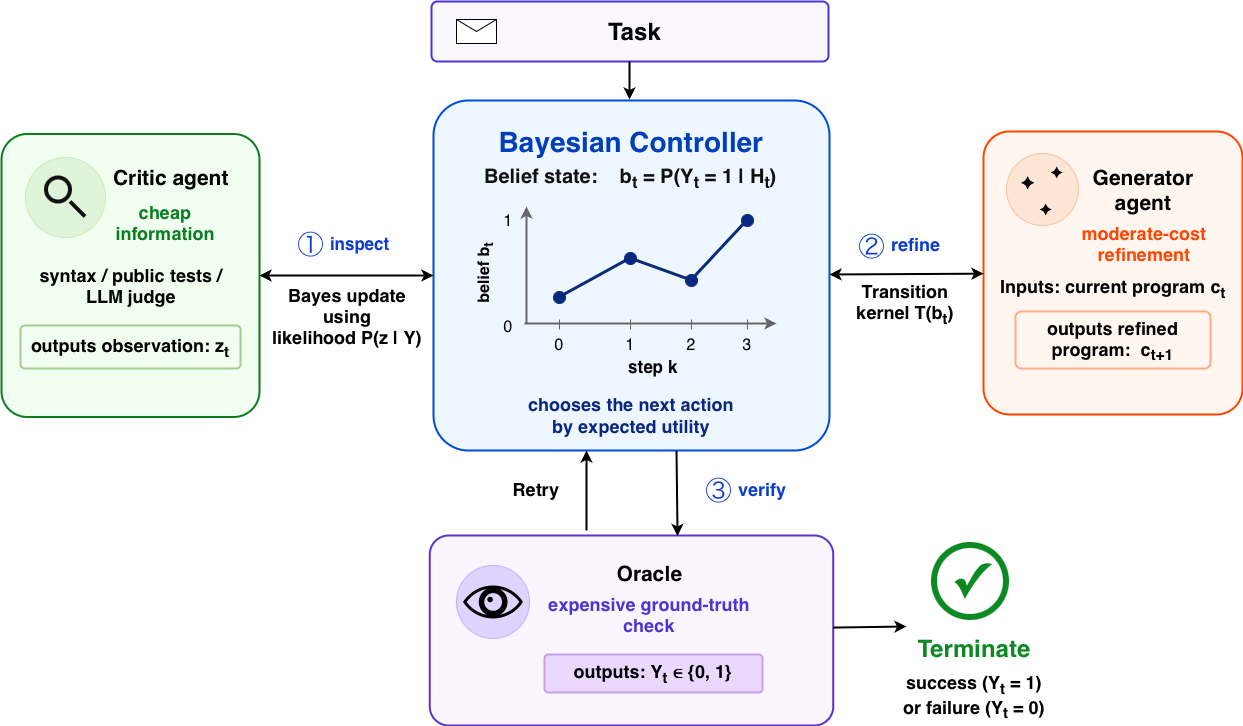}
    \caption{Illustration of the Bayesian control loop.}
    \label{fig:bayesian_agent_diagram}
\end{figure}

Many existing orchestration policies address this problem using fixed rules, such as always verifying, best-of-$N$ sampling, single-critic gates, or predefined generate–critique–regenerate–verify loops~\citep{inala2022coderanker,li2022alphacode,chen2023codet,huang2023agentcoder}. Although such policies can be competitive, they do not maintain a posterior over candidate correctness or explicitly weigh the value of critic calls against their cost. As a result, they are not designed to adapt stopping decisions to prior task difficulty, critic reliability, generator repair probability, or verifier cost.

In this work, following~\citet{papamarkou2026position}, we study code-generation control as a Bayesian decision problem. The latent state is candidate correctness, $Y\in\{0,1\}$. The controller maintains a belief $b=P(Y=1\mid\text{evidence})$ over whether the current candidate will pass the verifier. Critic calls are treated as noisy observations, generator calls as stochastic transitions that may fix or break a candidate, and verifier calls as costly terminal actions. The objective is expected utility, that is, reward for a correct solution minus the accumulated costs of generation, criticism, refinement, and verification. This yields a sequential hypothesis-testing problem that can be encoded as a Partially Observable Markov Decision Process (POMDP)~\citep{kaelbling1998pomdp}, whose Bellman equation expresses the value of information. 
Figure~\ref{fig:bayesian_agent_diagram} illustrates our Bayesian control loop. 

We implement this idea with two controllers derived from the Bellman equation. The first is a one-step Bayesian greedy controller that uses critic likelihoods to decide whether to call another critic, verify, regenerate, or stop. The second is a finite-horizon Bayesian dynamic-programming (DP) controller that performs backward induction and uses a measured generator transition kernel, $\hat P(\text{fix}\mid\text{broken})$ and $\hat P(\text{break}\mid\text{correct})$, estimated from iterative refinement trajectories. Both controllers operate on top of frozen LLMs: the base generators are not trained, and the optimization problem is entirely at the control layer.

We ask a series of empirical questions: when does a Bayesian belief-state controller improve cost-adjusted code generation; when does a simple heuristic suffice; and when is verification so cheap or the prior probability of success so high that no orchestration is needed?

We evaluate six generators across nine coding benchmarks 
and compare policies in terms of expected utility under varying cost–reward trade-offs. The results show that Bayesian control is most useful in low-prior regimes with informative but imperfect critics, while simple public test gating or always-verifying policies are preferable when public tests are highly predictive of hidden-test success, candidate patches are already likely to be correct, or verification is cheap. Dynamic programming adds value only when measured repair transitions make multi-step refinement worthwhile. Our empirical study supports these conclusions through out-of-sample validation, comparisons with stronger refinement baselines, and negative results that identify when naive Bayesian variants fail.

\textbf{The contributions} of this paper are as follows. (1) We formulate code-generation control as a cost-sensitive sequential hypothesis-testing problem over candidate correctness. (2) We derive two Bayesian belief-state controllers from the Bellman equation, namely a one-step greedy controller and a finite-horizon dynamic-programming controller. (3)  Experiments across diverse generators and benchmarks reveal three regimes governing which orchestration strategy dominates and clarify when DP planning beats the greedy policy. (4) We further show that the posterior beliefs induced by our hypothesis-testing framework serve as effective uncertainty scores for the coding agent that could be used to detect low-quality agent outputs.

\section{Related Work}
\label{sec:related_work}

\textbf{Bayesian decision theory.}
Our approach uses sequential analysis and the value of information. \citet{wald1947sequential} studied sequential tests that trade error against sampling cost, while \citet{howard1966information} formalized the economic value of evidence. POMDPs generalize this logic to control under latent state uncertainty \citep{kaelbling1998pomdp}. The recent work of  \citet{papamarkou2026position} argues that the control layer of agentic AI should maintain beliefs over task-relevant latent quantities and act by utility. Our paper implements this idea for code generation. The latent variable is candidate correctness, the observations are syntax checks, public tests, and LLM judges, the transitions are refinement steps, and the terminal action is an expensive verifier.

\textbf{POMDP methods for LLM agents.}
Recent work has begun to use Bayesian information acquisition for LLM agents. \citet{choudhury2026bedllm} use sequential Bayesian experimental design to choose questions that maximize expected information gain in multi-turn information gathering. \citet{suri2025structured} model tool-argument clarification as structured uncertainty and use a POMDP to choose clarifying questions. These papers are close in spirit because they treat interaction as costly information acquisition. They differ from our setting in the latent state, action space, and empirical target. We study code-generation control, where the agent must decide whether to buy more diagnostic evidence, refine a candidate, or pay for verification, and we evaluate utility across coding benchmarks rather than clarification success or information-gathering games.

\textbf{Coding agents.}
A large line of work is devoted to coding agents. Generated tests and reranking have been used to select better candidates \citep{chen2023codet,inala2022coderanker,zhang2023coderreviewer}. Iterative self-feedback and verbal memory methods refine model outputs over multiple turns \citep{madaan2023self,shinn2023reflexion}. Multi-agent and tree-search systems such as AgentCoder and LATS introduce richer generate-test-refine or planning loops \citep{huang2023agentcoder,zhou2024lats}. Repository-level agents such as SWE-agent and OpenHands focus on interfaces, sandboxed execution, and autonomous tool use for realistic software engineering tasks \citep{yang2024sweagent,wang2025openhands}. Our empirical setting follows established code-generation and program-repair benchmarks, including LiveCodeBench, SWE-Bench, EvalPlus, HumanEvalFix, and CodeContests \citep{jain2024livecodebench,jimenez2024swebench,liu2023mbpp+,muennighoff2024octopack,li2022alphacode}. 

Together, these methods and benchmarks demonstrate the value of external feedback, iterative control, and agentic stopping policies, and they provide the experimental basis for our evaluation. The distinguishing feature of our approach is that it quantifies the agent's uncertainty as a posterior belief over candidate correctness. This posterior provides both a decision state for cost-aware control and an uncertainty score. As a result, our framework is complementary to existing agentic harnesses: their tool-use and stopping loops can be augmented with posterior beliefs rather than replaced. Moreover, our characterization of cost-and-reliability regimes identifies when such beliefs improve over fixed workflows and when simpler policies are sufficient.

\section{Task and Environment}
\label{sec:preliminaries}

We consider the problem of automated program synthesis under resource constraints. This section formalizes the task environment, the abstraction of computational agents, and the decision-theoretic objective that drives control.

Let $\mathcal{X}$ denote the space of natural language specifications and $\mathcal{C}$ the space of executable programs. A task instance consists of a specification $x \in \mathcal{X}$ and an oracle verification function $\mathcal{O} : \mathcal{C} \times \mathcal{X} \to \{0, 1\}$, which returns $1$ if and only if program $c$ satisfies the requirements of $x$. Access to $\mathcal{O}$ is mediated by a \emph{verifier agent} $a_{\mathrm{ver}}$, hereafter called the \emph{oracle}, whose invocation incurs a significant computational or monetary cost $C_{\mathrm{ver}}$. For any candidate $c_t$, let $Y_t = \mathcal{O}(c_t, x) \in \{0, 1\}$ be its latent correctness, indicating whether the candidate passes oracle verification ($Y_t = 1$) or not ($Y_t = 0$).

To actively improve the candidate program, we can invoke an LLM $G_\theta$ with fixed parameters $\theta$. We denote this \emph{generator agent} by $a_{\mathrm{gen}}$. At iteration $t$, given a specification $x$ and a current candidate $c_t$, the generator produces a new candidate $c_{t+1} \sim G_\theta(\cdot \mid x, c_t)$ at cost $C_{\mathrm{gen}}$. The new candidate patch $c_{t+1}$ may improve the program, but it may also leave it unchanged or make it worse.

We assume access to $k$ critic agents $\{a^1_{\mathrm{crit}}, \ldots, a^k_{\mathrm{crit}}\}$ that inspect $c_t$ without modifying it. Critic $i$ returns a cost-effective noisy diagnostic observation $z_t^{(i)}\in\{\mathrm{pass},\mathrm{fail}\}$ with likelihood $P_i(z\mid Y_t)$ and cost $C^i_{\mathrm{crit}}$.

A control policy $\pi$ orchestrates the program generation process with the goal of producing a program that satisfies $x$. It interacts with a set of agents $\mathcal{A} = \{a_{\mathrm{gen}}, a_{\mathrm{ver}}, a^1_{\mathrm{crit}}, \ldots, a^k_{\mathrm{crit}}\}$ over discrete time steps $t=1, \ldots, T$. At each step, the policy selects an agent $a_t \in \mathcal{A}$ based on the current history
$\mathcal{H}_t = \left(x, c_1, a_1, o_1, \ldots, a_{t-1}, o_{t-1}, c_t\right)$,
where $o_t$ denotes the observation returned by the selected agent. The quality of a policy is measured by its expected utility, defined as the value of the final outcome minus the accumulated agent costs. Formally, if an episode terminates at a step $T$ with candidate $c_T$, the expected utility is $U(\pi) = \mathbb{E}_{\pi}\left[R \cdot Y_T - \sum_{t=1}^{T} C(a_t)\right]$, where $R$ represents the reward for a correct solution ($Y_T=1$), and $C(a_t)$ is the cost of the selected agent: $C(a_{\mathrm{gen}}) = C_{\mathrm{gen}}$, $C(a^i_{\mathrm{crit}}) = C^i_{\mathrm{crit}}$, and $C(a_{\mathrm{ver}}) = C_{\mathrm{ver}}$.

Our goal is to derive a policy that maximizes this utility by estimating the \emph{action value} (or $Q$-\emph{value}) of each agent, thus quantifying the trade-off between the cost of an action and its expected contribution to solving the task.

\section{Bayesian Decision-Theoretic Control}
\label{sec:method}

\textbf{General design of Bayesian control.}
We adopt Bayesian decision theory through a POMDP abstraction. At time step $t$, the controller holds a candidate program $c_t$ and a belief state $b_t=P(Y_t=1\mid \mathcal{H}_t)$ over its correctness, and selects actions by trading off the cost of information, improvement and verification. The belief update process constitutes a sequential hypothesis test, iteratively accumulating evidence to decide between the null hypothesis $Y_t=0$ and the alternative $Y_t=1$. 

If $a_t = a_{\mathrm{gen}}$, the belief is pushed through a transition kernel $\mathcal{T}(b_t)$, which accounts for the probability that the generator introduces a bug into a correct program or fixes an incorrect one,
\[
\small
\mathcal{T} =
\begin{pmatrix}
P(Y_{t+1}=0 \mid Y_t=0) & P(Y_{t+1}=1 \mid Y_t=0) \\
P(Y_{t+1}=0 \mid Y_t=1) & P(Y_{t+1}=1 \mid Y_t=1)
\end{pmatrix},
\]
parameterized by the empirical fix probability $p_{01}=P(Y_{t+1}=1 \mid Y_t=0)$ and break probability $p_{10}=P(Y_{t+1}=0 \mid Y_t=1)$. Thus, after invoking the generator, the controller propagates its belief as $b_{t+1} = b_t(1-p_{10}) + (1-b_t)p_{01}$.

If $a_t = a^i_{\mathrm{crit}}$ and $z_t^{(i)}=z$, the belief is updated via Bayes' rule:
\[
\small
\begin{aligned}
b_{t+1}
&= P(Y_t=1 \mid \mathcal{H}_t, z_t^{(i)}=z) \\
&= \frac{P_i(z \mid Y_t=1)b_t}
        {P_i(z \mid Y_t=1)b_t + P_i(z \mid Y_t=0)(1-b_t)}.
\end{aligned}
\]
Critic $i$ reduces uncertainty about $Y_t$. This interaction constitutes an information purchase; the controller pays a low cost $C^i_{\mathrm{crit}}$ to acquire evidence before committing to refinement or verification.

The controller can submit the candidate to the oracle to resolve the episode. In this case, the controller receives the definitive outcome $Y_t = \mathcal{O}(c_t, x)$. While this interaction provides perfect information and determines the reward $R$, it incurs the highest cost $C_{\mathrm{ver}}$.

\textbf{Control via value of information.}
The controller implements a policy $\pi$ that maps the current belief state to an agent selection, $\pi(b_t) \in \mathcal{A}$, with the goal of maximizing expected future utility. We write the value function as $V(b_t)$; in finite-horizon implementations, the state also includes the remaining budget and which critics have already been queried. The Bellman optimality equation is
$V(b_t) =
\max \left\{
Q_{\mathrm{gen}}(b_t),
% Q_{\mathrm{crit}}(b_t),
\max_{i} Q_{\mathrm{crit}}^{i}(b_t),
Q_{\mathrm{ver}}(b_t)\right\}$,
where the action values for the generator, critic, and verifier are defined, respectively, as
\begin{align*}
Q_{\mathrm{gen}}(b_t)
&= \mathbb{E}[V(b_{t+1}) \mid b_t, a_{\mathrm{gen}}] -C_{\mathrm{gen}}, \\
% Q_{\mathrm{crit}}(b_t)
% &= -C_{\mathrm{crit}} + \mathbb{E}_{z}[V(b_{t+1}) \mid b_t, a_{\mathrm{crit}}], \\
Q_{\mathrm{crit}}^{i}(b_t)
&= \mathbb{E}_{z}[V(b_{t+1}) \mid b_t, a^i_{\mathrm{crit}}] -C^i_{\mathrm{crit}}, \\
Q_{\mathrm{ver}}(b_t)
&= b_t R- C_{\mathrm{ver}}.
\end{align*}

$Q_{\mathrm{gen}}$ represents the expected long-term value of improving the candidate program via the generator agent. $Q_{\mathrm{crit}}^{i}$ captures the expected value of acquiring information from critic $i$. $Q_{\mathrm{ver}}$ represents the immediate expected utility of stopping through oracle verification. To compute the optimal policy, we solve this Bellman recursion by finite-horizon dynamic programming over the belief state.

\textbf{Optimal control vs heuristics.} Orchestration heuristics, such as thresholding a confidence score, collapse the decision problem into a scalar boundary. This can be insufficient because the optimal policy depends on the posterior probability of correctness, action-specific likelihoods, costs, remaining budget, and which information sources have been consumed. For example, a low-confidence candidate before any critic has been queried may have high information value, making a critic call preferable to refinement. A low-confidence candidate after a failed public test may have little remaining diagnostic value, making refinement preferable. A static threshold can treat these cases identically, whereas the Bellman update separates them through the $Q$-values of the generator, critic, and verifier actions.

\textbf{Estimating critic likelihoods.}
For each critic $i$, we estimate the conditional distributions (critic likelihoods) $P_i(z \mid Y=1)$ and $P_i(z \mid Y=0)$ on a held-out calibration set of programs with oracle-labeled outcomes, using add-one smoothing under a beta-binomial model.

\textbf{Estimating transition probabilities.}
We estimate the transition probabilities $p_{01}$ and $p_{10}$ in the transition matrix $\mathcal{T}$ from refinement trajectories with oracle labels at consecutive steps, aggregating observed $(Y_t,Y_{t+1})$ pairs and applying beta-binomial smoothing. These counts come from a held-out iterative-refinement corpus, as detailed in Section \ref{sec:experiments}.

\section{Experimental Setup}
\label{sec:experiments}

%We experiment on three
%coding-task families and six generators. 
We investigate the conditions under which the Bayesian controller outperforms heuristic baselines and its failure modes. Additional setup details are provided in Appendix~\ref{appendix:setup}.

\subsection{Benchmarks and Generators}
\label{ssec:benchmarks}

\textbf{Benchmarks.} We experiment with nine benchmarks spanning three task families (Table~\ref{tab:benchmarks}). The \emph{function-level synthesis benchmarks} are LCB at three difficulty tiers, covering LeetCode problems from releases v1 through v6, MBPP+~\citep{liu2023mbpp+}, and HumanEval+~\citep{liu2023mbpp+}. The \emph{repository-level patch-generation benchmarks} are SWE-Bench Lite and SWE-Bench Verified~\citep{jimenez2024swebench}. The \emph{bug-fixing benchmarks} are HumanEvalFix~\citep{muennighoff2024octopack} and CodeContests~\citep{li2022alphacode}.

\textbf{Generators.}
We test six LLMs across two access modes (Table~\ref{tab:generators}), with 7--200B effective parameters. For each generator-benchmark cell, the generator's single-shot pass rate defines the initial prior pass probability $b_0=P(Y_0{=}1)$, the probability that a fresh candidate passes the oracle verifier before any critic evidence is observed. Across cells, $b_0$ ranges from $0.05$ to $0.96$ (Table~\ref{tab:prior_pass_rate}).

\subsection{Critics and Oracle Verifier}
\label{ssec:critics}
 
The oracle verifier represents an expensive-to-run test suite that deterministically checks the correctness of the solution. For LCB, it is a full hidden test suite; for SWE-Bench, it is a Podman-based harness; and for MBPP+/HumanEval+, it is represented by assert-based unit tests.

Critics provide lightweight but noisy evidence. The syntax critic $\text{Cr}_\text{syn}$ runs \texttt{ast.parse} on each modified file and is essentially free in wall time. The test critic $\text{Cr}_\text{test}$ runs public tests, using LiveCodeBench public examples (and the analogous public tests for MBPP+, HumanEval+, HumanEvalFix, and CodeContests). It serves as the main critic signal on these benchmarks. SWE-Bench has no public-test split distinct from the oracle verifier, so $\text{Cr}_\text{test}$ is undefined for SWE-Bench. The LLM critic $\text{Cr}_\text{llm}$ uses a separate small model, Claude Haiku-4.5 by default, prompted with the output of the \texttt{diff} command and the problem statement to return a one-token PASS/FAIL judgment. In Appendix \ref{ssec:critic_inform} (Table~\ref{tab:critic_gap}, Fig.~\ref{fig:critic_grid}), we report the per-critic informativeness measured as $\gamma_i = P_i(z{=}\mathrm{pass}\mid Y{=}1) - P_i(z{=}\mathrm{pass}\mid Y{=}0)$ across the full panel: $\gamma_i \approx 0$ means the critic carries no useful signal, while $|\gamma_i| \to 1$ means it is highly informative; values near $1$ indicate near-oracle alignment with correctness, and values near $-1$ indicate an inverted signal.

\textbf{Critic and oracle verifier cost vectors}
are deployment-specific hyperparameters. To obtain the cost vectors used in our experiments, we collect telemetry from a representative deployment. We measure generation cost, critic costs, and oracle verifier cost. For each action, we log wall-clock latency, token usage, and API cost. We use wall-clock latency as the primary proxy cost metric because it best captures the operational bottlenecks that matter in practice, including local process overhead, CI waiting time, container execution, and external API latency. Details on measurements and the exact vectors are given in Table~\ref{tab:action_latency}.

We report experimental results under two cost regimes that span the typical oracle/critic ratio space. The \emph{``slow oracle, fast critics''} regime sets $C_{\mathrm{ver}}{=}90$, $C_{\mathrm{crit}}^{\mathrm{syn}}{=}1$ for the syntax critic, $C_{\mathrm{crit}}^{\mathrm{test}}{=}2$ for the public-test critic, $C_{\mathrm{crit}}^{\mathrm{llm}}{=}5$ for the LLM critic, $C_{\mathrm{gen}}{=}10$ (e.g., a CI pipeline with frontier LLM judges). The \emph{``fast oracle, balanced critics''} regime sets $C_{\mathrm{ver}}{=}5$, $C_{\mathrm{crit}}^{\mathrm{syn}}{=}C_{\mathrm{crit}}^{\mathrm{test}}{=}C_{\mathrm{crit}}^{\mathrm{llm}}{=}1$, $C_{\mathrm{gen}}{=}10$ (e.g., a unit-test loop where verification is cheap). The fast-oracle regime represents bug-fixing settings such as HumanEvalFix and CodeContests, where verification is relatively cheap and generation is the dominant costly action. The slow-oracle regime represents SWE-Bench-like deployments, where full verification is a major bottleneck and critic gating has higher value. 
We set $R{=}100$ as a utility normalization rather than a measured latency. One verified correct solution is worth 100 utility units, which makes $C_{\mathrm{ver}}/R$ the posterior success threshold at which verification becomes cost-neutral, and we report $R$-$C_{\mathrm{ver}}$ sweeps to check that the qualitative regimes are not tied to this single scale choice.

\subsection{Baseline Policies}
\label{ssec:methods_compared}

We compare the Bayesian controllers against eight baseline policies grouped into two families. The stateless baselines operate over the same action space $\{\text{Cr}_\text{syn}, \text{Cr}_\text{test}, \text{Cr}_\text{llm}, \texttt{verify}, \texttt{regen}, \texttt{stop}\}$, while the iterative-refinement baselines use the same verifier and refinement budget but follow their published update rules.

\textbf{Stateless per-patch baselines}  take action on each patch in isolation, with no belief state carried across actions. \texttt{always\_verify} skips every critic and calls the oracle verifier on every patch. \texttt{best\_of\_N} generates $N{=}3$ independent patches, runs the oracle on each, and returns the best, paying $3 C_{\mathrm{gen}} + 3 C_{\mathrm{ver}}$ per instance. It isolates pure generator-diversity gains without critic assessment. The \texttt{gate\_}$\text{Cr}_i, i \in \{\mathrm{syn},\mathrm{test},\mathrm{llm}\}$ baselines run critic $\text{Cr}_i$ once and call the oracle only if it returns PASS; otherwise, request regeneration. \texttt{fixed\_pipeline} runs all critics in the order $\text{Cr}_\text{syn} \to \text{Cr}_\text{test} \to \text{Cr}_\text{llm}$ and verifies only if every critic returns PASS, testing the simple strategy of stacking all critics together with an AND gate.

\textbf{Iterative-refinement baselines.} We compare against the \emph{Self-Refine}~\citep{madaan2023self} and \emph{Reflexion}~\citep{shinn2023reflexion} coding-agent baselines. Both are run with up to four refinement steps per instance and the same verifier budget as the Bayesian controllers, and neither uses policy replays on our calibration trajectories. After each failed verification, Self-Refine prompts the generator to critique its own patch and regenerate, stopping when the verifier returns PASS or the refinement budget is exhausted. Reflexion extends Self-Refine with a verbal-memory buffer of past failures that is fed back into the regeneration prompt.

\subsection{Details of Bayesian Belief-State Policies}
\label{ssec:dp_vs_greedy}

\textbf{Bayesian controllers.} Our framework derives two controllers from the Bellman equation of
the POMDP (see also Section~\ref{sec:method}). Both maintain a
posterior $b = P(Y{=}1 \mid \text{evidence})$ and select actions to
maximize expected utility.

The \textbf{\texttt{bayesian\_greedy}} controller performs one-step lookahead. At each decision point, with current candidate $c_t$ and belief $b_t$, it evaluates each action's immediate $Q$-value under the assumption that the next move is \texttt{verify}, and selects the action with largest value. When computing the value of a generator call, the controller treats the resulting patch as a fresh sample and resets its belief to the prior pass rate $b_0=P(Y_0{=}1)$, rather than propagating the current belief through the measured fix/break probabilities. As a result, it cannot value multi-step refinement chains.

The \textbf{\texttt{bayesian\_DP}} controller performs finite-horizon backward induction over states $(b,h)$, where $b$ is the posterior probability of correctness and $h\in\{0,\ldots,H\}$ is the remaining planning depth. We discretize the belief interval to a uniform 51-point grid and solve the Bellman recursion backward over horizon $H{=}3$, using the empirically estimated transition kernel $\hat P(\text{fix}\mid\text{broken})$ and $\hat P(\text{break}\mid\text{correct})$ for the regeneration branch. The kernel propagates beliefs across regenerations, allowing DP to value multi-step critic-refine-verify chains that the greedy controller cannot see.

\textbf{Estimating the parameters of Bayesian control policies.}
We estimate the quantities required by the Bayesian controllers from
held-out calibration data (split 75\% for training, 25\% for testing), namely the initial belief $b_0=P(Y_0{=}1)$, also
called the prior pass probability
(Table~\ref{tab:prior_pass_rate}), the critic likelihoods, and the
refinement-transition probabilities $\hat P(\text{fix}\mid\text{broken})$
and $\hat P(\text{break}\mid\text{correct})$.

Single-shot calibration draws $m{=}3$ independent patches per instance and runs the full critic stack, followed by the oracle verifier. From the resulting per-patch records, we estimate the prior pass probability $b_0=P(Y_0{=}1)$, which is the probability that a fresh patch is correct before observing critic outputs, and the critic likelihoods $P_i(z{=}\text{pass}\mid Y_0{=}y)$. 

For each critic $i$ and oracle label $y\in\{0,1\}$, we estimate the time-homogeneous critic likelihood $\hat{P}_i(z=\text{pass}\mid Y_t{=}y)$ as the empirical fraction of patches with label $y$ on which critic $i$ returns PASS, with one pseudo-PASS and one pseudo-FAIL added for smoothing:
\[
\hat{P}_i(z=\text{pass}\mid Y_t{=}y)
=
\frac{n_{y,i}^{\text{pass}} + 1}{n_{y,i} + 2}.
\]
Here, $n_{y,i}^{\text{pass}}$ is the number of calibration patches with the oracle label $y$ on which the critic $i$ returns PASS, and $n_{y,i}$ is the total number of calibration patches with the oracle label $y$ assessed by the critic $i$. We use the same critic likelihood model at each decision step.

Transition probabilities are estimated from consecutive
parent-child generation pairs in the iterative-refinement corpus,
using smoothing:
\[
\hat{p}_{01} = \frac{n_{0\to1}+1}{n_{\text{from }0}+2},
~
\hat{p}_{10} = \frac{n_{1\to0}+1}{n_{\text{from }1}+2}.
\]

\subsection{Evaluation Metric}
\label{ssec:per_cell}

For each generator/benchmark/cost-vector triple, we evaluate the ten policies above on the same instance cohort and patches, yielding a vector of mean utilities $\bar U_\pi$. The headline metric is $\Delta_\pi = \bar U_\pi - \bar U_{\texttt{always\_verify}}$, the per-instance utility gap against the reference policy. Thus, $\Delta_\pi > 0$ beats the ``oracle on every patch'' default, $\Delta_\pi = 0$ ties, and $\Delta_\pi < 0$ underperforms. Paired 95\% bootstrap CIs on $\Delta_\pi$ are used to identify the winning policies. 

\section{Results}
\label{sec:results}

We organize the results around three questions: \emph{where} in the
$(P(Y{=}1),\,C_{\mathrm{ver}}/R)$ plane each policy dominates
(\S\ref{ssec:regimes}); \emph{which} policy wins on held-out eval
splits when all baselines -- including the published refinement
agents Self-Refine~\citep{madaan2023self} and
Reflexion~\citep{shinn2023reflexion} -- are evaluated head-to-head
(\S\ref{ssec:policy_eval}); and \emph{what} signal makes the
controller's choice rational (\S\ref{ssec:critic_inform_main}). We then
sweep the cost vector (\S\ref{ssec:sensitivity}) and discuss how
our framework enables better uncertainty quantification for agents
(\S\ref{ssec:uncertainty}).

\subsection{Regimes of Control Policy Prevalence Based on Priors and Cost-Reward Ratios}
\label{ssec:regimes}
\begin{figure}[t]
\centering
\includegraphics[width=\linewidth]{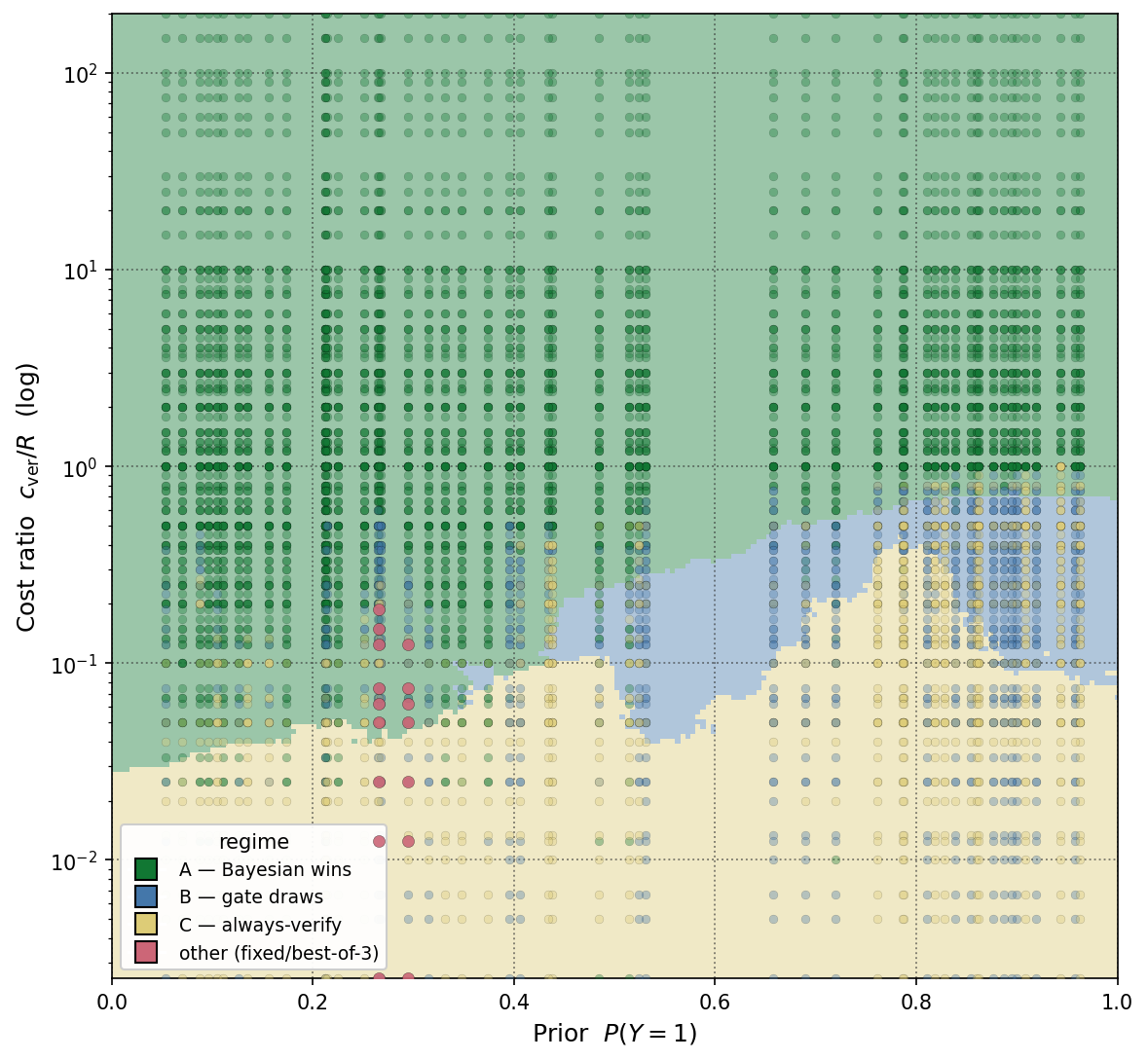}
\caption{Empirical decision regions over the
$(P(Y{=}1),\,C_{\mathrm{ver}}/R)$ plane: $k$-NN ($k{=}300$) over
7{,}020 sweep points. Colors: \textbf{A} (green) Bayesian;
\textbf{B} (blue) \texttt{gate($\text{Cr}_\text{test}$)};
\textbf{C} (yellow) \texttt{always\_verify}; \emph{other} (coral).
See \S\ref{ssec:regimes} for the full regime description.}
\label{fig:regime_map}
\end{figure}

Figure~\ref{fig:regime_map} aggregates 7{,}020 sample points
pooled across all (benchmark, generator) pairs and the full
$(R,\,C_{\mathrm{ver}})$ sweep grid into a single empirical
decision map (grid resolution in
Appendix~\ref{appendix:regime_grid}). The plot indicates \emph{which policy wins} at each
coordinate in the $(P(Y{=}1),\,C_{\mathrm{ver}}/R)$ plane.
Three regimes emerge, each
driven by a distinct mechanism that ties back to the per-cell
prior (Table~\ref{tab:prior_pass_rate}) and per-critic
informativeness (Table~\ref{tab:critic_gap}).

\textbf{Regime A: Bayesian wins (green, $\mathbf{C_{\mathrm{ver}}/R \gtrsim 1}$).}
When verification is at least as expensive as the reward,
paying $C_{\mathrm{ver}}$ on every patch is structurally
unprofitable. The Bayesian policies maintain a posterior
$P(Y{=}1\mid z^{(1)},\ldots,z^{(k)})$ over the critics and
verify only when the expected value of information exceeds
$C_{\mathrm{ver}}$. 
\texttt{bayesian\_DP} also
pulls measurably ahead of the myopic \texttt{bayesian\_greedy}:
when each round of verification is unaffordable, planning
multiple steps ahead over the transition kernel
lets the controller defer
verification across refinement rounds until the posterior is
sharp enough to commit.

\textbf{Regime B: gate($\text{Cr}_\text{test}$) draws (blue wedge, $\mathbf{10^{-1} \lesssim C_{\mathrm{ver}}/R \lesssim 1}$).}
At moderate verification cost, the optimal Bayes decision often
is a gate on the public-test critic. On cells where
$\text{Cr}_\text{test}$ is highly informative
($\gamma_\text{test} \gtrsim 0.7$ on LCB and on most HumanEvalFix
and CodeContests cells; $\gamma_\text{test}\gtrsim 0.9$ on LCB-easy
specifically, Table~\ref{tab:critic_gap}; SWE-Bench has no
public-test critic distinct from the oracle and is excluded), a single
PASS observation moves the posterior past the verification
threshold by itself, so adding $\text{Cr}_\text{syn}$ or
$\text{Cr}_\text{llm}$ to the update rarely changes the decision
$\text{Cr}_\text{test}$ already implies. The Bayesian
controllers and \texttt{gate}($\text{Cr}_\text{test}$) make the
same decision in most instances, but the gate baseline ties or
marginally wins from the absence of inference overhead. The
wedge narrows on cells where the public test is a weaker signal
(MBPP+, HumanEval+: $\gamma_\text{test} \in [0.07, 0.75]$).

\textbf{Regime C: always\_verify wins (yellow, $\mathbf{C_{\mathrm{ver}}/R \lesssim 10^{-1}}$).}
When verification is essentially free, the cost of running any
critic -- let alone of a Bayesian update -- approaches or
exceeds $C_{\mathrm{ver}}$ itself, so blind verification
dominates. Two pockets appear. The bottom-right (high
prior, low cost) covers cells where the step-0 prior is already
above the indifference threshold -- almost every patch passes
and \texttt{always\_verify} simply collects the reward without
paying for information it does not need. The bottom-left
(low prior, low cost) covers cells where most patches fail; the
few passes are cheap to harvest by verifying everything, and
any critic-based skip risks dropping a true pass for a saving
that does not justify the loss.

\subsection{Policy Comparison}
\label{ssec:policy_eval}

\begin{figure}[t]
\centering
\begin{subfigure}[b]{\linewidth}
  \centering
  \includegraphics[width=0.75\linewidth]{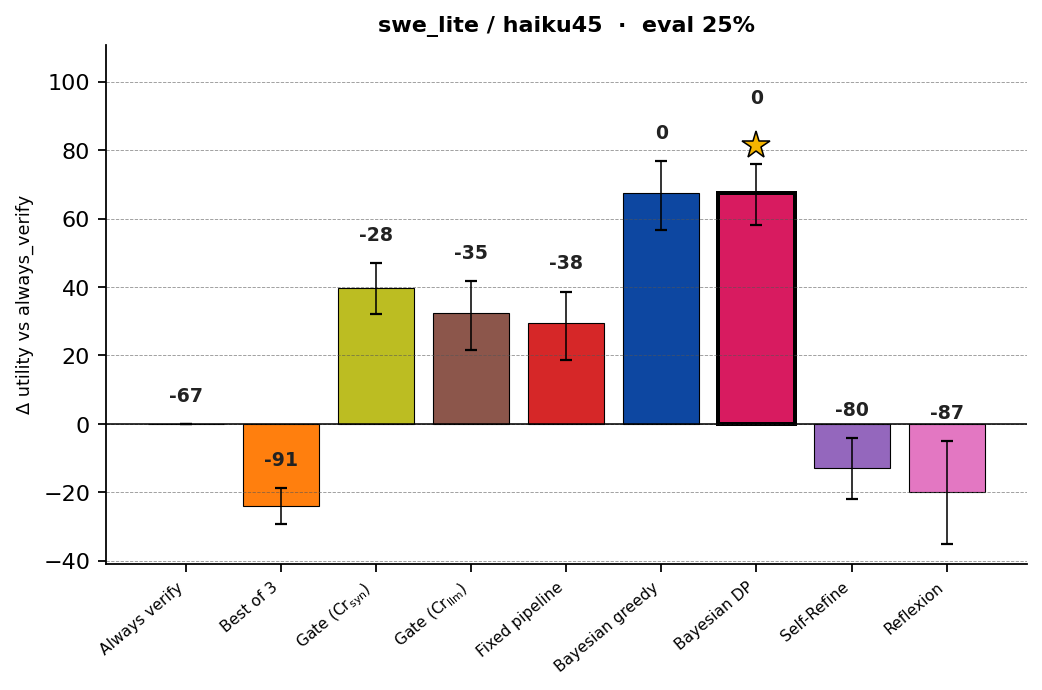}
  \label{fig:policies_main_eval}
\end{subfigure}
\caption{Per-policy $\Delta$ utility against \texttt{always\_verify}
on SWE-Bench Lite / \texttt{claude-haiku-4.5}. 
Gold stars mark per-cell winners; numbers above
bars are absolute mean utility; error bars are paired-bootstrap
95\% CIs ($B{=}1000$). Discussion in \S\ref{ssec:policy_eval};
per-cell evaluation-set bars across all (benchmark, generator) pairs in Appendix~\ref{appendix:policies_per_model}; full train-vs-evaluation panel in Appendix~\ref{appendix:train_vs_eval}.}
\label{fig:policies_eval_main}
\end{figure}

Figure~\ref{fig:policies_eval_main} presents a representative experiment comparing all policies on SWE-Bench Lite with \texttt{claude-haiku-4.5}. SWE-Bench Lite uses the slow oracle ($C_{\mathrm{ver}}{=}90$), so its cells fall into regime~A of Figure~\ref{fig:regime_map}, where the framework predicts that the Bayesian variants win. The two Bayesian variants are effectively tied for the best performance: \texttt{bayesian\_greedy} and \texttt{bayesian\_DP} both substantially improve over \texttt{always\_verify}. Among the remaining baselines, the syntax gate \texttt{gate(\(\text{Cr}_{\text{syn}}\))} is the strongest but still well below the Bayesian controllers; the LLM-judge gate \texttt{gate(\(\text{Cr}_{\text{llm}}\))} and \texttt{fixed\_pipeline} improve over \texttt{always\_verify} by smaller margins. The public-test gate \texttt{gate(\(\text{Cr}_{\text{test}}\))} is omitted on SWE-Bench because there is no public-test critic distinct from the oracle verifier (see~\S\ref{ssec:critics}). The refinement baselines SR and Reflexion, together with \texttt{best\_of\_3}, underperform \texttt{always\_verify} in this setting. Detailed results for each dataset and generator are presented in Table~\ref{tab:full_results} in Appendix~\ref{sec:full_results}.

\subsection{Critic Informativeness}
\label{ssec:critic_inform_main}
Bayesian controllers are effective only to
the extent that the underlying critic likelihoods provide useful
evidence (the critic informativeness is reported in Appendix \ref{ssec:critic_inform}).
They are particularly valuable when this evidence is distributed across
multiple imperfect critics rather than relying on a single near-oracle
critic. By combining several moderately informative signals, the
Bayesian update creates a sharper, more accurate posterior belief than
any individual critic could provide on its own. This is achieved because
the controllers automatically weight each critic according to its
likelihood gap, denoted as $\gamma_i$. 
Consequently, low-gap critics
have very little effect on the update; in many function-level
synthesis cells, this includes the syntax critic
($\text{Cr}_\text{syn}$). In contrast, informative and complementary
critics, such as the public-test critic ($\text{Cr}_\text{test}$), are weighted heavily and
meaningfully improve the belief state.
Ultimately, this
automatic weighting and aggregation of imperfect signals directly
explain the performance gains observed with the
\texttt{bayesian\_greedy} and \texttt{bayesian\_DP} controllers in
Regime~A.

\begin{table}[t]
\centering
\resizebox{0.48\textwidth}{!}{\Large
\begin{tabular}{l|cc|c}
\toprule
\textbf{Method} & \textbf{LCB-Medium} & \textbf{LCB-Hard} & \textbf{Avg} \\
\midrule
Perplexity & .156 & .578 & .367 \\
Seq. Prob. & .789 & \underline{.814} & \underline{.801} \\
Tool Success Rate & \textbf{.842} & .747 & .795 \\
Bayes Belief State & \underline{.820} & \textbf{.911} & \textbf{.866} \\
\bottomrule
\end{tabular}
}
\caption{PRR$\uparrow$ of uncertainty quantification methods before the final verification step on LCB-Medium and LCB-Hard, using the SAGE agent~\cite{suri2025structured} with \texttt{gpt-oss-20b}. The \textit{Avg} column reports the average score across the two subsets.
}
\label{tab:lcb_uncertainty}
\end{table}

\subsection{Cost-Regime and Parameter Sensitivity}
\label{ssec:sensitivity}
\begin{figure}[t]
\centering
\includegraphics[width=\linewidth]{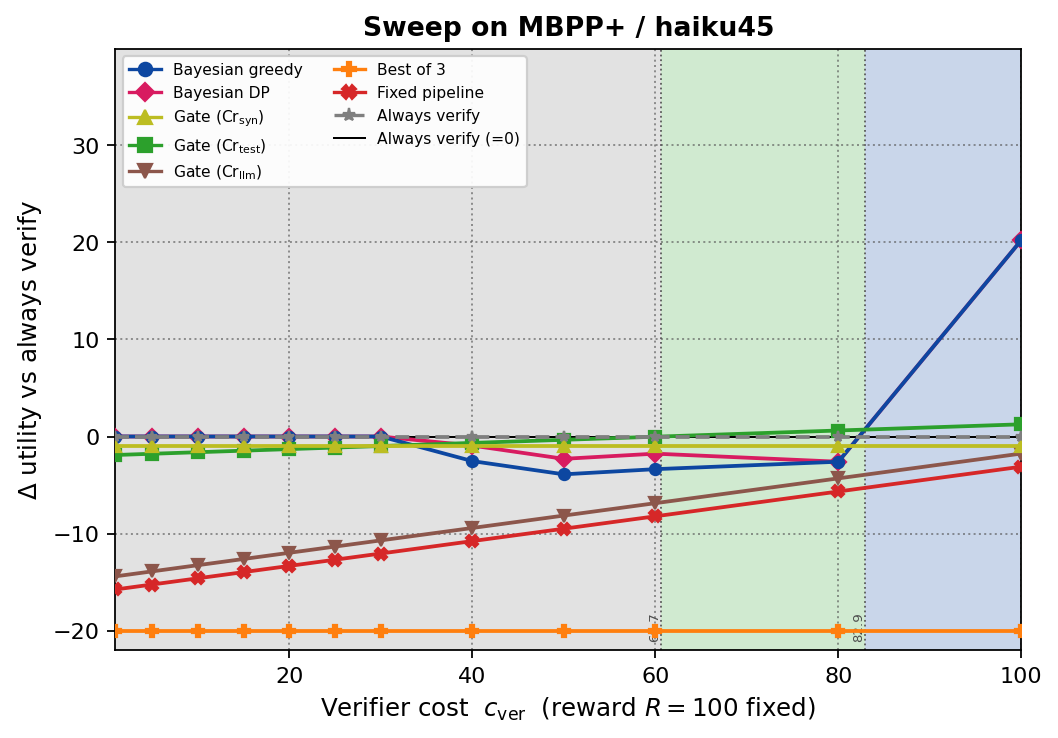}
\caption{Per-policy $\Delta$ utility relative to \texttt{always\_verify}
on MBPP+ / \texttt{claude-haiku-4.5} as a function of verifier cost
$C_{\mathrm{ver}}$ ($C_{\mathrm{crit}}^{\mathrm{syn}}{=}1$,
$C_{\mathrm{crit}}^{\mathrm{test}}{=}2$,
$C_{\mathrm{crit}}^{\mathrm{llm}}{=}5$, $C_{\mathrm{gen}}{=}10$, $R{=}100$ fixed). Shaded regions mark the empirically dominant policy in each verifier-cost sub-range; gray: \texttt{always\_verify} / near-tied region at low verifier cost; green: \texttt{gate($\text{Cr}_\text{test}$)}; blue: Bayesian. Dotted verticals mark empirical leader-change boundaries obtained from intersections of the plotted $\Delta$ segments; their locations are benchmark- and model-dependent. Per-benchmark and per-(benchmark, generator) panels in Appendix~\ref{appendix:per_bench_cver}.}
\label{fig:cver_sweep_main}
\end{figure}

Figure~\ref{fig:cver_sweep_main} sweeps $C_{\mathrm{ver}} \in [1,
100]$ on MBPP+ / \texttt{claude-haiku-4.5} at fixed $R{=}100$, with
the other costs held at their default values. The sweep traverses
all three regimes within a single cell, illustrating the C\,$\to$\,B
$\to$\,A handoff: at low $C_{\mathrm{ver}}$, \texttt{always\_verify}
wins because the prior is high enough that blind verification has
a positive expected value; at moderate $C_{\mathrm{ver}}$,
\texttt{gate($\text{Cr}_\text{test}$)} exploits the public-test
critic; at high $C_{\mathrm{ver}}$, the Bayesian variants take over
because the gate policy is forced to refuse too many uncertain
patches. The per-benchmark/per-generator $C_{\mathrm{ver}}$ sweeps
covering all 54 (benchmark, model) pairs are in Appendix~\ref{appendix:per_bench_cver}.

\subsection{Uncertainty for Coding Agents}
\label{ssec:uncertainty}

Beyond using Bayesian controllers to optimize expected utility, we investigate whether the belief state can quantify uncertainty for state-of-the-art LLM-based controllers (see details in Appendix \ref{sec:uq_results}). In this experiment, we use a SAGE agent \citep{suri2025structured} with \texttt{gpt-oss-20b}. The Bayesian model is used only post hoc to aggregate the critic and verifier feedback into a belief that the code is 
correct before the final oracle verification outcome is observed. Table~\ref{tab:lcb_uncertainty} reports results on LCB-Medium and LCB-Hard. On average, the Bayesian belief state achieves the best performance in ranking incorrect code generations compared to token-probability uncertainty baselines and a raw tool success rate baseline that simply averages observed tool outcomes. This suggests that a belief state obtained using
Bayesian aggregation of the critic and non-final verifier observations can provide a useful uncertainty score even for coding agents controlled by an LLM.

\section{Conclusion}
\label{sec:conclusion}

We formulated LLM-based code-generation orchestration as a cost-sensitive sequential hypothesis testing problem, where the controller maintains a posterior belief over candidate correctness and uses it to choose between criticism, refinement, verification, and stopping. Our experiments reveal three regimes. (A) Bayesian controllers work best when verification is expensive and critics are informative but imperfect, because they can aggregate multiple noisy signals and downweight weak critics. (B) A simple public-test gate is sufficient when the public-test critic is already near-oracle, leaving little benefit for more complex Bayesian aggregation. (C) Unconditional verification dominates when verification is cheap or the generator's prior probability of correctness is high, since gathering additional information is then not worth the cost. This helps identify when Bayesian control is useful and when simpler orchestration policies are sufficient.

Beyond control, our results suggest that the belief can be used as an interpretable uncertainty score to identify whether the generated code is correct. This makes the proposed Bayesian framework complementary to existing coding agents: it can be used either to control the agentic loop directly or to provide uncertainty scores for trajectories produced by external controllers. Investigating this research direction more systematically is an important direction for future work.

\section*{Limitations}
The benchmarks used in our evaluation primarily focus on Python, since most widely used datasets and benchmarks are Python-based. However, because modern LLMs have broad reasoning capabilities across code, we are confident that our findings generalize well to other programming languages. These benchmarks also contain a limited number of instances. Still, our experiments cover six models and nine datasets, which supports the robustness of our findings. Finally, different prompts or judge models may affect the results of the LLM-based critic. However, we use a standard prompt and a modern default model, which helps support the reliability of our findings. We focus only on coding-agents and code generation; however, the proposed framework could be utilized in any agentic workflow with orchestration.

% \bibliographystyle{acl_natbib} % The ACL style loads this automatically
%\bibliography{references}

\appendix
\clearpage
\section{Additional Details on Experimental Setup}
\label{appendix:setup}

\subsection{Models}

Table~\ref{tab:generators} lists the closed-API and open-weight generator models.

\begin{table}[h]
\centering\small
\begin{tabular}{ll}
\toprule
\textbf{Access} & \textbf{Model} \\
\midrule
Closed-API & \texttt{gpt-5-mini} \\
           & \texttt{qwen3-coder} \\
           & \texttt{claude-haiku-4.5} \\
           & \texttt{claude-sonnet-4.5} \\
\midrule
Open-weight & \texttt{Qwen2.5-Coder-32B-Instruct} \\
            & \texttt{gpt-oss-20b} \\
\bottomrule
\end{tabular}
\caption{Generator models used in the experiments. Closed-API models are served via OpenRouter; open-weight models are served locally via vLLM.}
\label{tab:generators}
\end{table}

\subsection{Dataset Statistics}

Table~\ref{tab:benchmarks} presents the benchmarks and datasets used in the experiments, their evaluation-set sizes, and verifier types. Table~\ref{tab:action_latency} reports the measured per-action wall-clock latency across all benchmarks.

\begin{table}[h]
\centering\small
\begin{tabular}{lrl}
\toprule
\textbf{Benchmark} & \textbf{N} & \textbf{Verifier} \\
\midrule
\multicolumn{3}{l}{\emph{Function-level synthesis}} \\
LCB-hard           & 102 & hidden tests \\
LCB-medium         & 207 & hidden tests \\
LCB-easy           & 135 & hidden tests \\
MBPP+              & 378 & assert tests \\
HumanEval+         & 164 & assert tests \\
\midrule
\multicolumn{3}{l}{\emph{Repository-level patch generation}} \\
SWE-Bench Lite     & 300 & SWE harness \\
SWE-Bench Verified & 500 & SWE harness \\
\midrule
\multicolumn{3}{l}{\emph{Bug-fixing}} \\
HumanEvalFix       & 164 & assert tests \\
CodeContests       & 165 & hidden tests \\
\bottomrule
\end{tabular}
\caption{Evaluation benchmarks grouped by task, with dataset size and verifier type.
\label{tab:benchmarks}}
\end{table}

\begin{table*}[t]
\centering\small
\setlength{\tabcolsep}{4pt}
\begin{tabular}{lrrrrrrrr}
\toprule
& \multicolumn{2}{c}{$a_{\mathrm{gen}}$ (s)} & \multicolumn{2}{c}{$a_{\mathrm{ver}}$ (s)} & $L_0$ & $L_2$ & $L_3$ & Gen \$ \\
\cmidrule(lr){2-3}\cmidrule(lr){4-5}\cmidrule(lr){6-8}
Benchmark    & med    & p90    & med        & p90    & med                & med                & med    & ($10^{-3}$) \\
\midrule
\multicolumn{9}{l}{\emph{Function-level synthesis}} \\
LCB-easy     & $3.4$  & $5.2$  & $0.20$     & $0.33$ & $<\!0.001$         & $0.054$            & $1.46$ & $2.4$ \\
LCB-medium   & $6.8$  & $8.6$  & $0.25$     & $0.41$ & $<\!0.001$         & $0.059$            & $1.74$ & $5.5$ \\
LCB-hard     & $7.6$  & $9.8$  & $0.30$     & $0.50$ & $0.001$            & $0.062$            & $1.74$ & $6.3$ \\
MBPP+        & $1.3$  & $2.8$  & $0.05$     & $0.08$ & $<\!0.001$         & $0.063$            & $1.24$ & $0.5$ \\
HumanEval+   & $2.2$  & $3.1$  & $0.02$     & $0.03$ & $<\!0.001$         & $0.037$            & $1.14$ & $1.6$ \\
\midrule
\multicolumn{9}{l}{\emph{Bug-fixing}} \\
HumanEvalFix & $2.1$  & $3.4$  & $0.043$    & $0.050$ & $0.021^{\ddagger}$ & $0.021^{\ddagger}$ & $1.84$ & $3.0$ \\
CodeContests & $4.0$  & $9.0$  & $0.068$    & $0.879$ & $0.021^{\ddagger}$ & $0.021^{\ddagger}$ & $1.84$ & $5.0$ \\
\midrule
\multicolumn{9}{l}{\emph{Repository-level patches }} \\
SWE-Bench    & $10.0$ & $31.4$ & $1.9^{\|}$ & $682$  & $0.362$            & $2.02$             & $1.91$ & $6.0$ \\
\bottomrule
\end{tabular}
\caption{Measured per-action wall-clock latency across all benchmarks.
Claude-haiku-4.5 is used throughout, except the SWE-Bench row, which
pools $5$ models on $4$ held-out instances. Synthesis rows
(LCB-easy/medium/hard, MBPP+, HumanEval+) are calibration runs with
$n{=}30$ instances per benchmark ($n{=}29$ for LCB-hard). Bug-fixing
rows (HumanEvalFix, CodeContests) use the instrumented haiku-4.5 pilot
($15$ tasks $\times 2$ variants); their $L_0$ and $L_2$ entries report
the local-process critic median pooled over $L_0$--$L_2$ ($n{=}17$).
SWE-Bench critics are Docker pytest runs ($n{=}183$ pooled
$L_0$--$L_3$). All timing is in seconds (median / p90 where reported).
$L_0{=}$regex syntax check, $L_2{=}$local public-test execution (Docker
on SWE-Bench), and $L_3{=}$LLM judge instrumented separately on $60$
HEFix patches with haiku-4.5; SWE-Bench $L_3$ reports Docker
\texttt{critic\_mid}. The last column is per-generate API cost in
milli-dollars at OpenRouter haiku-4.5 rates; local critics and the local
verifier incur no API charge. $^{\|}$SWE-Bench Docker verifier is
bimodal: heavy-suite \texttt{psf/requests} median ${\approx}672$\,s
($n{=}35$) versus fast-fail sub-second ($n{=}25$); pooled median
$1.9$\,s, pooled p90 $682$\,s. $^{\ddagger}$Local critic median pooled
across $L_0$--$L_2$.}
\label{tab:action_latency}
\end{table*}

\subsection{Critic Informativeness}
\label{ssec:critic_inform}

\begin{figure*}[!t]
  \centering
  \includegraphics[width=\linewidth]{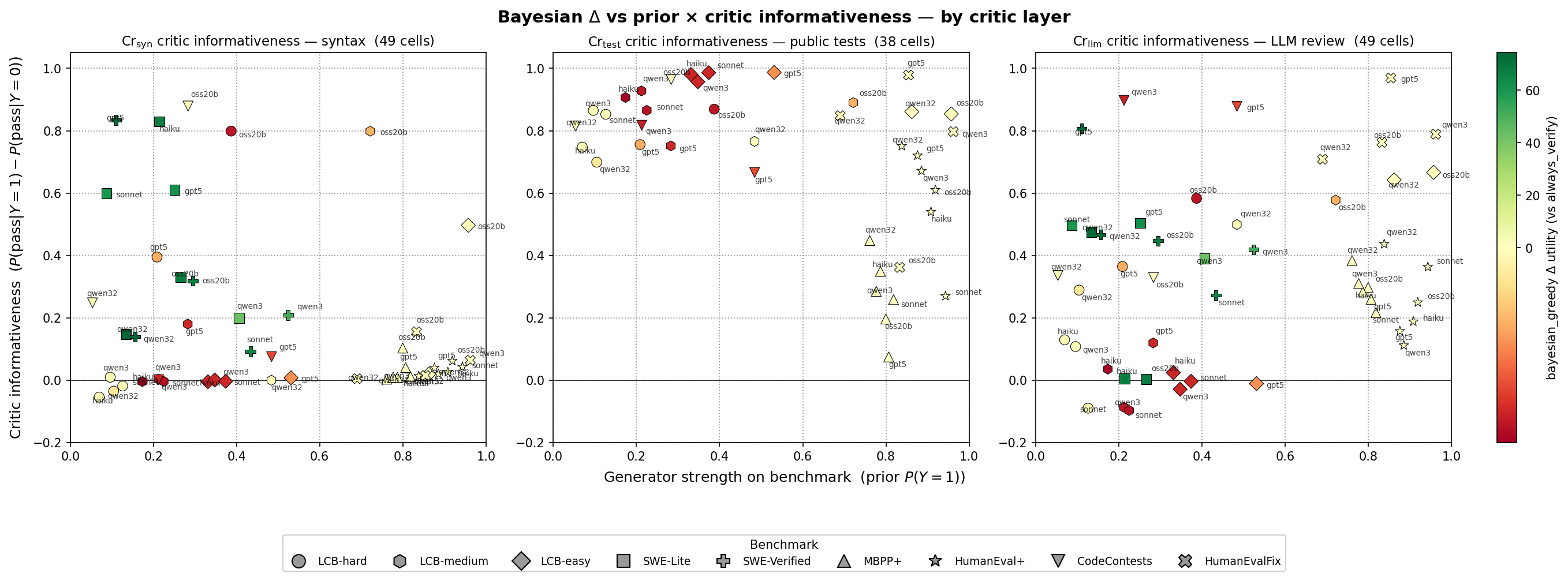}
  \caption{Critic informativeness $\gamma_i =
  P_i(z{=}\mathrm{pass}\mid Y{=}1) -
  P_i(z{=}\mathrm{pass}\mid Y{=}0)$ vs prior $P(Y{=}1)$ across
  the available benchmark--generator pairs, one subplot per critic layer.
  Point color indicates the Bayesian $\Delta$ for each benchmark--generator pair.
  SWE-Bench cells do not appear in the $\text{Cr}_\text{test}$ (public tests)
  panel: SWE-Bench has no public-test split distinct from the oracle
  verifier (the \texttt{test\_patch} is the verifier), so there is no
  independent public-test critic to estimate a gap for on those cells.}
  \label{fig:critic_grid}
\end{figure*}

The controller's posterior $P(Y{=}1\mid z^{(1)},\ldots,z^{(k)})$ is only as
useful as the underlying likelihoods $P_i(z \mid Y)$. We measure
the per-critic informativeness $\gamma_i =
P_i(z{=}\mathrm{pass}\mid Y{=}1) -
P_i(z{=}\mathrm{pass}\mid Y{=}0)$ across the full panel:
$\gamma_i \approx 0$ means the critic provides little information about correctness,
$\gamma_i \to 1$ means it is near-oracle (Table~\ref{tab:critic_gap},
Fig.~\ref{fig:critic_grid}). 

First, $\text{Cr}_\text{syn}$ is the least informative critic on
average. On MBPP+, HumanEval+, and LCB-easy every generator has
$|\gamma_\text{syn}| \leq 0.10$, and on LCB-medium four of six
generators have $|\gamma_\text{syn}| \leq 0.005$ -- the syntax
check passes on both $Y{=}1$ and $Y{=}0$ candidates and tells the
controller nothing, because modern instruction-tuned generators
produce parseable code regardless of whether it is semantically
correct.

Second, $\text{Cr}_\text{test}$'s informativeness tracks the
relationship between public and hidden tests on each benchmark, not
the generator. It is near-oracle ($\gamma_\text{test} \gtrsim 0.7$)
on LCB at all three difficulty tiers, on both SWE-Bench variants,
and on CodeContests -- benchmarks where the public test suite either
\emph{is} the hidden suite or is strongly predictive of it. On MBPP+
and HumanEval+, $\gamma_\text{test}$ drops to $0.07$--$0.75$ because
the public-test critic catches surface failures, but the hidden test
suite catches behavioral failures the public asserts pass on these
benchmarks; the public-test critic mainly detects failures covered by the visible assertions on those pairs. On average,
$\text{Cr}_\text{test}$ is the most informative of the three
critics. This explains why single-critic gating on
$\text{Cr}_\text{test}$ is the dominant policy over an entire
sub-range of verifier costs on some benchmark--generator pairs -- e.g.\ the green band
in Fig.~\ref{fig:cver_sweep_main} on MBPP+ /
\texttt{claude-haiku-4.5}, where \texttt{gate($\text{Cr}_\text{test}$)}
beats both \texttt{always\_verify} and the Bayesian variants for
moderate $C_{\mathrm{ver}}$.

Third, $\text{Cr}_\text{llm}$ is consistently weaker than $\text{Cr}_\text{test}$
on benchmark--generator pairs where the latter is near-oracle ($\gamma_\text{llm} < 0.5$
throughout LCB and SWE-Bench, often below $0.3$ on the LCB rows),
but becomes the dominant signal on HumanEvalFix, where
$\gamma_\text{llm} \geq 0.7$ for all six generators and the problem
statement gives the judge enough context to read the diff
semantically. This is what gives \texttt{bayesian\_greedy} and
\texttt{bayesian\_DP} their regime-A wins: when no single critic is
near-oracle, the controller updates the belief through Bayes' rule
on each critic's per-setup likelihood $P_i(z\mid Y)$, so the
contribution of critic $i$ scales with its informativeness
$\gamma_i$ -- a near-zero-gap critic barely moves the belief, while
two moderately-informative critics compose into a posterior sharper
than either alone produces.

\subsection{Regime-Map Sweep Grid}
\label{appendix:regime_grid}

The empirical decision map in Figure~\ref{fig:regime_map} pools
$7{,}020$ sample points. These are the Cartesian product of all
$54$ (benchmark, generator) pairs with a $(C_{\mathrm{ver}},R)$
sweep grid of $13$ verifier costs
$C_{\mathrm{ver}}\in\{1,5,10,20,25,30,50,60,75,90,100,150,200\}$
and $10$ rewards
$R\in\{1,10,20,25,50,75,100,150,200,400\}$, giving
$54\times13\times10=7{,}020$ points. At each point, we record the
utility-maximizing policy; the background fill in
Figure~\ref{fig:regime_map} is a $k$-NN ($k{=}300$) smoothing of
that per-point winner over the
$(P(Y{=}1),\,C_{\mathrm{ver}}/R)$ plane. The remaining costs are
held at the canonical values used throughout the paper
($c_\mathrm{gen}{=}10$, $c_{C_\text{syn}}{=}1$,
$c_{C_\text{test}}{=}2$, $c_{C_\text{llm}}{=}5$).

\subsection{Per-benchmark Generator Locations}
\label{appendix:per_bench_grids}

\begin{figure*}[!htb]
  \centering
  \includegraphics[width=\linewidth]{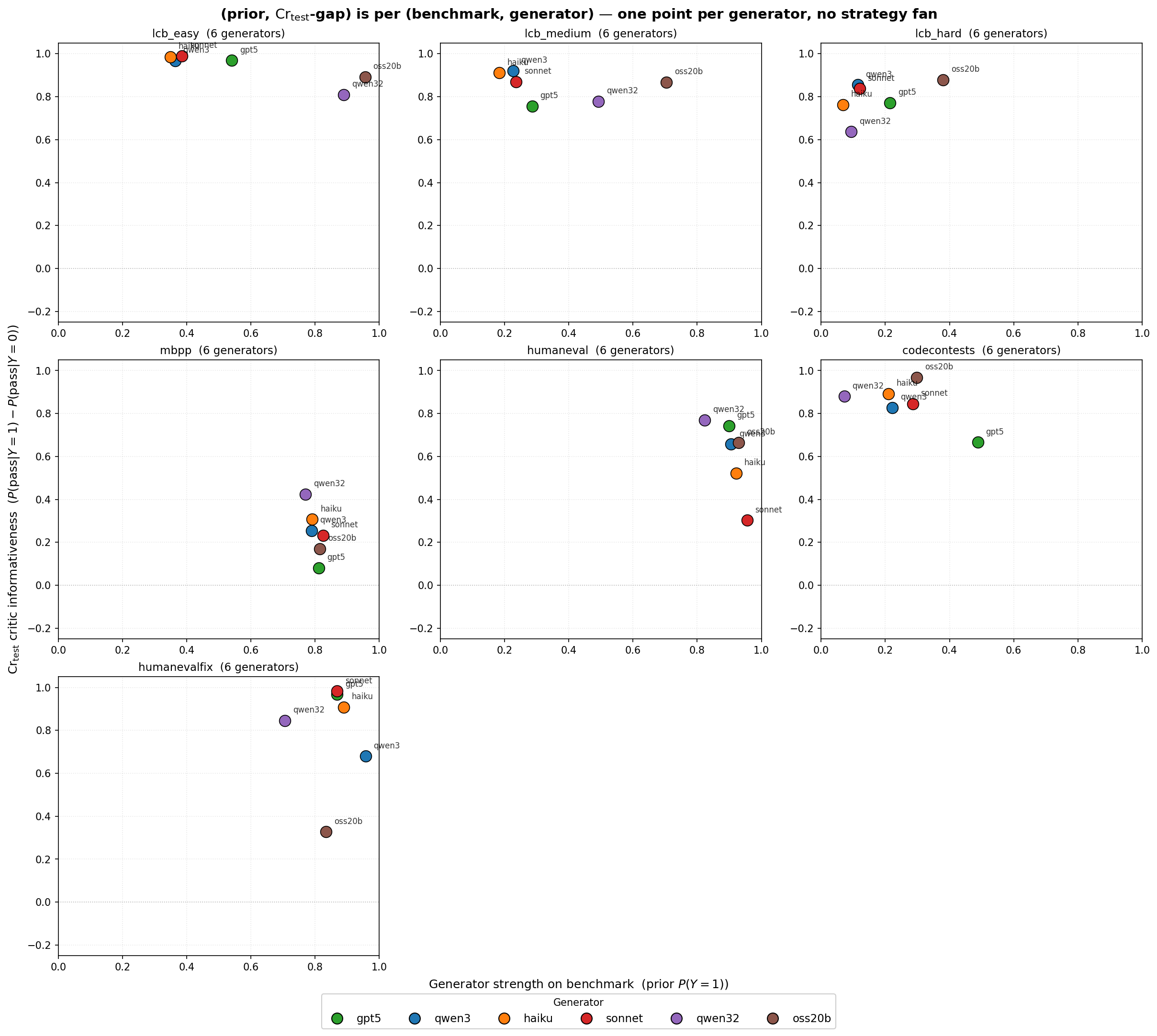}
  \caption{Per-benchmark location of each generator in
  ($P(Y{=}1)$, $\text{Cr}_\text{test}$ critic informativeness
  $\gamma_{\mathrm{test}} =
  P_{\mathrm{test}}(z{=}\text{pass}\mid Y{=}1) -
  P_{\mathrm{test}}(z{=}\text{pass}\mid Y{=}0)$). One sub-panel per
  benchmark, each dot one generator. The position of each cell in
  this plane determines which regime (A/B/C in
  Figure~\ref{fig:regime_map}) it falls into.
  SWE-Bench Lite and SWE-Bench Verified are omitted: their oracle
  verifier is the project's own test suite, so there is no public-test
  critic $\text{Cr}_\text{test}$ distinct from the verifier to plot.}
  \label{fig:generator_locations}
\end{figure*}

Figure~\ref{fig:generator_locations} plots where each benchmark–generator pair falls in the $(P(Y{=}1),\,\text{Cr}_\text{test}\text{-gap})$ plane that drives the regime structure of Figure~\ref{fig:regime_map}.

\subsection{Policy Specifications}
\label{appendix:policies}

This section specifies each policy used in Table~\ref{tab:full_results} and Figures~\ref{fig:regime_map},~\ref{fig:policies_eval_main}, and~\ref{fig:cver_sweep_main}.
Each entry describes the decision rule, data dependence, and the
empirical regime in which the policy is competitive. Throughout
$b$ denotes the controller's posterior belief, $P(Y{=}1\mid\text{evidence})$,
$z^{(i)}$ the observation returned by critic $\text{Cr}_i$, and $R, c_a$ -- 
the reward and per-action cost parameters from \S\ref{sec:method}.

Policies fall into two families: (i) \textbf{unsupervised baselines}
with no learnable parameters; (ii) \textbf{Bayesian replay
controllers} (\texttt{bayesian\_greedy}, \texttt{bayesian\_DP}) that
operate over pre-collected calibration trajectories. Self-Refine and Reflexion are evaluated both as real implementations of the published methods and as policy replays over our iterative-refinement corpus.

\subsubsection{Unsupervised Baselines}
\label{appendix:policies_unfitted}

\textbf{\texttt{always\_verify}.}
Reference policy: ignore every critic, call the verifier on every
patch, return $Y$. Per-instance utility $R\cdot Y - c_{\mathrm{ver}}$
on the first patch verified; if $Y{=}0$, the controller consumes the
next pre-sampled patch from the 3-patch pool. Has no free
parameters; serves as the $\Delta_\pi{=}0$ baseline against which
every $\Delta_\pi$ in the paper is computed. Dominates regime~C
(high prior and/or cheap verification).

\textbf{\texttt{best\_of\_3}.}
Generate $N{=}3$ patches, verify each, and return the patch with the
highest $Y$. Pays $3 c_{\mathrm{gen}} + 3 c_{\mathrm{ver}}$ per
instance and never under-spends on verification. Isolates the
generator-diversity contribution: any advantage of \texttt{best\_of\_3}
over \texttt{always\_verify} is attributable to taking the maximum
over $3$ i.i.d.\ draws rather than to any orchestration logic. On the
paper's panel, it is rarely the per-cell winner.

\textbf{\texttt{gate($\text{Cr}_i$)} for $i \in \{\mathrm{syn},\mathrm{test},\mathrm{llm}\}$.}
Single-critic gate: query critic $\text{Cr}_i$ on the current patch; verify
iff $z^{(i)}{=}\text{pass}$, else regenerate (consume the next patch
in the 3-pool). Three variants span the cost/informativeness
spectrum: \texttt{gate($\text{Cr}_\text{syn}$)} uses the (essentially-free, weak)
\texttt{ast.parse} check; $\texttt{gate}(Cr_\text{test})$ uses the
mid-cost, near-oracle public-test critic;\texttt{gate($\text{Cr}_\text{llm}$)}
uses the mid-cost LLM judge. The decision rule is hand-coded:
PASS$\to$verify; FAIL$\to$regen. Across our panel,
\texttt{gate($\text{Cr}_\text{test}$)} dominates regime~B by exploiting
$\text{Cr}_\text{test}$'s near-oracle informativeness at low cost; the other two
variants are rarely competitive.

\textbf{\texttt{fixed\_pipeline}.}
Run critics in fixed order $\text{Cr}_\text{syn} \to \text{Cr}_\text{test} \to \text{Cr}_\text{llm}$ and verify iff
every critic returns PASS (logical AND-gate). Tests the naive
``stack everything'' intuition. Two structural issues hurt its
performance: (i) the AND-gate multiplies false-negative rates across
critics, missing $Y{=}1$ patches whenever any single critic fires
incorrectly; (ii) every patch pays the full critic stack even when
$\text{Cr}_\text{syn}$ or $\text{Cr}_\text{test}$ would have already settled the decision. Rarely the
per-cell winner across the panel.

\textbf{Self-Refine \citep{madaan2023self}.}
After each verify-fail, prompt the generator to critique its own
patch and produce a refined patch. The agent loops up to $K{=}4$
refinement steps; at each step, it pays $c_{\mathrm{gen}}$ for the
refinement and $c_{\mathrm{ver}}$ for the verification. Self-Refine
has no fitted parameters: the generator adapts via in-context
feedback rather than via learned weights. We report it both as an executable implementation (\S\ref{ssec:methods_compared}) and as a
policy replay over the iterative-refinement corpus. On 22 of 24 LCB cells, the Bayesian
framework dominates Self-Refine; the two exceptions (LCB-easy
gpt-5-mini and haiku-4.5) are high-$P_{\mathrm{fix}}$ cells where
mechanical retry has positive expected value.

\textbf{Reflexion \citep{shinn2023reflexion}.}
Extends Self-Refine with an explicit verbal-memory buffer of past
failures, fed back into the regeneration prompt. Same cost
structure and same lack of fitted parameters as Self-Refine. The
buffer is local to the current instance; nothing is fit from
historical data. Reflexion outperforms Self-Refine on the two
high-$P_{\mathrm{fix}}$ LCB-easy cells; otherwise, the two
heuristics track each other closely.

\subsubsection{Bayesian Controllers}
\label{appendix:policies_bayes}

Both Bayesian controllers share the same Bellman backbone
(\S\ref{sec:method}) and the same fitted Beta(1,1)-Laplace estimates:
the prior $\hat{P}(Y{=}1)$, the critic likelihoods
$\hat{P}_i(z\mid Y{=}y)$, and (for DP) the transition kernel
$\hat{P}(\text{fix}\mid\text{broken}),\hat{P}(\text{break}\mid\text{correct})$.
They differ in the lookahead horizon and in how the regenerate
branch is valued.

\textbf{\texttt{bayesian\_greedy} (BG).}
One-step Bellman lookahead. At every state $(b, t)$ the controller
computes $Q$-values for each available action and picks the argmax:
\begin{align*}
Q_{\mathrm{verify}}(b) &= R \cdot b - c_{\mathrm{ver}}, \\
Q_{\mathrm{stop}} &= 0, \\
Q_{\mathrm{gen}}(b) &= -c_{\mathrm{gen}} + \bigl(\hat{P}(Y{=}1)\cdot R - c_{\mathrm{ver}}\bigr), \\
Q_{\mathrm{crit}}^i(b) &= -c_{\mathrm{crit}}^i
+ \mathbb{E}_{z\sim \hat{P}_i(\cdot\mid b)}[Q_{\mathrm{verify}}(b_{t+1})],
\end{align*}
where $b_{t+1}$ is the Bayes posterior from Section~\ref{sec:method}. The
defining property: the regenerate branch substitutes the prior
$\hat{P}(Y{=}1)$ for the post-regen belief, so BG cannot value
multi-step refinement chains. Dominates regime~A on cells where the
critic stack is informative, but no single critic is near-oracle.

\textbf{\texttt{bayesian\_DP} (BDP).}
Full backward induction on the $(b, t)$ state space. We discretize
$b$ to a uniform $51$-point grid on $[0, 1]$ and solve the Bellman
recursion backwards over horizon $H{=}3$. The defining differences
from BG are: (i) the regenerate branch uses the \emph{measured}
transition kernel, so the post-regen belief is
$b\,(1-\hat{P}_{\mathrm{break}}) + (1-b)\,\hat{P}_{\mathrm{fix}}$
rather than the prior; (ii) critic Q-values recurse on $V_t$ rather
than truncating to $Q_{\mathrm{verify}}$, so BDP can plan
multi-step critic–refine–verify chains. At inference, the
precomputed policy $\pi_t(b)$ is an $O(1)$ lookup. BDP strictly
dominates BG only on cells where the measured
$\hat{P}_{\mathrm{fix}} \gtrsim 0.15$ makes multi-step chains
positive-EV; on the rest of the panel, BDP and BG converge
(\S\ref{ssec:dp_vs_greedy}).

\textbf{Replay convention.}
Both BG and BDP are evaluated by replaying their decision rule over
the pre-collected calibration corpus $\texttt{critic\_results.jsonl}$:
the patches and their critic verdicts are fixed (3 patches per
instance, $\texttt{seed}{=}42$); the controller selects which
observations to consult and when to verify or regenerate. The
``regenerate'' action consumes the next patch from the pool; the
controller never generates new patches in replay mode.

\begin{table}[t]
\centering
\scriptsize
\setlength{\tabcolsep}{4pt}
\renewcommand{\arraystretch}{1.05}
\begin{tabular}{ll rr}
\toprule
\textbf{Benchmark} & \textbf{Generator} & $\hat P(Y{=}1)$ & \textbf{95\% CI} \\
\midrule
\multicolumn{4}{l}{\emph{Function-level synthesis}} \\
\midrule
LCB-hard          & \texttt{gpt-5-mini}            & $0.209$ & $[0.16,\,0.26]$ \\
                  & \texttt{qwen3-coder}           & $0.096$ & $[0.06,\,0.14]$ \\
                  & \texttt{claude-haiku-4.5}      & $0.070$ & $[0.04,\,0.11]$ \\
                  & \texttt{claude-sonnet-4.5}     & $0.126$ & $[0.09,\,0.17]$ \\
                  & \texttt{Qwen2.5-Coder-32B}     & $0.104$ & $[0.07,\,0.15]$ \\
                  & \texttt{gpt-oss-20b}           & $0.387$ & $[0.33,\,0.45]$ \\
LCB-medium        & \texttt{gpt-5-mini}            & $0.283$ & $[0.24,\,0.32]$ \\
                  & \texttt{qwen3-coder}           & $0.212$ & $[0.18,\,0.25]$ \\
                  & \texttt{claude-haiku-4.5}      & $0.173$ & $[0.14,\,0.21]$ \\
                  & \texttt{claude-sonnet-4.5}     & $0.225$ & $[0.19,\,0.26]$ \\
                  & \texttt{Qwen2.5-Coder-32B}     & $0.484$ & $[0.44,\,0.53]$ \\
                  & \texttt{gpt-oss-20b}           & $0.722$ & $[0.68,\,0.76]$ \\
LCB-easy          & \texttt{gpt-5-mini}            & $0.531$ & $[0.48,\,0.59]$ \\
                  & \texttt{qwen3-coder}           & $0.348$ & $[0.30,\,0.40]$ \\
                  & \texttt{claude-haiku-4.5}      & $0.331$ & $[0.28,\,0.38]$ \\
                  & \texttt{claude-sonnet-4.5}     & $0.374$ & $[0.32,\,0.43]$ \\
                  & \texttt{Qwen2.5-Coder-32B}     & $0.862$ & $[0.82,\,0.90]$ \\
                  & \texttt{gpt-oss-20b}           & $0.957$ & $[0.93,\,0.98]$ \\
MBPP+             & \texttt{gpt-5-mini}            & $0.807$ & $[0.78,\,0.83]$ \\
                  & \texttt{qwen3-coder}           & $0.777$ & $[0.75,\,0.80]$ \\
                  & \texttt{claude-haiku-4.5}      & $0.787$ & $[0.76,\,0.81]$ \\
                  & \texttt{claude-sonnet-4.5}     & $0.819$ & $[0.79,\,0.84]$ \\
                  & \texttt{Qwen2.5-Coder-32B}     & $0.761$ & $[0.73,\,0.79]$ \\
                  & \texttt{gpt-oss-20b}           & $0.800$ & $[0.77,\,0.83]$ \\
HumanEval+        & \texttt{gpt-5-mini}            & $0.876$ & $[0.84,\,0.91]$ \\
                  & \texttt{qwen3-coder}           & $0.886$ & $[0.85,\,0.92]$ \\
                  & \texttt{claude-haiku-4.5}      & $0.908$ & $[0.88,\,0.94]$ \\
                  & \texttt{claude-sonnet-4.5}     & $0.943$ & $[0.92,\,0.96]$ \\
                  & \texttt{Qwen2.5-Coder-32B}     & $0.838$ & $[0.80,\,0.87]$ \\
                  & \texttt{gpt-oss-20b}           & $0.919$ & $[0.89,\,0.94]$ \\
\midrule
\multicolumn{4}{l}{\emph{Repository-level patch generation}} \\
\midrule
SWE-Bench Lite    & \texttt{gpt-5-mini}            & $0.251$ & $[0.22,\,0.28]$ \\
                  & \texttt{qwen3-coder}           & $0.406$ & $[0.37,\,0.44]$ \\
                  & \texttt{claude-haiku-4.5}      & $0.214$ & $[0.18,\,0.25]$ \\
                  & \texttt{claude-sonnet-4.5}     & $0.087$ & $[0.07,\,0.11]$ \\
                  & \texttt{Qwen2.5-Coder-32B}     & $0.134$ & $[0.11,\,0.16]$ \\
                  & \texttt{gpt-oss-20b}           & $0.266$ & $[0.23,\,0.30]$ \\
SWE-Bench Verified& \texttt{gpt-5-mini}            & $0.111$ & $[0.08,\,0.14]$ \\
                  & \texttt{qwen3-coder}           & $0.524$ & $[0.48,\,0.57]$ \\
                  & \texttt{claude-haiku-4.5}      & $0.438$ & $[0.39,\,0.48]$ \\
                  & \texttt{claude-sonnet-4.5}     & $0.434$ & $[0.39,\,0.48]$ \\
                  & \texttt{Qwen2.5-Coder-32B}     & $0.156$ & $[0.14,\,0.18]$ \\
                  & \texttt{gpt-oss-20b}           & $0.295$ & $[0.27,\,0.32]$ \\
\midrule
\multicolumn{4}{l}{\emph{Bug-fixing}} \\
\midrule
HumanEvalFix      & \texttt{gpt-5-mini}            & $0.854$ & $[0.82,\,0.89]$ \\
                  & \texttt{qwen3-coder}           & $0.962$ & $[0.94,\,0.98]$ \\
                  & \texttt{claude-haiku-4.5}      & $0.895$ & $[0.86,\,0.92]$ \\
                  & \texttt{claude-sonnet-4.5}     & $0.887$ & $[0.85,\,0.92]$ \\
                  & \texttt{Qwen2.5-Coder-32B}     & $0.690$ & $[0.64,\,0.74]$ \\
                  & \texttt{gpt-oss-20b}           & $0.833$ & $[0.79,\,0.87]$ \\
CodeContests      & \texttt{gpt-5-mini}            & $0.484$ & $[0.42,\,0.55]$ \\
                  & \texttt{qwen3-coder}           & $0.213$ & $[0.16,\,0.26]$ \\
                  & \texttt{claude-haiku-4.5}      & $0.213$ & $[0.16,\,0.26]$ \\
                  & \texttt{claude-sonnet-4.5}     & $0.265$ & $[0.21,\,0.32]$ \\
                  & \texttt{Qwen2.5-Coder-32B}     & $0.053$ & $[0.03,\,0.08]$ \\
                  & \texttt{gpt-oss-20b}           & $0.283$ & $[0.24,\,0.33]$ \\
\bottomrule
\end{tabular}
\caption{Per-cell single-shot prior pass rate $\hat P(Y{=}1)$
of the LLM-generated step-0 code, with Wilson 95\% confidence
intervals. The prior anchors the regime structure of
Figure~\ref{fig:regime_map}: cells with high prior favour
\texttt{always\_verify} (regime C), low priors favour the
Bayesian variants (regime A) when verification is expensive.
Priors span more than an order of magnitude across cells.}
\label{tab:prior_pass_rate}
\end{table}

\begin{table}[t]
\centering
\scriptsize
\setlength{\tabcolsep}{4pt}
\renewcommand{\arraystretch}{1.05}
\begin{tabular}{ll rrr}
\toprule
\textbf{Benchmark} & \textbf{Generator} & $\gamma_\text{syn}$ & $\gamma_\text{test}$ & $\gamma_\text{llm}$ \\
\midrule
\multicolumn{5}{l}{\emph{Function-level synthesis}} \\
\midrule
LCB-hard          & \texttt{gpt-5-mini}            & $0.395$ & $0.756$ & $0.365$ \\
                  & \texttt{qwen3-coder}           & $0.009$ & $0.865$ & $0.108$ \\
                  & \texttt{claude-haiku-4.5}      & $-0.054$ & $0.747$ & $0.129$ \\
                  & \texttt{claude-sonnet-4.5}     & $-0.018$ & $0.853$ & $-0.090$ \\
                  & \texttt{Qwen2.5-Coder-32B}     & $-0.035$ & $0.699$ & $0.289$ \\
                  & \texttt{gpt-oss-20b}           & $0.799$ & $0.869$ & $0.583$ \\
LCB-medium        & \texttt{gpt-5-mini}            & $0.180$ & $0.751$ & $0.120$ \\
                  & \texttt{qwen3-coder}           & $0.004$ & $0.928$ & $-0.087$ \\
                  & \texttt{claude-haiku-4.5}      & $-0.004$ & $0.907$ & $0.035$ \\
                  & \texttt{claude-sonnet-4.5}     & $-0.004$ & $0.865$ & $-0.097$ \\
                  & \texttt{Qwen2.5-Coder-32B}     & $-0.000$ & $0.766$ & $0.499$ \\
                  & \texttt{gpt-oss-20b}           & $0.799$ & $0.890$ & $0.578$ \\
LCB-easy          & \texttt{gpt-5-mini}            & $0.008$ & $0.987$ & $-0.011$ \\
                  & \texttt{qwen3-coder}           & $0.001$ & $0.957$ & $-0.029$ \\
                  & \texttt{claude-haiku-4.5}      & $-0.005$ & $0.980$ & $0.024$ \\
                  & \texttt{claude-sonnet-4.5}     & $-0.003$ & $0.986$ & $-0.004$ \\
                  & \texttt{Qwen2.5-Coder-32B}     & $0.019$ & $0.861$ & $0.642$ \\
                  & \texttt{gpt-oss-20b}           & $0.497$ & $0.854$ & $0.667$ \\
MBPP+             & \texttt{gpt-5-mini}            & $0.041$ & $0.074$ & $0.259$ \\
                  & \texttt{qwen3-coder}           & $0.009$ & $0.285$ & $0.310$ \\
                  & \texttt{claude-haiku-4.5}      & $0.009$ & $0.349$ & $0.283$ \\
                  & \texttt{claude-sonnet-4.5}     & $0.011$ & $0.258$ & $0.217$ \\
                  & \texttt{Qwen2.5-Coder-32B}     & $0.003$ & $0.448$ & $0.384$ \\
                  & \texttt{gpt-oss-20b}           & $0.103$ & $0.196$ & $0.297$ \\
HumanEval+        & \texttt{gpt-5-mini}            & $0.039$ & $0.720$ & $0.157$ \\
                  & \texttt{qwen3-coder}           & $0.020$ & $0.671$ & $0.112$ \\
                  & \texttt{claude-haiku-4.5}      & $0.026$ & $0.540$ & $0.188$ \\
                  & \texttt{claude-sonnet-4.5}     & $0.043$ & $0.270$ & $0.363$ \\
                  & \texttt{Qwen2.5-Coder-32B}     & $0.013$ & $0.751$ & $0.436$ \\
                  & \texttt{gpt-oss-20b}           & $0.062$ & $0.610$ & $0.250$ \\
\midrule
\multicolumn{5}{l}{\emph{Repository-level patch generation}} \\
\midrule
SWE-Bench Lite    & \texttt{gpt-5-mini}            & $0.610$ & -- & $0.504$ \\
                  & \texttt{qwen3-coder}           & $0.200$ & -- & $0.390$ \\
                  & \texttt{claude-haiku-4.5}      & $0.830$ & -- & $0.005$ \\
                  & \texttt{claude-sonnet-4.5}     & $0.599$ & -- & $0.497$ \\
                  & \texttt{Qwen2.5-Coder-32B}     & $0.148$ & -- & $0.475$ \\
                  & \texttt{gpt-oss-20b}           & $0.330$ & -- & $0.004$ \\
SWE-Bench Verified& \texttt{gpt-5-mini}            & $0.834$ & -- & $0.807$ \\
                  & \texttt{qwen3-coder}           & $0.209$ & -- & $0.419$ \\
                  & \texttt{claude-haiku-4.5}      & $0.370$ & -- & $0.381$ \\
                  & \texttt{claude-sonnet-4.5}     & $0.092$ & -- & $0.273$ \\
                  & \texttt{Qwen2.5-Coder-32B}     & $0.139$ & -- & $0.466$ \\
                  & \texttt{gpt-oss-20b}           & $0.319$ & -- & $0.447$ \\
\midrule
\multicolumn{5}{l}{\emph{Bug-fixing}} \\
\midrule
HumanEvalFix      & \texttt{gpt-5-mini}            & $0.015$ & $0.979$ & $0.969$ \\
                  & \texttt{qwen3-coder}           & $0.064$ & $0.797$ & $0.789$ \\
                  & \texttt{claude-haiku-4.5}      & $0.222$ & $0.947$ & $0.938$ \\
                  & \texttt{claude-sonnet-4.5}     & $0.276$ & $0.974$ & $0.965$ \\
                  & \texttt{Qwen2.5-Coder-32B}     & $0.005$ & $0.848$ & $0.708$ \\
                  & \texttt{gpt-oss-20b}           & $0.156$ & $0.362$ & $0.763$ \\
CodeContests      & \texttt{gpt-5-mini}            & $0.075$ & $0.666$ & $0.878$ \\
                  & \texttt{qwen3-coder}           & $0.002$ & $0.818$ & $0.897$ \\
                  & \texttt{claude-haiku-4.5}      & $0.037$ & $0.882$ & $0.947$ \\
                  & \texttt{claude-sonnet-4.5}     & $0.013$ & $0.863$ & $0.926$ \\
                  & \texttt{Qwen2.5-Coder-32B}     & $0.248$ & $0.815$ & $0.336$ \\
                  & \texttt{gpt-oss-20b}           & $0.879$ & $0.965$ & $0.329$ \\
\bottomrule
\end{tabular}
\caption{Per-cell critic informativeness gap $\gamma_\ell =
P(z_\ell{=}\text{pass}\mid Y{=}1) - P(z_\ell{=}\text{pass}\mid Y{=}0)$
estimated with Beta(1,1) smoothing on the calibration cohort.
$\gamma_\ell{\to}1$ corresponds to a near-oracle critic;
$\gamma_\ell{\to}0$ corresponds to a critic that carries no
Bayes signal. \texttt{Cr\_lint} is omitted -- it is a no-op
pass-through in this codebase so its gap is structurally
zero on every cell. \texttt{Cr\_test} is shown as ``--'' on
all SWE-Bench cells: SWE-Bench has no public-test split
distinct from the oracle verifier (the \texttt{test\_patch}
is the verifier), so there is no independent public-test
critic to estimate a gap for. The gap structure here drives
the regime-B vs regime-A boundary in
Figure~\ref{fig:regime_map}: cells with
$\gamma_\text{test}{\to}1$ favour single-critic gating,
cells with moderate $\gamma$ across multiple critics favour
Bayesian composition.}
\label{tab:critic_gap}
\end{table}

\section{Detailed Results}
\label{sec:full_results}

Table~\ref{tab:full_results} reports the per-cell utility
difference $\Delta_\pi$ against \texttt{always\_verify} for every
(generator, benchmark, policy) triple in the empirical panel, under
the ``slow oracle, fast critics'' cost vector; bug-fixing rows use the
fast-oracle variant noted in the caption. Bold marks the per-cell
winner among the reported methods. Columns \textbf{BG} and
\textbf{BDP} denote \texttt{bayesian\_greedy} and
\texttt{bayesian\_DP}.

\begin{table*}[t]
\centering
\scriptsize
\setlength{\tabcolsep}{3pt}
\renewcommand{\arraystretch}{1.05}
\begin{tabular}{ll rrrrrrrrr}
\toprule
\textbf{Benchmark} & \textbf{Generator}
 & \textbf{BoN}
 & \textbf{$t_{C_\text{syn}}$}
 & \textbf{$t_{C_\text{test}}$}
 & \textbf{$t_{C_\text{llm}}$}
 & \textbf{FP}
 & \textbf{SR}
 & \textbf{Rfx}
 & \textbf{BG}
 & \textbf{BDP} \\
\midrule
\multicolumn{11}{l}{\emph{Function-level synthesis}} \\
\midrule
LCB-hard          & \texttt{gpt-5-mini}            & $-20.0$ & $+3.1$ & $\mathbf{+4.3}$ & $-6.0$ & $-7.5$ & $-33.2$ & $-4.4$ & $-7.6$ & $+0.3$ \\
                  & \texttt{qwen3-coder}           & $-20.0$ & $-1.4$ & $-12.8$ & $-10.0$ & $-12.1$ & $-25.6$ & $-15.4$ & $\mathbf{+2.4}$ & $\mathbf{+2.4}$ \\
                  & \texttt{claude-haiku-4.5}      & $-20.0$ & $-1.0$ & $-13.8$ & $-14.3$ & $-16.4$ & $-10.8$ & $-29.3$ & $\mathbf{+2.8}$ & $+2.1$ \\
                  & \texttt{claude-sonnet-4.5}     & $-23.8$ & $-1.0$ & $-8.7$ & $-16.1$ & $-18.3$ & $-5.4$ & $-5.2$ & $\mathbf{+2.5}$ & $\mathbf{+2.5}$ \\
                  & \texttt{Qwen2.5-Coder-32B}     & $-31.5$ & $\mathbf{-1.0}$ & $-9.7$ & $-26.2$ & $-28.8$ & $-17.7$ & $-15.3$ & $-2.6$ & $-2.6$ \\
                  & \texttt{gpt-oss-20b}           & $-24.0$ & $+0.0$ & $+5.1$ & $-1.1$ & $-3.0$ & $-34.4$ & $\mathbf{+25.8}$ & $-17.0$ & $+4.6$ \\
LCB-medium        & \texttt{gpt-5-mini}            & $-11.3$ & $-1.2$ & $+0.7$ & $+1.7$ & $+0.3$ & $-6.5$ & $-14.1$ & $-15.7$ & $\mathbf{+1.9}$ \\
                  & \texttt{qwen3-coder}           & $-18.1$ & $-1.0$ & $-11.8$ & $-6.5$ & $-8.3$ & $-15.0$ & $-23.2$ & $-16.9$ & $\mathbf{+2.2}$ \\
                  & \texttt{claude-haiku-4.5}      & $-18.1$ & $-1.0$ & $-13.1$ & $-3.3$ & $-4.8$ & $-10.4$ & $-17.9$ & $-18.8$ & $\mathbf{+2.6}$ \\
                  & \texttt{claude-sonnet-4.5}     & $-18.1$ & $-1.3$ & $-10.0$ & $-10.2$ & $-11.8$ & $+6.0$ & $\mathbf{+10.6}$ & $-17.3$ & $+2.3$ \\
                  & \texttt{Qwen2.5-Coder-32B}     & $-21.9$ & $-1.0$ & $-4.3$ & $-12.1$ & $-13.9$ & $-16.0$ & $-16.4$ & $\mathbf{+0.0}$ & $\mathbf{+0.0}$ \\
                  & \texttt{gpt-oss-20b}           & $-23.8$ & $+2.6$ & $+2.4$ & $-7.5$ & $-9.2$ & $-35.0$ & $\mathbf{+6.1}$ & $-7.2$ & $+2.3$ \\
LCB-easy          & \texttt{gpt-5-mini}            & $-25.9$ & $-1.0$ & $\mathbf{+10.8}$ & $-5.3$ & $-6.4$ & $-11.2$ & $+4.3$ & $-9.1$ & $+0.6$ \\
                  & \texttt{qwen3-coder}           & $-20.0$ & $-1.0$ & $-11.0$ & $-1.3$ & $-2.4$ & $-10.6$ & $-20.7$ & $-15.7$ & $\mathbf{+1.4}$ \\
                  & \texttt{claude-haiku-4.5}      & $-20.0$ & $-1.0$ & $-11.0$ & $-2.6$ & $-3.8$ & $-10.9$ & $-0.1$ & $-15.7$ & $\mathbf{+1.9}$ \\
                  & \texttt{claude-sonnet-4.5}     & $-17.1$ & $-1.0$ & $-8.2$ & $-2.5$ & $-3.6$ & $-11.5$ & $+0.5$ & $-15.7$ & $\mathbf{+1.9}$ \\
                  & \texttt{Qwen2.5-Coder-32B}     & $-20.0$ & $-1.0$ & $-1.0$ & $-5.6$ & $-6.8$ & $-17.6$ & $-1.0$ & $\mathbf{+0.0}$ & $-1.0$ \\
                  & \texttt{gpt-oss-20b}           & $-20.0$ & $-1.0$ & $\mathbf{+4.2}$ & $+1.0$ & $-0.1$ & $-22.9$ & $-3.6$ & $+0.0$ & $+0.0$ \\
MBPP+             & \texttt{gpt-5-mini}            & $-21.1$ & $-0.0$ & $\mathbf{+1.5}$ & $-8.6$ & $-9.9$ & $-21.3$ & $-46.0$ & $+0.0$ & $+0.0$ \\
                  & \texttt{qwen3-coder}           & $-30.7$ & $-1.0$ & $-3.7$ & $-16.1$ & $-17.5$ & $-28.8$ & $-7.8$ & $\mathbf{+0.0}$ & $\mathbf{+0.0}$ \\
                  & \texttt{claude-haiku-4.5}      & $-20.0$ & $-1.0$ & $-1.8$ & $-8.5$ & $-9.8$ & $-9.0$ & $\mathbf{+2.2}$ & $+0.0$ & $+0.0$ \\
                  & \texttt{claude-sonnet-4.5}     & $-21.1$ & $-1.0$ & $-1.4$ & $-13.6$ & $-14.9$ & $-16.1$ & $-3.0$ & $\mathbf{+0.0}$ & $\mathbf{+0.0}$ \\
                  & \texttt{Qwen2.5-Coder-32B}     & $-22.1$ & $-1.0$ & $-3.3$ & $-16.9$ & $-18.4$ & $-35.7$ & $-125.9$ & $\mathbf{+0.0}$ & $\mathbf{+0.0}$ \\
                  & \texttt{gpt-oss-20b}           & $-18.9$ & $-1.0$ & $-1.0$ & $-6.3$ & $-7.7$ & $-34.3$ & $-125.5$ & $\mathbf{+0.0}$ & $\mathbf{+0.0}$ \\
HumanEval+        & \texttt{gpt-5-mini}            & $-17.6$ & $-1.0$ & $\mathbf{+3.3}$ & $-21.0$ & $-22.3$ & $-32.0$ & $-80.3$ & $+0.0$ & $+0.0$ \\
                  & \texttt{qwen3-coder}           & $-17.4$ & $-1.0$ & $\mathbf{+3.0}$ & $-20.1$ & $-21.5$ & $-9.2$ & $-12.5$ & $+0.0$ & $+0.0$ \\
                  & \texttt{claude-haiku-4.5}      & $-22.4$ & $-1.0$ & $-1.0$ & $-20.8$ & $-22.1$ & $-11.2$ & $-4.5$ & $\mathbf{+0.0}$ & $\mathbf{+0.0}$ \\
                  & \texttt{claude-sonnet-4.5}     & $-22.4$ & $-1.0$ & $-1.0$ & $-21.0$ & $-22.3$ & $-12.0$ & $-1.0$ & $\mathbf{+0.0}$ & $\mathbf{+0.0}$ \\
                  & \texttt{Qwen2.5-Coder-32B}     & $-17.6$ & $-1.0$ & $\mathbf{+3.6}$ & $-16.1$ & $-17.7$ & $-26.1$ & $-119.8$ & $+0.0$ & $+0.0$ \\
                  & \texttt{gpt-oss-20b}           & $-17.6$ & $-1.0$ & $\mathbf{+3.3}$ & $-10.6$ & $-11.9$ & $-48.8$ & $-82.5$ & $+0.0$ & $+0.0$ \\
\midrule
\multicolumn{11}{l}{\emph{Repository-level patch generation}} \\
\midrule
SWE-Bench Lite    & \texttt{gpt-5-mini}            & $-17.3$ & $+21.7$ & -- & $+13.7$ & $+11.6$ & $-24.4$ & $-102.7$ & $\mathbf{+62.0}$ & $\mathbf{+62.0}$ \\
                  & \texttt{qwen3-coder}           & $-20.0$ & $+1.7$ & -- & $+9.0$ & $+7.1$ & $-57.7$ & $-136.0$ & $\mathbf{+43.3}$ & $\mathbf{+43.3}$ \\
                  & \texttt{claude-haiku-4.5}      & $-24.0$ & $+39.7$ & -- & $+32.3$ & $+29.3$ & $-12.9$ & $-20.0$ & $+67.3$ & $\mathbf{+67.6}$ \\
                  & \texttt{claude-sonnet-4.5}     & $-50.7$ & $+28.9$ & -- & $+22.3$ & $+19.9$ & $-24.8$ & $-93.3$ & $\mathbf{+59.3}$ & $\mathbf{+59.3}$ \\
                  & \texttt{Qwen2.5-Coder-32B}     & $-17.3$ & $+6.1$ & -- & $+27.9$ & $+25.5$ & $-16.0$ & $-268.0$ & $\mathbf{+74.0}$ & $\mathbf{+74.0}$ \\
                  & \texttt{gpt-oss-20b}           & $-12.0$ & $+8.3$ & -- & $+32.3$ & $+29.3$ & $-52.0$ & $-256.0$ & $\mathbf{+67.3}$ & $\mathbf{+67.3}$ \\
SWE-Bench Verified& \texttt{gpt-5-mini}            & $-24.0$ & $+41.9$ & -- & $+34.7$ & $+32.3$ & $-55.0$ & $-226.0$ & $\mathbf{+74.0}$ & $\mathbf{+74.0}$ \\
                  & \texttt{qwen3-coder}           & $-16.0$ & $+2.8$ & -- & $+5.6$ & $+3.8$ & $-42.6$ & $-116.0$ & $\mathbf{+52.0}$ & $\mathbf{+52.0}$ \\
                  & \texttt{claude-haiku-4.5}      & $-18.0$ & $+13.9$ & -- & $+7.2$ & $+5.0$ & $-14.6$ & $-166.0$ & $\mathbf{+56.0}$ & $\mathbf{+56.0}$ \\
                  & \texttt{claude-sonnet-4.5}     & $-20.0$ & $+1.7$ & -- & $+2.4$ & $+0.8$ & $-27.4$ & $-122.0$ & $\mathbf{+66.0}$ & $\mathbf{+66.0}$ \\
                  & \texttt{Qwen2.5-Coder-32B}     & $-19.2$ & $+8.6$ & -- & $+27.4$ & $+25.2$ & $-14.6$ & $-266.4$ & $\mathbf{+74.8}$ & $\mathbf{+74.8}$ \\
                  & \texttt{gpt-oss-20b}           & $-8.8$ & $+12.5$ & -- & $+14.5$ & $+12.7$ & $-47.0$ & $-180.0$ & $\mathbf{+68.4}$ & $\mathbf{+68.4}$ \\
\midrule
\multicolumn{11}{l}{\emph{Bug-fixing}} \\
\midrule
HumanEvalFix      & \texttt{gpt-5-mini}            & $-22.4$ & $-1.0$ & $+0.3$ & $+0.3$ & $-0.8$ & $-32.7$ & $\mathbf{+5.2}$ & $+0.0$ & $+0.1$ \\
                  & \texttt{qwen3-coder}           & $-20.0$ & $-1.0$ & $-1.4$ & $-1.8$ & $-2.9$ & $-37.6$ & $\mathbf{+1.2}$ & $+0.0$ & $+0.0$ \\
                  & \texttt{claude-haiku-4.5}      & $-21.8$ & $-0.6$ & $+3.8$ & $+3.0$ & $+1.9$ & $-35.1$ & $+1.0$ & $-13.3$ & $-3.8$ \\
                  & \texttt{claude-sonnet-4.5}     & $-16.3$ & $+1.3$ & $+5.9$ & $+5.2$ & $+3.9$ & $-12.9$ & $+4.2$ & $-17.6$ & $-2.4$ \\
                  & \texttt{Qwen2.5-Coder-32B}     & $-22.4$ & $\mathbf{+0.9}$ & $-2.7$ & $-30.5$ & $-32.4$ & $-30.7$ & $-102.8$ & $+0.0$ & $+0.0$ \\
                  & \texttt{gpt-oss-20b}           & $-22.5$ & $-1.0$ & $-1.9$ & $-5.3$ & $-6.8$ & $-27.5$ & $-103.9$ & $\mathbf{+0.0}$ & $\mathbf{+0.0}$ \\
CodeContests      & \texttt{gpt-5-mini}            & $-15.5$ & $\mathbf{-1.0}$ & $-7.7$ & $-4.9$ & $-6.9$ & $-49.1$ & $-7.9$ & $-13.5$ & $-4.3$ \\
                  & \texttt{qwen3-coder}           & $-41.4$ & $-1.0$ & $-9.9$ & $-11.3$ & $-13.5$ & $-40.0$ & $-15.1$ & $-16.2$ & $\mathbf{+2.0}$ \\
                  & \texttt{claude-haiku-4.5}      & $-25.3$ & $+1.1$ & $-7.8$ & $-9.6$ & $-12.1$ & $-12.9$ & $-11.5$ & $-19.4$ & $\mathbf{+2.7}$ \\
                  & \texttt{claude-sonnet-4.5}     & $-25.3$ & $-1.0$ & $-12.4$ & $-11.9$ & $-14.4$ & $-22.4$ & $-30.6$ & $-17.4$ & $\mathbf{+2.3}$ \\
                  & \texttt{Qwen2.5-Coder-32B}     & $-22.4$ & $-5.7$ & $-7.8$ & $-21.8$ & $-24.6$ & $-18.0$ & $-47.2$ & $\mathbf{+3.3}$ & $\mathbf{+3.3}$ \\
                  & \texttt{gpt-oss-20b}           & $-13.8$ & $+7.7$ & $\mathbf{+12.5}$ & $-28.2$ & $-30.2$ & $-19.4$ & $+8.2$ & $+0.0$ & $+11.7$ \\
\bottomrule
\end{tabular}
\caption{Per-cell utility difference $\Delta_\pi = \bar U_\pi -
\bar U_{\texttt{always\_verify}}$ across all generators, benchmarks,
and nine non-reference methods. FP denotes \texttt{fixed\_pipeline}.
All values computed on the held-out 25\% eval split with the anchored
cal $\cap$ SR $\cap$ Rfx pool
(\textsc{stat\_anchor}). The reference column \texttt{always\_verify}
is omitted (all entries $\Delta_\pi=0$ by definition). Bold marks the
per-cell winner across the nine method columns. ``--'' indicates a
cell with no data available yet.
$t_{C_\text{test}}$ is omitted on SWE-Bench cells (Repository-level patch
generation): SWE-Bench has no public-test split distinct from the oracle
verifier, so the public-test critic is undefined there.}
\label{tab:full_results}
\end{table*}

\section{Bayesian Belief States for Uncertainty Quantification of Coding Agents}
\label{sec:uq_results}

In this experiment, we explore whether the Bayesian belief states can quantify uncertainty for an externally controlled agentic coding trajectory before the final oracle verification outcome is observed. We use SAGE agent~\cite{suri2025structured} using \texttt{gpt-oss-20b} as the controller in this experiment. We run the agent on LCB-Medium and LCB-Hard, use the training split to estimate the initial prior and critic likelihoods, and evaluate uncertainty scores for the final candidate on the held-out test split.

We experiment with four uncertainty quantification methods. First, we use two standard methods: \textit{perplexity} and \textit{sequence probability}~\cite{fomicheva-etal-2020-unsupervised}, which  are computed from the token log-probabilities of the final generated code. Second, the \textit{Tool Success Rate} is the empirical success rate of non-final critic and verifier calls observed before any successful verifier call. The terminal verifier decision itself is excluded to avoid label leakage. When no such tool evidence is available, we set the score to the train-set prior. Finally, \textit{Bayes Belief State} corresponds to a posterior probability of correctness, initialized from the train-set prior and updated using fitted critic likelihoods and generation transition dynamics. As with the tool success rate, the belief state is measured before the terminal correctness label is revealed.

Table~\ref{tab:lcb_uncertainty} shows the Prediction Rejection Ratio (PRR)~\cite{malinin2020uncertainty,vashurin-etal-2025-benchmarking}, which measures the area under the rejection curve. This curve plots the average quality of the remaining generations as we progressively abstain from an increasing fraction of the most uncertain predictions. Following standard practice~\cite{vashurin-etal-2025-benchmarking}, we compute PRR up to a rejection rate of 50\%.

The Bayesian belief state yields the best average results, outperforming the second-best method by 6.5\% on average and is particularly strong on LCB-Hard. This suggests that aggregating multiple tool observations using a Bayesian model is more informative than treating all tool outcomes uniformly. The tool success rate performs best on LCB-Medium, where many successful episodes are solved directly and the train-set prior is already high. Sequence probability is also competitive, indicating that the model's own generation probability contains substantial information about final correctness. In contrast, its length-normalized variant, perplexity, is less reliable, especially on LCB-Medium, where normalization discards much of the length-sensitive signal captured by sequence probability.

Overall, this result demonstrates that aggregating multiple observations from critics and verifiers using a Bayesian belief state provides a highly effective uncertainty measure for arbitrary external agents, showing strong performance in identifying erroneous generations.

\section{Additional Results}
\label{appendix:figures}

This appendix collects the per-benchmark, per-strategy, per-critic, and per-(benchmark, generator) slices of the regime map that don't
fit in the 8-page main paper. All figures are reproduced from the
same single-shot calibration + iterative-refinement corpus described
in \S\ref{sec:experiments}, covering the 9-benchmark $\times$
6-generator panel. Unless otherwise noted, figures use the
per-benchmark cost vectors: \emph{fast-oracle}
($C_{\mathrm{ver}}{=}5$,
$C_{\mathrm{crit}}^{\mathrm{syn}}{=}C_{\mathrm{crit}}^{\mathrm{test}}{=}C_{\mathrm{crit}}^{\mathrm{llm}}{=}1$,
$C_{\mathrm{gen}}{=}10$, $R{=}100$) on the function-level synthesis
and bug-fixing benchmarks; \emph{slow-oracle}
($C_{\mathrm{ver}}{=}90$, $C_{\mathrm{crit}}^{\mathrm{syn}}{=}1$,
$C_{\mathrm{crit}}^{\mathrm{test}}{=}2$,
$C_{\mathrm{crit}}^{\mathrm{llm}}{=}5$, $C_{\mathrm{gen}}{=}10$,
$R{=}100$) on the SWE-Bench rows.

\FloatBarrier
\clearpage

\subsection{Held-out Policy Comparison Across Benchmark--Generator Pairs}
\label{appendix:policies_per_model}

\begin{figure*}[!htb]
  \centering
  \includegraphics[width=\linewidth]{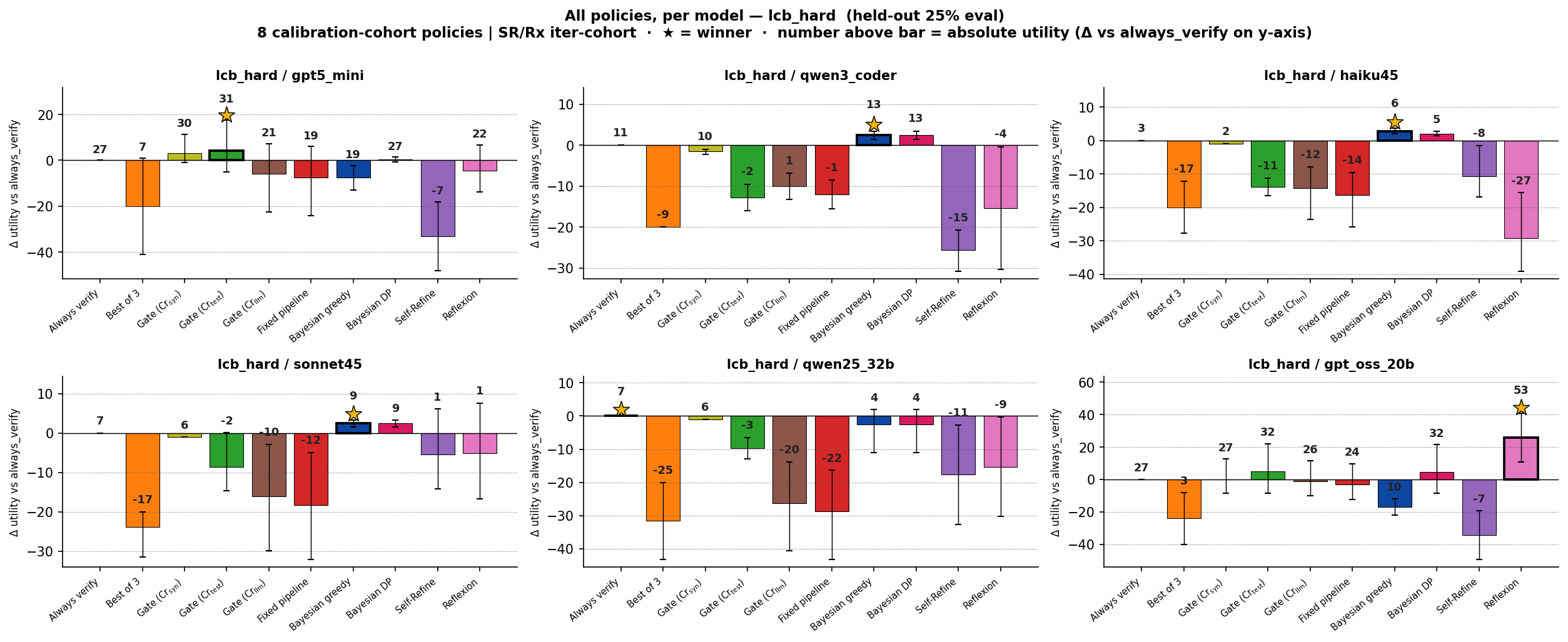}
  \caption{Held-out eval policy comparison on LCB-hard, all
  generators on one panel. Bayesian variants dominate this regime-A
  cell across every generator. Companion to
  Figure~\ref{fig:policies_eval_main} (SWE-Bench Lite /
  \texttt{claude-haiku-4.5} cell shown in the main text).}
  \label{fig:policies_eval_lcb_hard}
\end{figure*}

\begin{figure*}[!htb]
  \centering
  \includegraphics[width=\linewidth]{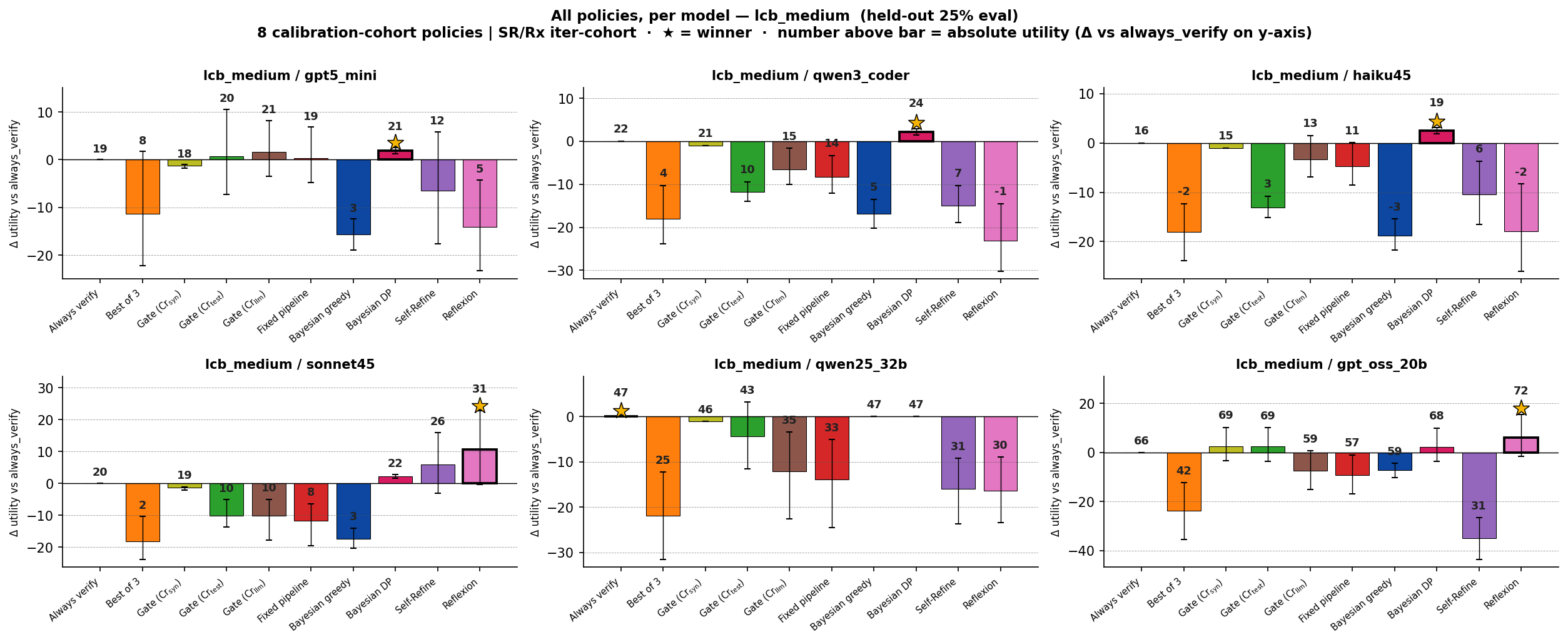}
  \caption{Held-out eval policy comparison on LCB-medium, all
  generators on one panel.}
  \label{fig:policies_eval_lcb_medium}
\end{figure*}

\begin{figure*}[!htb]
  \centering
  \includegraphics[width=\linewidth]{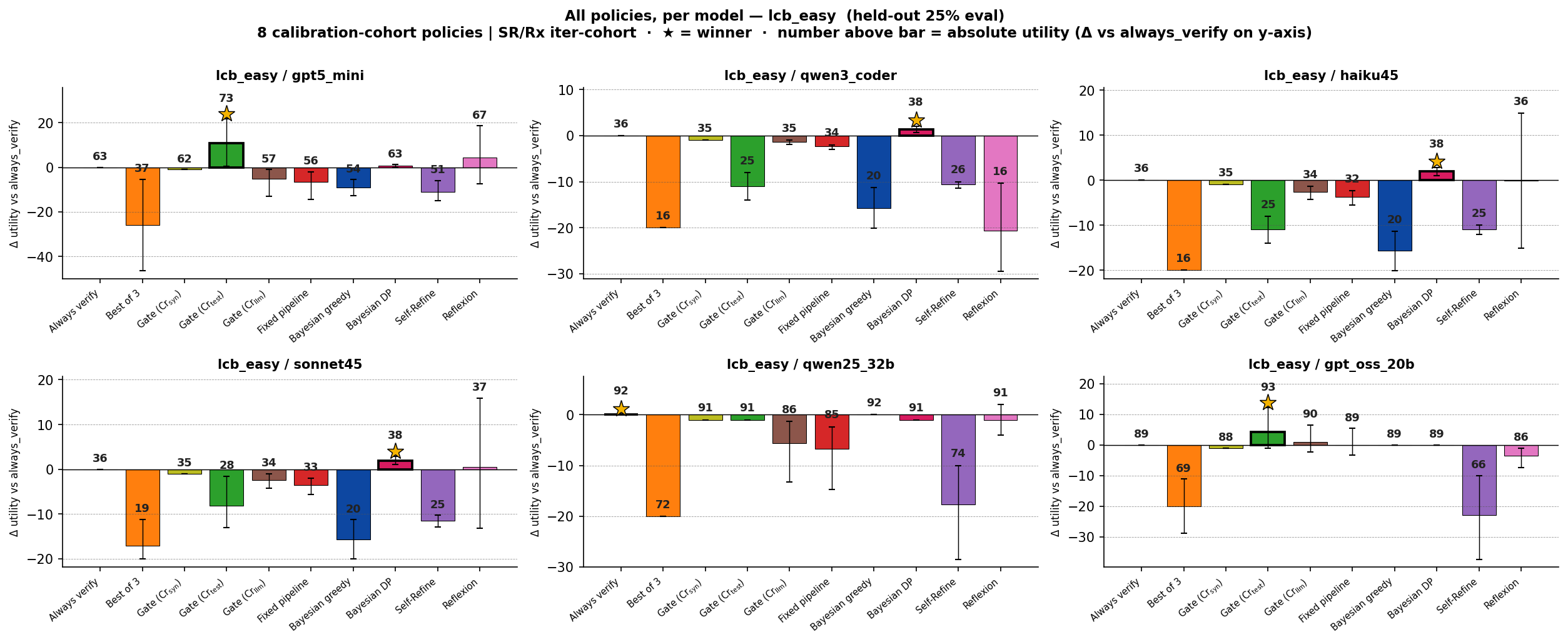}
  \caption{Held-out eval policy comparison on LCB-easy, all
  generators on one panel. \texttt{gate($\text{Cr}_\text{test}$)} wins
  on most cells here -- LCB-easy's near-oracle public-test critic
  places these cells in regime~B.}
  \label{fig:policies_eval_lcb_easy}
\end{figure*}

\begin{figure*}[!htb]
  \centering
  \includegraphics[width=\linewidth]{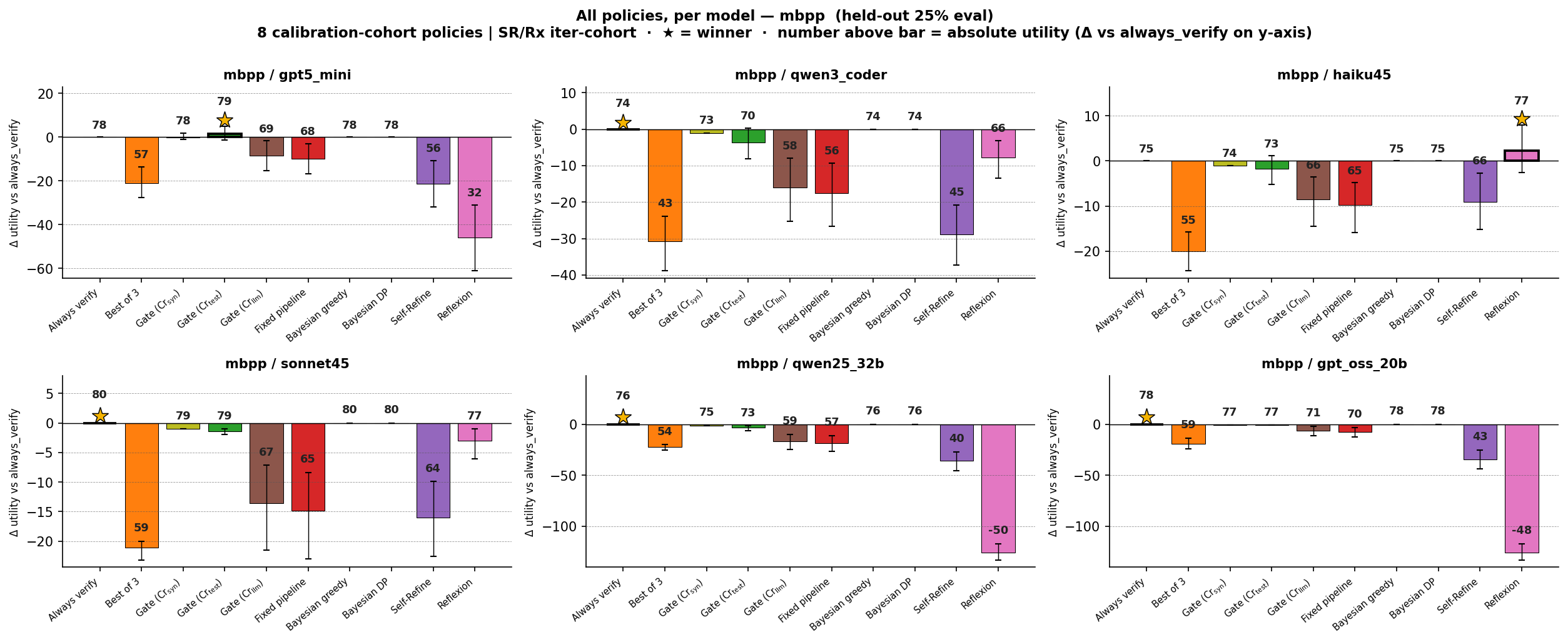}
  \caption{Held-out eval policy comparison on MBPP+, all generators
  on one panel.}
  \label{fig:policies_eval_mbpp}
\end{figure*}

\begin{figure*}[!htb]
  \centering
  \includegraphics[width=\linewidth]{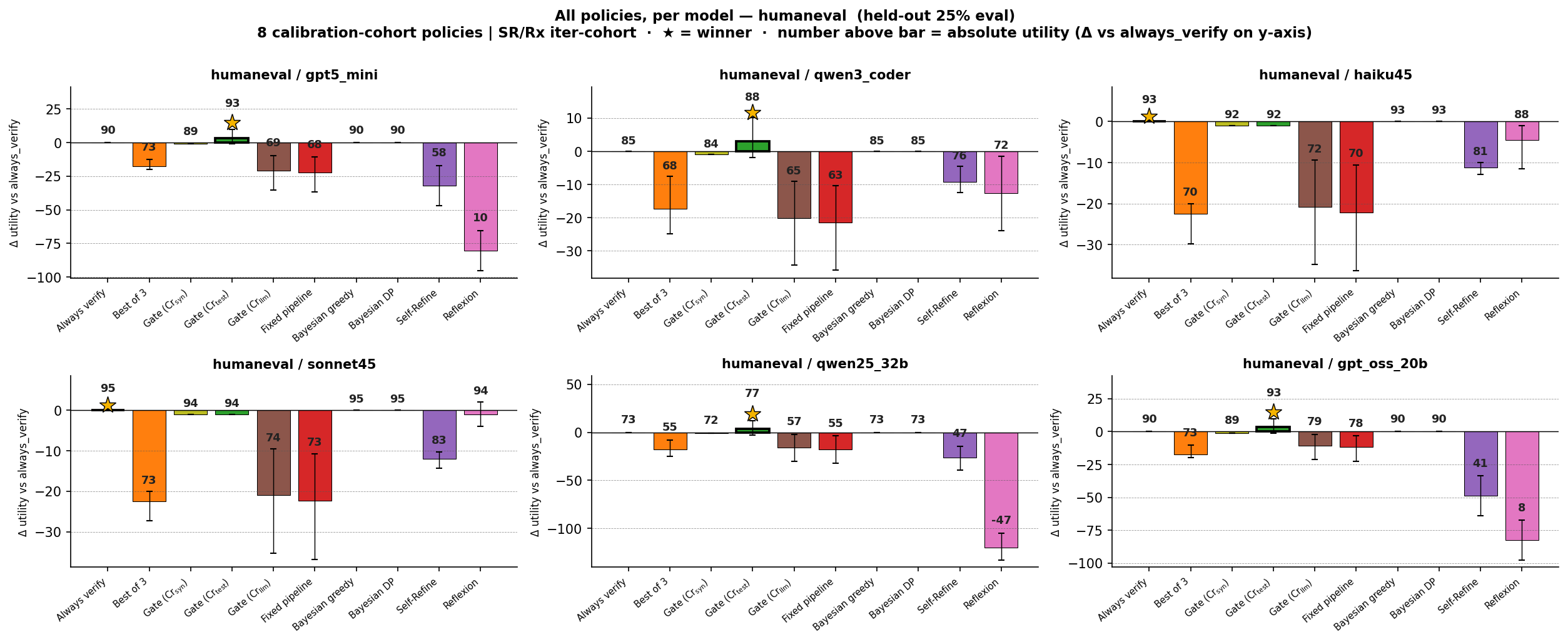}
  \caption{Held-out eval policy comparison on HumanEval+, all
  generators on one panel.}
  \label{fig:policies_eval_humaneval}
\end{figure*}

\begin{figure*}[!htb]
  \centering
  \includegraphics[width=\linewidth]{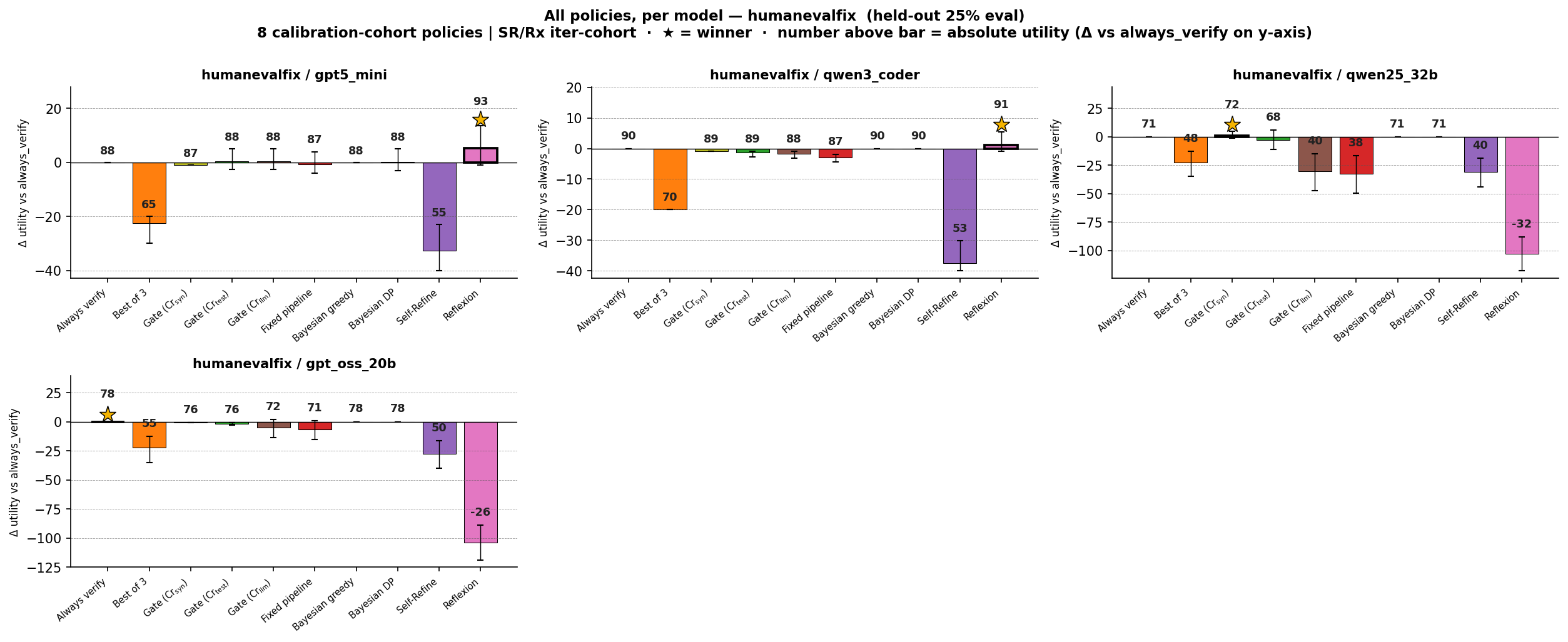}
  \caption{Held-out eval policy comparison on HumanEvalFix, all generators
  on one panel.}
  \label{fig:policies_eval_humanevalfix}
\end{figure*}

\begin{figure*}[!htb]
  \centering
  \includegraphics[width=\linewidth]{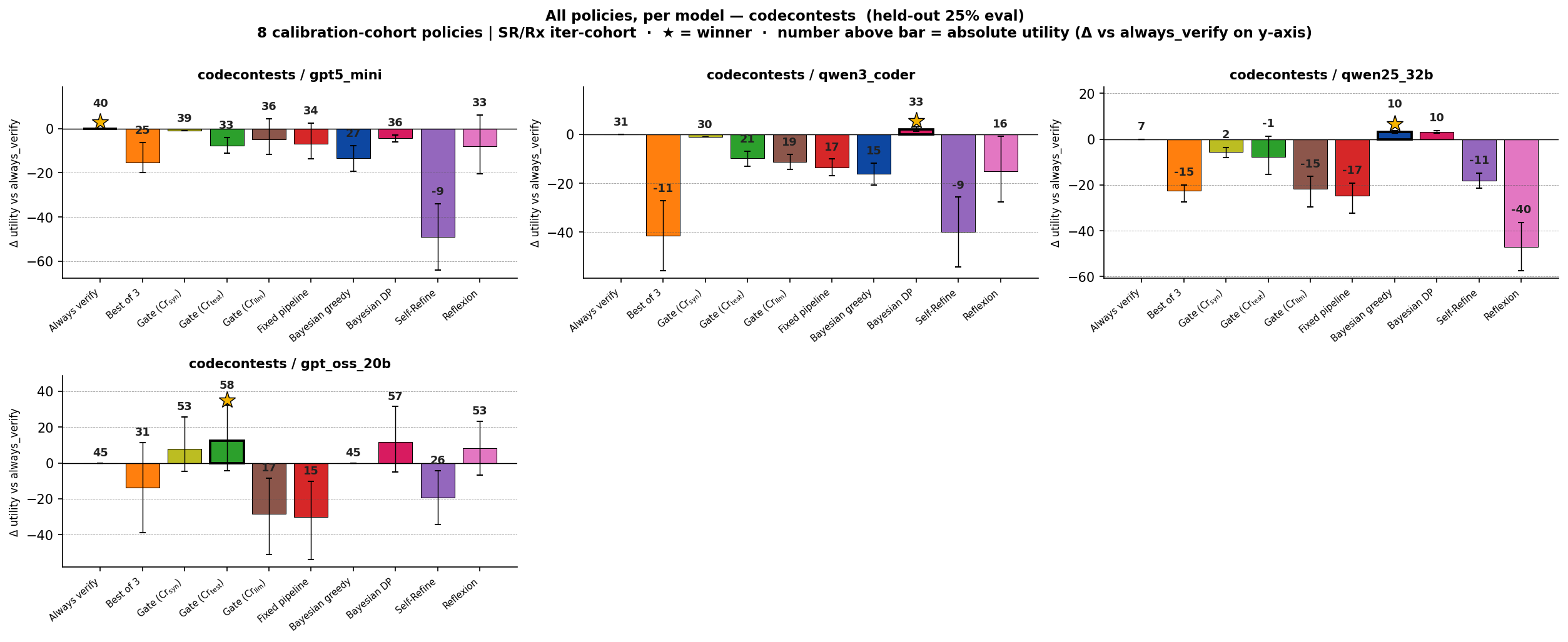}
  \caption{Held-out eval policy comparison on CodeContests, all generators
  on one panel.}
  \label{fig:policies_eval_codecontests}
\end{figure*}

\begin{figure*}[!htb]
  \centering
  \includegraphics[width=\linewidth]{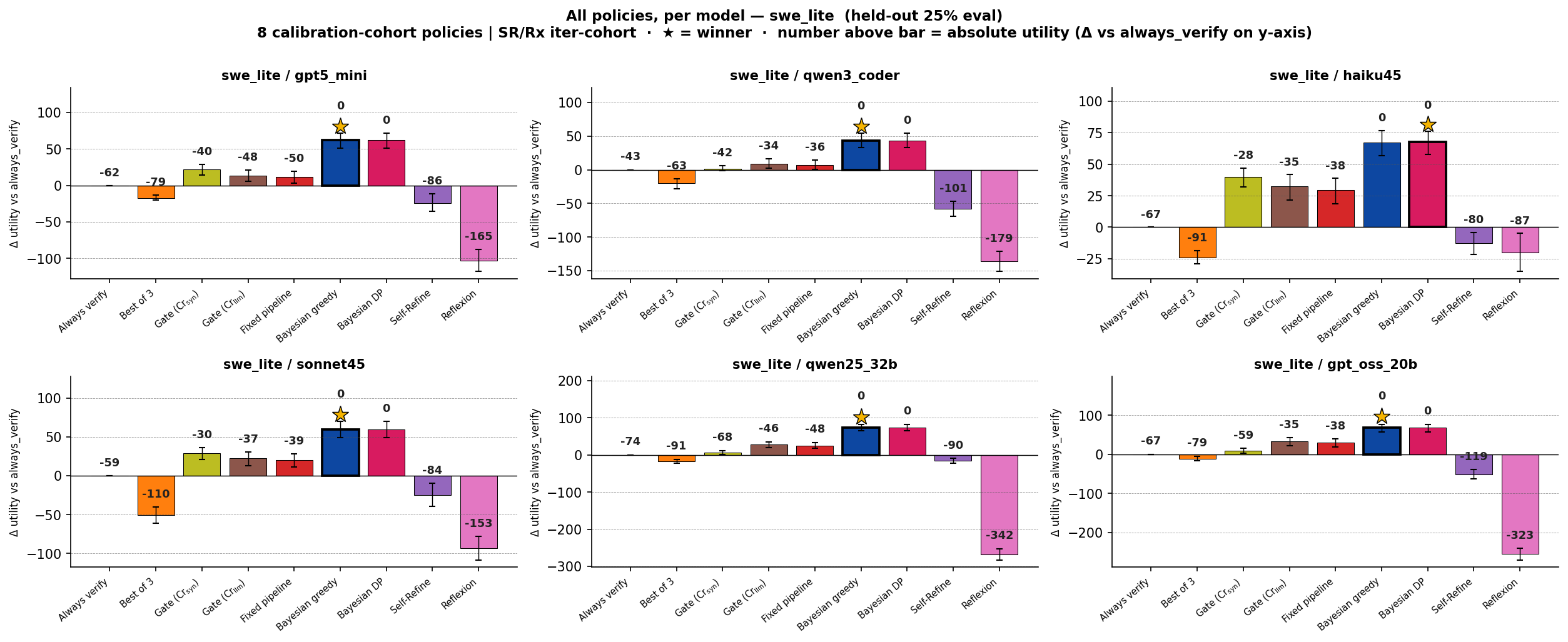}
  \caption{Held-out eval policy comparison on SWE-Bench Lite, all generators on one panel (the \texttt{claude-haiku-4.5} cell of this figure is the headline shown as Figure~\ref{fig:policies_eval_main} in the main text).}
  \label{fig:policies_eval_swe_lite}
\end{figure*}

\begin{figure*}[!htb]
  \centering
  \includegraphics[width=\linewidth]{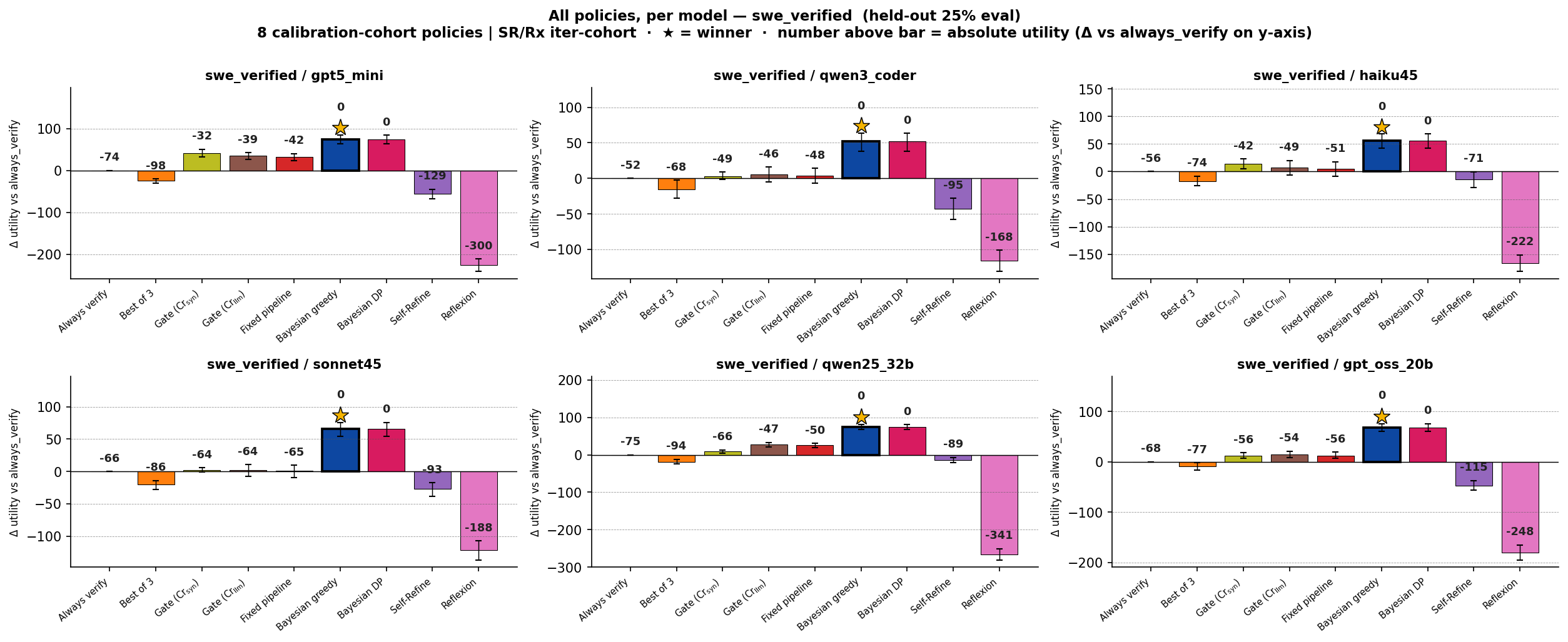}
  \caption{Held-out eval policy comparison on SWE-Bench Verified,
  all generators on one panel.}
  \label{fig:policies_eval_swe_verified}
\end{figure*}

Figures~\ref{fig:policies_eval_lcb_hard}--\ref{fig:policies_eval_swe_verified} expand the headline
Figure~\ref{fig:policies_eval_main} to every (benchmark, generator)
pair on the held-out 25\% eval split. One bar chart per cell; the
gold-starred bar marks the winner. The Bayesian variants win or tie
on the regime-A cells; \texttt{gate($\text{Cr}_\text{test}$)} wins on
regime-B; \texttt{always\_verify} wins on the saturated cells.
Self-Refine and Reflexion are evaluated as actual agents (not policy
replays) and consistently underperform on cells with weak refine
signals.

\FloatBarrier
\clearpage

\subsection{Train-vs-Evaluation Policy Comparison Across Benchmark--Generator Pairs}
\label{appendix:train_vs_eval}

\begin{figure*}[!htb]
  \centering
  \includegraphics[width=0.85\linewidth]{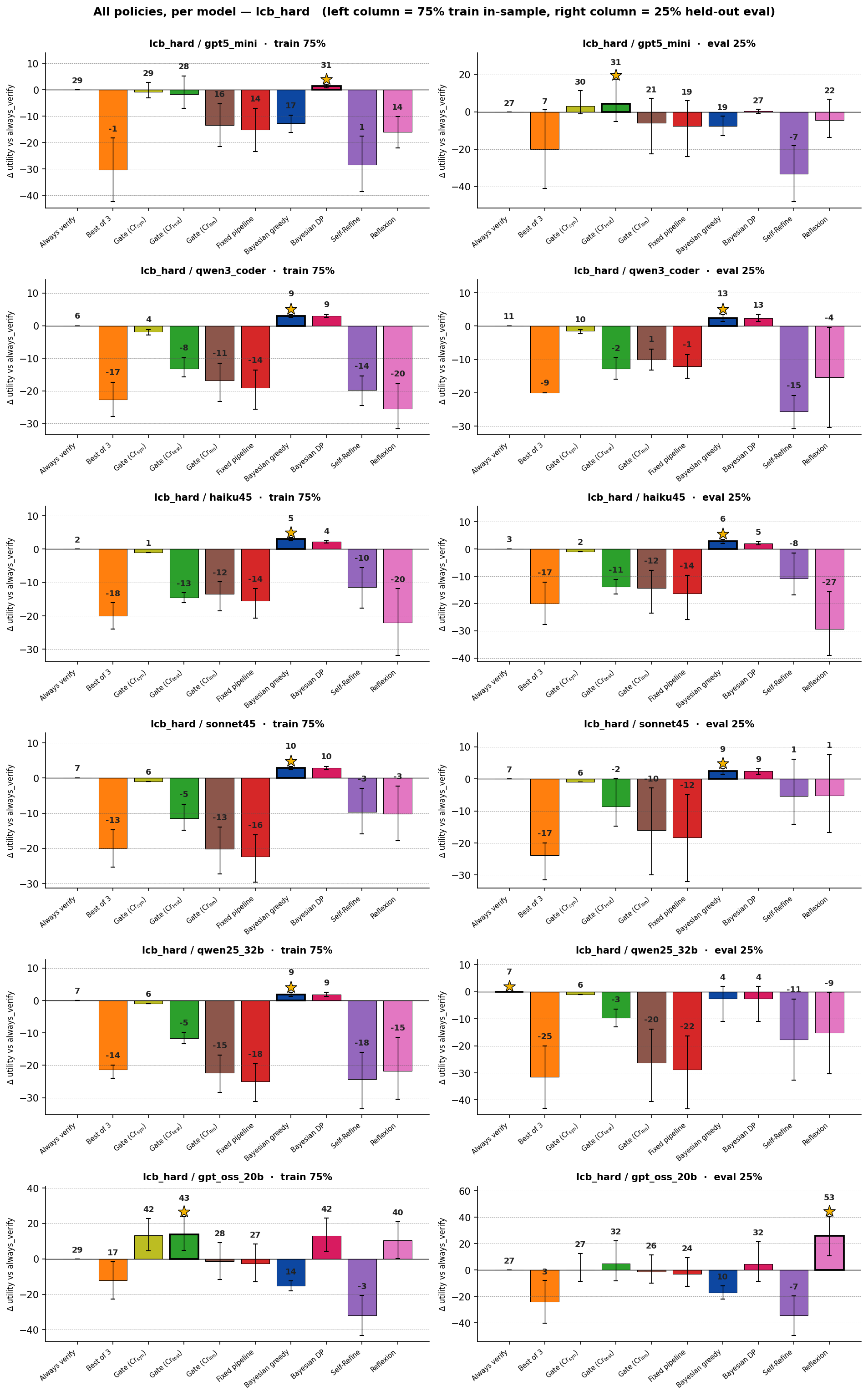}
  \caption{Train-vs-eval policy comparison on LCB-hard. Left
  column: 75\% train (in-sample fit). Right column: 25\% held-out
  eval. One row per generator.}
  \label{fig:tve_lcb_hard}
\end{figure*}

\begin{figure*}[!htb]
  \centering
  \includegraphics[width=0.85\linewidth]{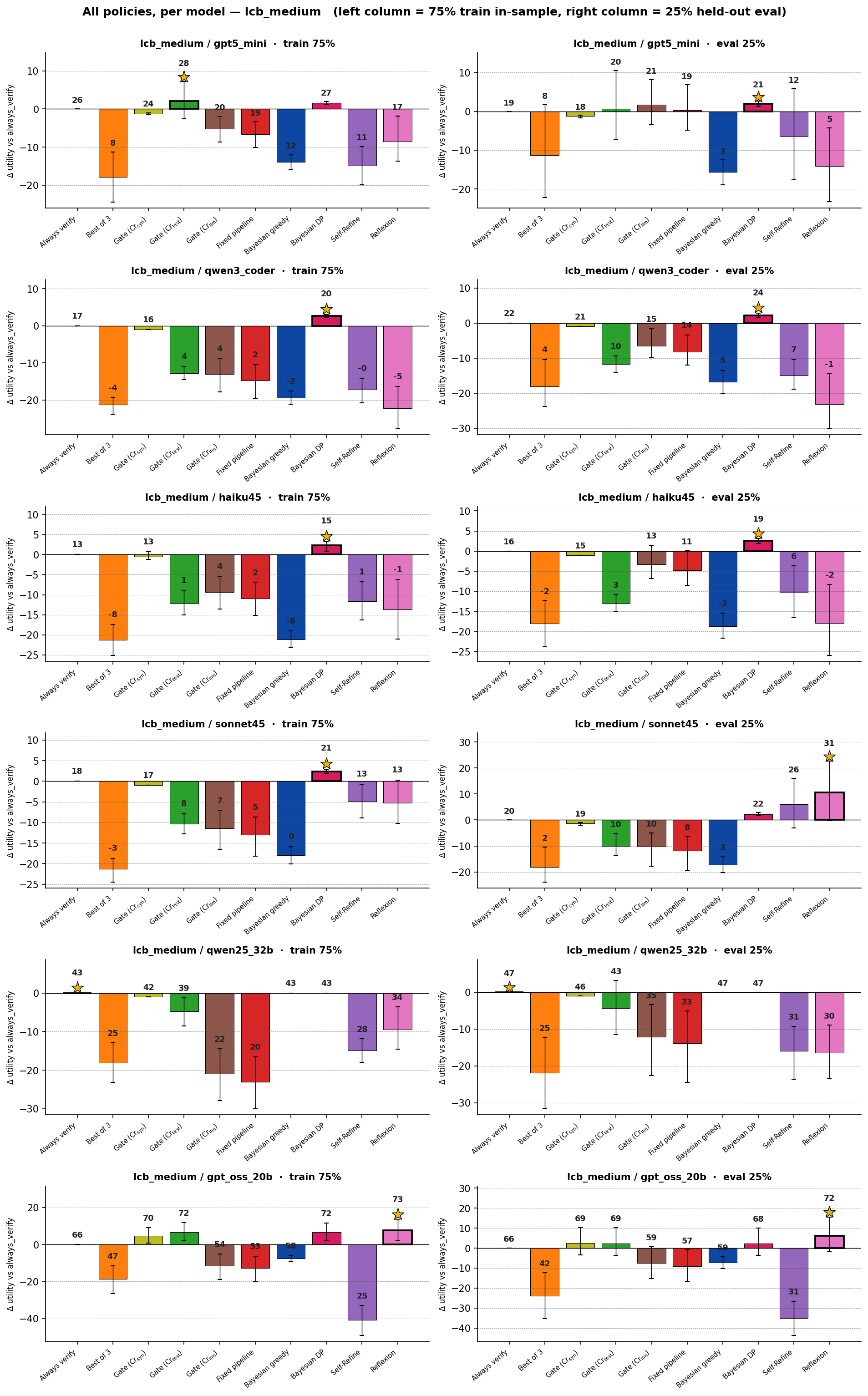}
  \caption{Train-vs-eval policy comparison on LCB-medium.}
  \label{fig:tve_lcb_medium}
\end{figure*}

\begin{figure*}[!htb]
  \centering
  \includegraphics[width=0.85\linewidth]{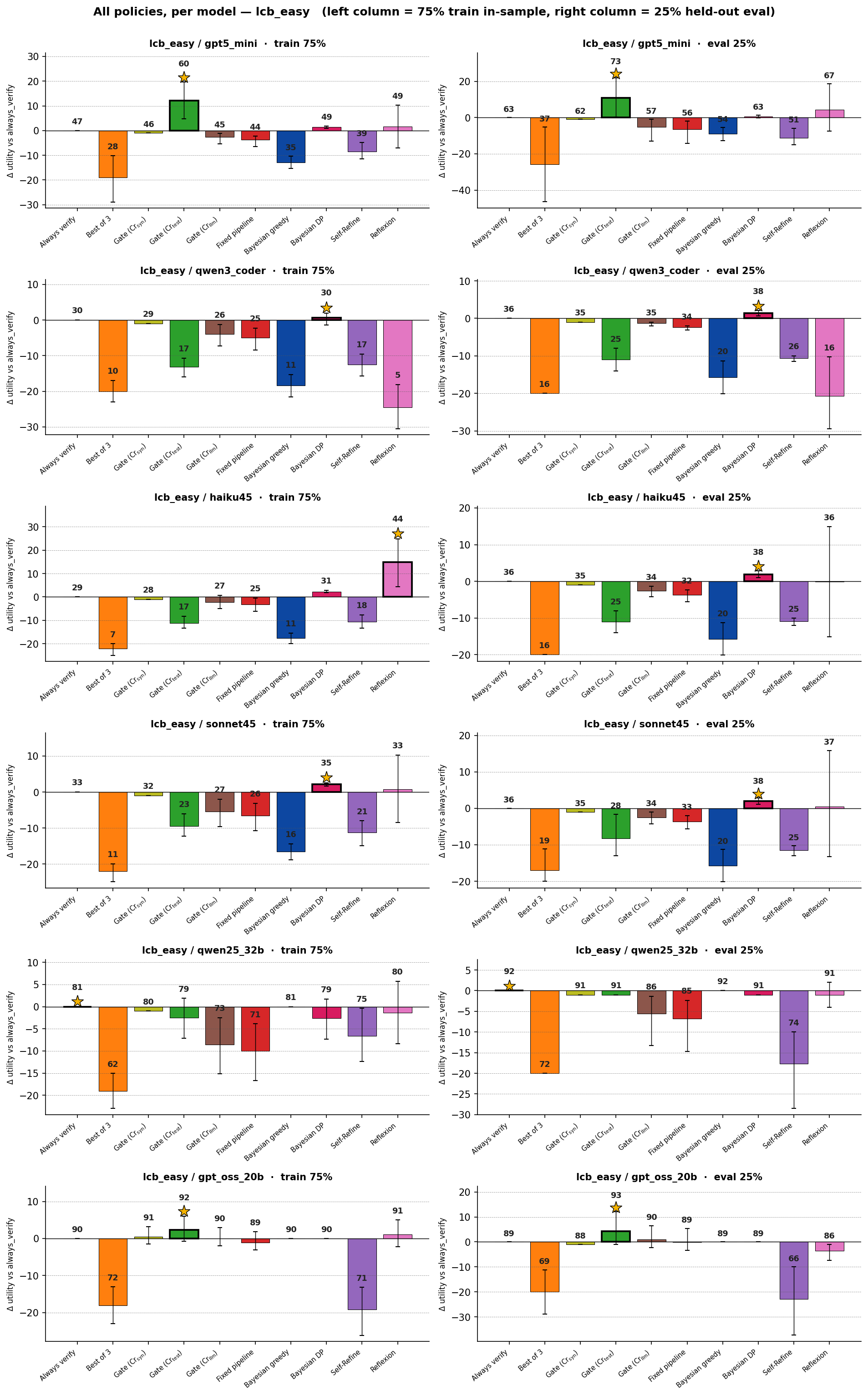}
  \caption{Train-vs-eval policy comparison on LCB-easy.}
  \label{fig:tve_lcb_easy}
\end{figure*}

\begin{figure*}[!htb]
  \centering
  \includegraphics[width=0.85\linewidth]{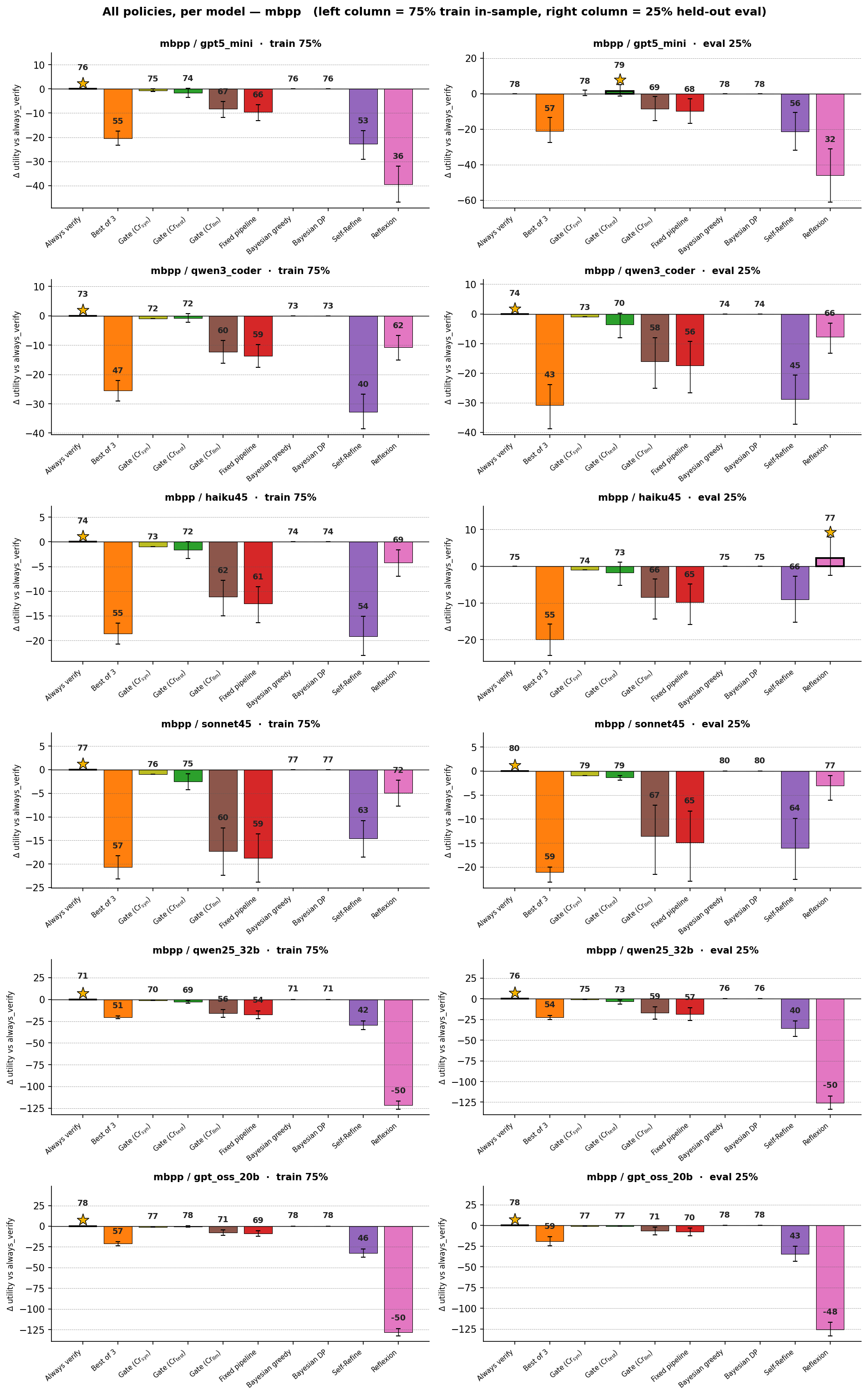}
  \caption{Train-vs-eval policy comparison on MBPP+.}
  \label{fig:tve_mbpp}
\end{figure*}

\begin{figure*}[!htb]
  \centering
  \includegraphics[width=0.85\linewidth]{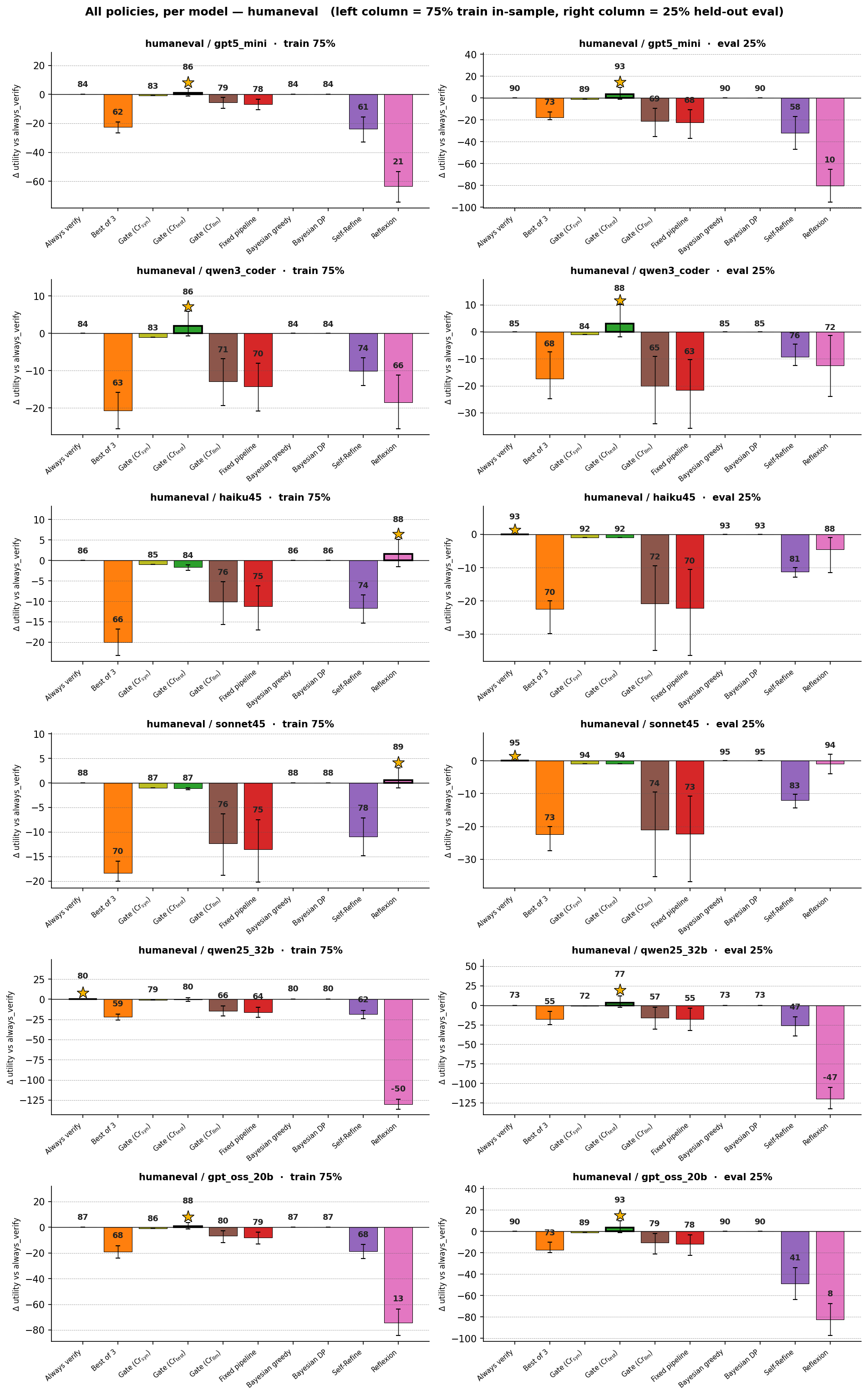}
  \caption{Train-vs-eval policy comparison on HumanEval+.}
  \label{fig:tve_humaneval}
\end{figure*}

\begin{figure*}[!htb]
  \centering
  \includegraphics[width=\linewidth]{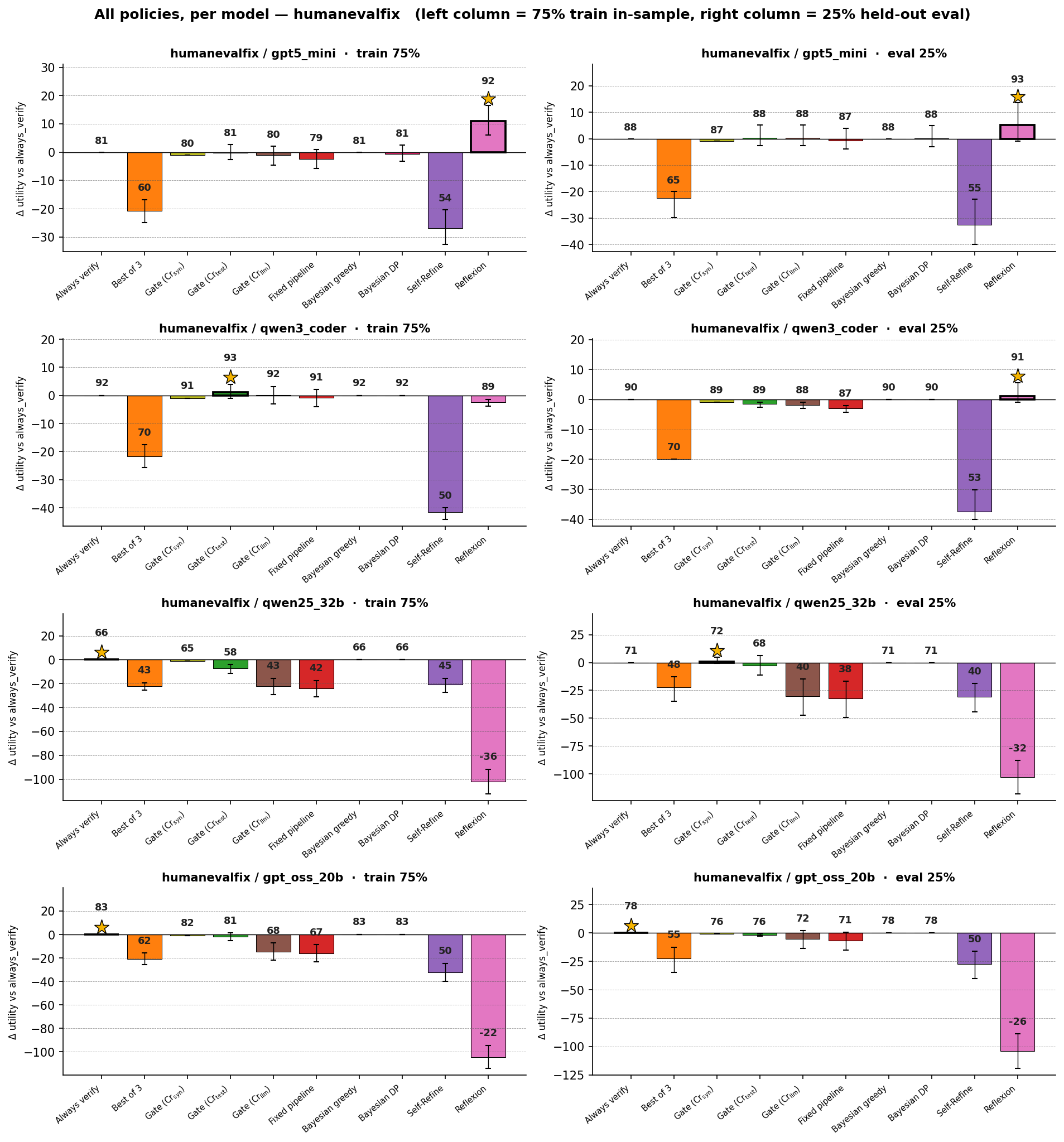}
  \caption{Train-vs-eval policy comparison on HumanEvalFix.}
  \label{fig:tve_humanevalfix}
\end{figure*}

\begin{figure*}[!htb]
  \centering
  \includegraphics[width=\linewidth]{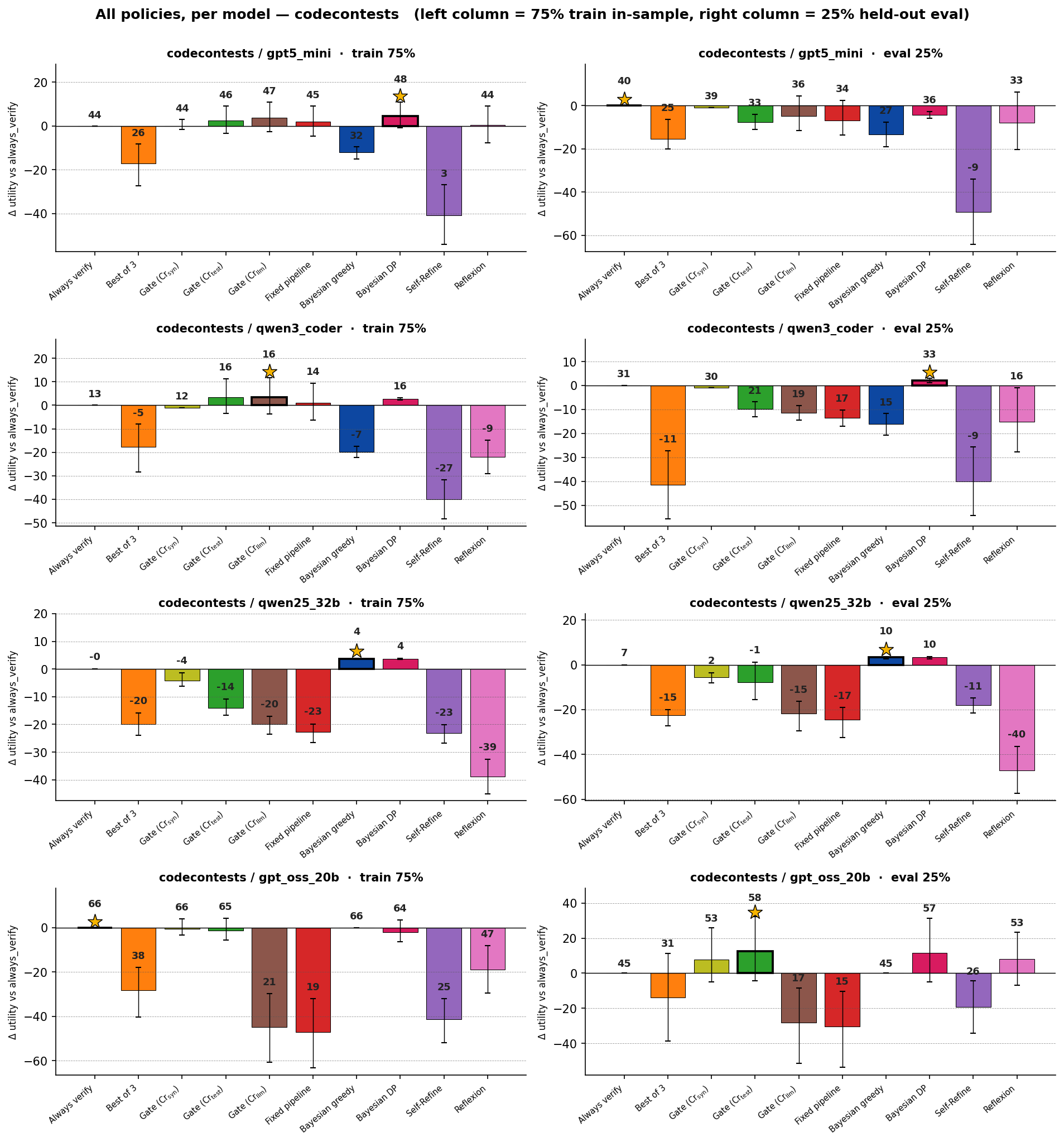}
  \caption{Train-vs-eval policy comparison on CodeContests.}
  \label{fig:tve_codecontests}
\end{figure*}

\begin{figure*}[!htb]
  \centering
  \includegraphics[width=0.85\linewidth]{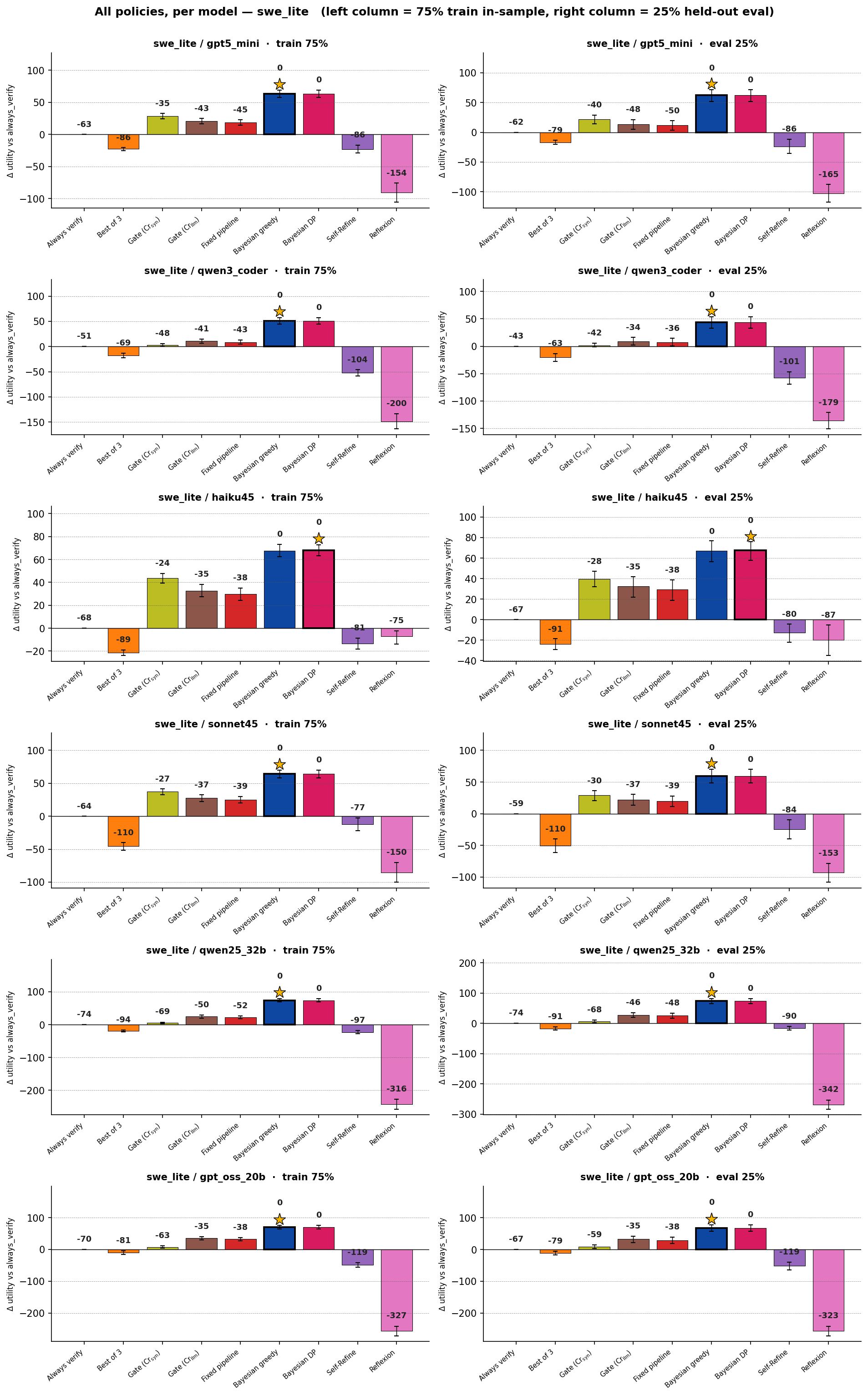}
  \caption{Train-vs-eval policy comparison on SWE-Bench Lite. The
  \texttt{claude-haiku-4.5} row is the headline shown as
  Figure~\ref{fig:policies_eval_main} in the main text.}
  \label{fig:tve_swe_lite}
\end{figure*}

\begin{figure*}[!htb]
  \centering
  \includegraphics[width=0.85\linewidth]{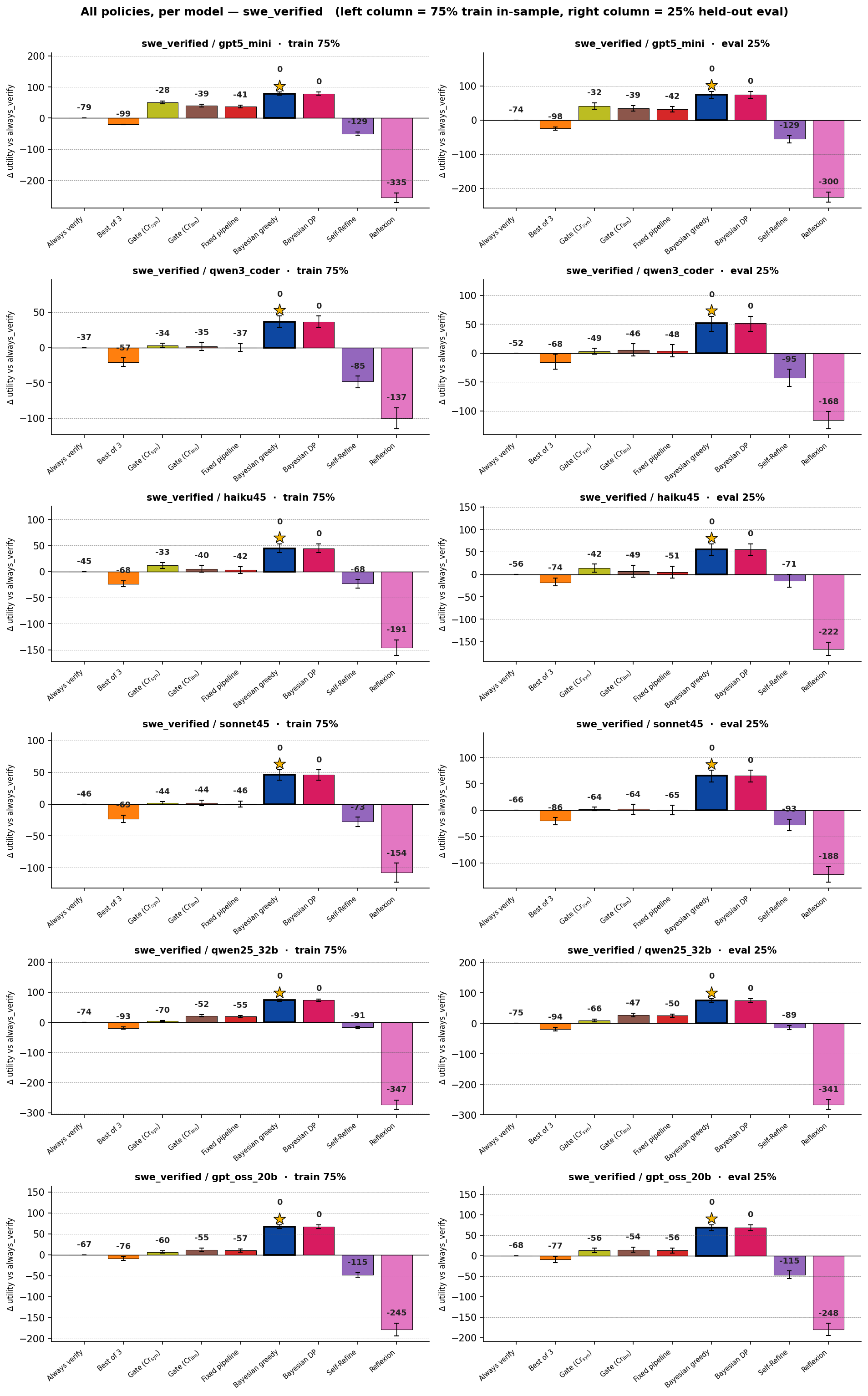}
  \caption{Train-vs-eval policy comparison on SWE-Bench Verified.}
  \label{fig:tve_swe_verified}
\end{figure*}

Figures~\ref{fig:tve_lcb_hard}--\ref{fig:tve_swe_verified} extend the
headline train-vs-eval comparison from
Figure~\ref{fig:policies_eval_main} (SWE-Bench Lite /
\texttt{claude-haiku-4.5}) to every benchmark. Each
figure shows one benchmark across all six generators; the left
column is the 75\% train cohort that the controllers were fit on,
the right column is the held-out 25\% eval cohort. The Bayesian
variants maintain positive $\Delta$ on \emph{both} cohorts in every
regime-A cell, ruling out overfitting of the likelihoods, prior, or
kernel to the calibration instances.

\FloatBarrier
\clearpage

\subsection{Per-Benchmark Verification-Cost Sweeps}
\label{appendix:per_bench_cver}

\begin{figure*}[!htb]
  \centering
  \includegraphics[width=\linewidth]{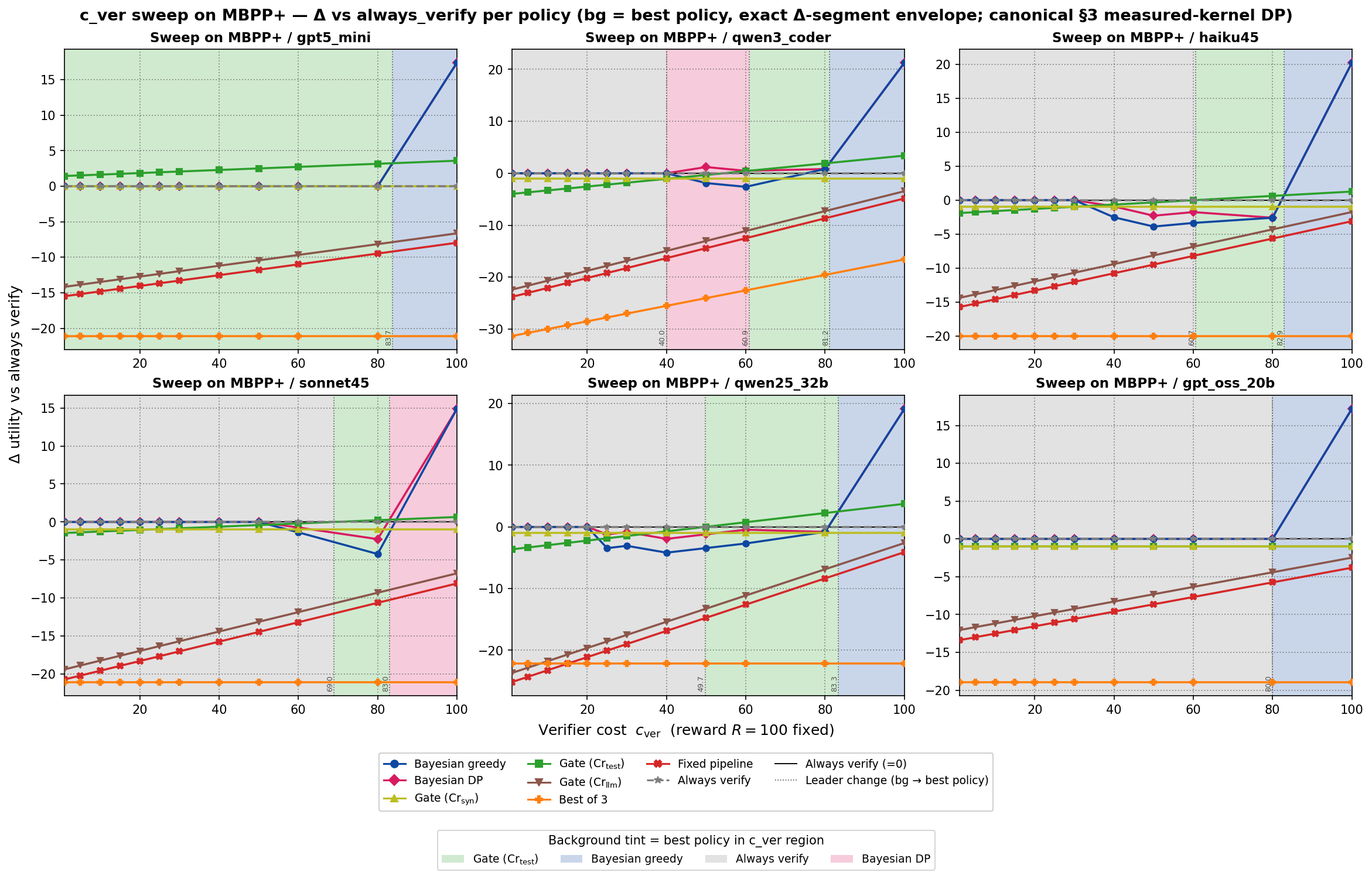}
  \caption{$C_{\mathrm{ver}}$ sweep on MBPP+, one sub-panel per
  generator (companion to Figure~\ref{fig:cver_sweep_main}). The
  C\,$\to$\,B handoff happens at slightly different
  $C_{\mathrm{ver}}$ values per generator, tracking each cell's
  prior $P(Y{=}1)$.}
  \label{fig:cver_sweep_mbpp}
\end{figure*}

\begin{figure*}[!htb]
  \centering
  \includegraphics[width=\linewidth]{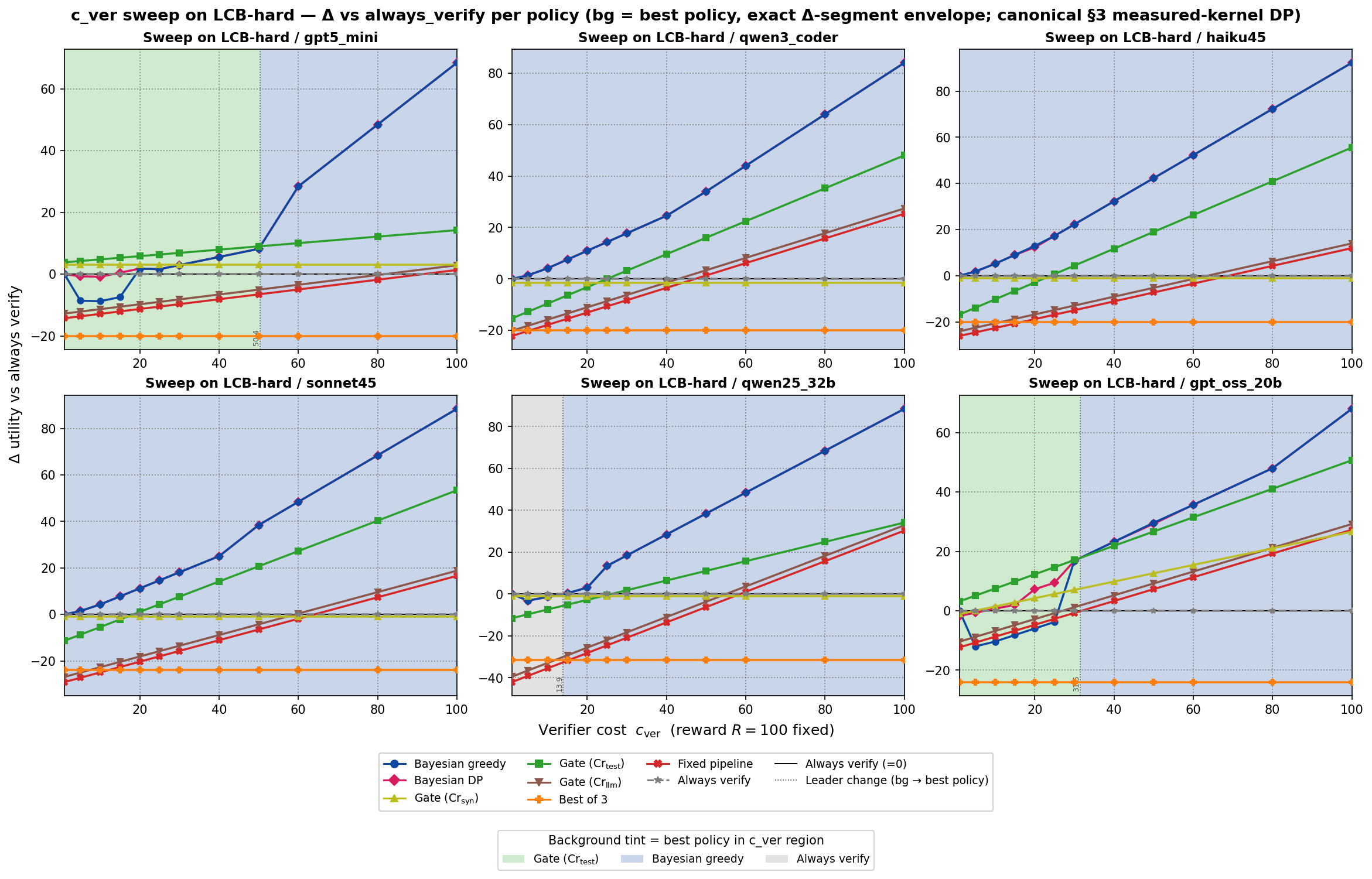}
  \caption{$C_{\mathrm{ver}}$ sweep on LCB-hard, one sub-panel per
  generator. LCB-hard's low prior ($P(Y{=}1)\approx 0.17$) places
  the analytic crossover at $C_{\mathrm{ver}}^* \approx 17$, so the
  cell sits in regime~A for nearly the entire sweep -- the
  Bayesian variants lead above $C_{\mathrm{ver}} \approx 20$. The
  two Bayesian curves overlap because the measured
  $P(\text{fix}\mid\text{broken}) \approx 0.07$ makes multi-step
  refinement negative-EV, so \texttt{bayesian\_DP} collapses to the
  same myopic policy as \texttt{bayesian\_greedy}.}
  \label{fig:cver_sweep_lcb_hard}
\end{figure*}

\begin{figure*}[!htb]
  \centering
  \includegraphics[width=\linewidth]{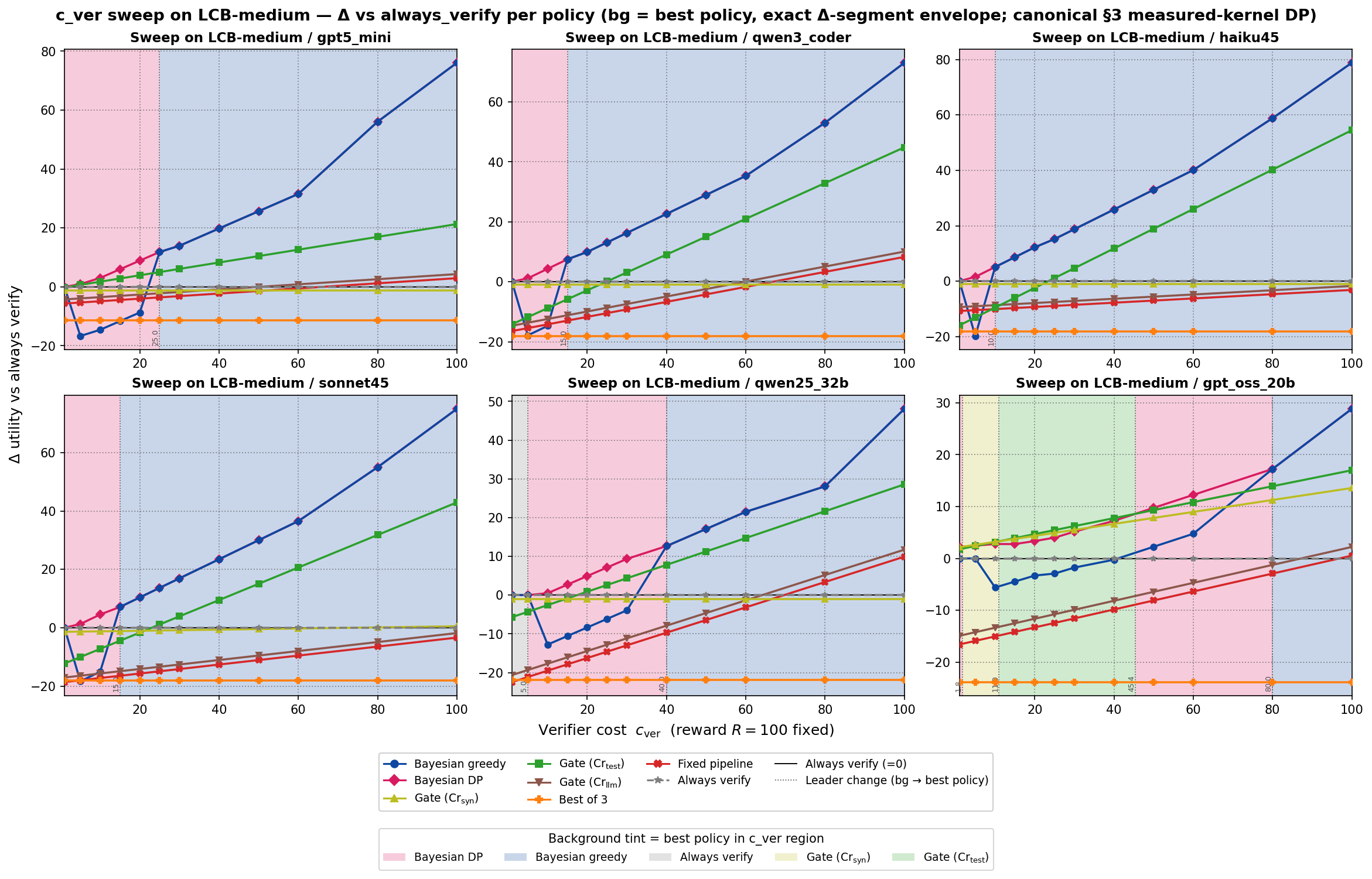}
  \caption{$C_{\mathrm{ver}}$ sweep on LCB-medium, one sub-panel per generator.}
  \label{fig:cver_sweep_lcb_medium}
\end{figure*}

\begin{figure*}[!htb]
  \centering
  \includegraphics[width=\linewidth]{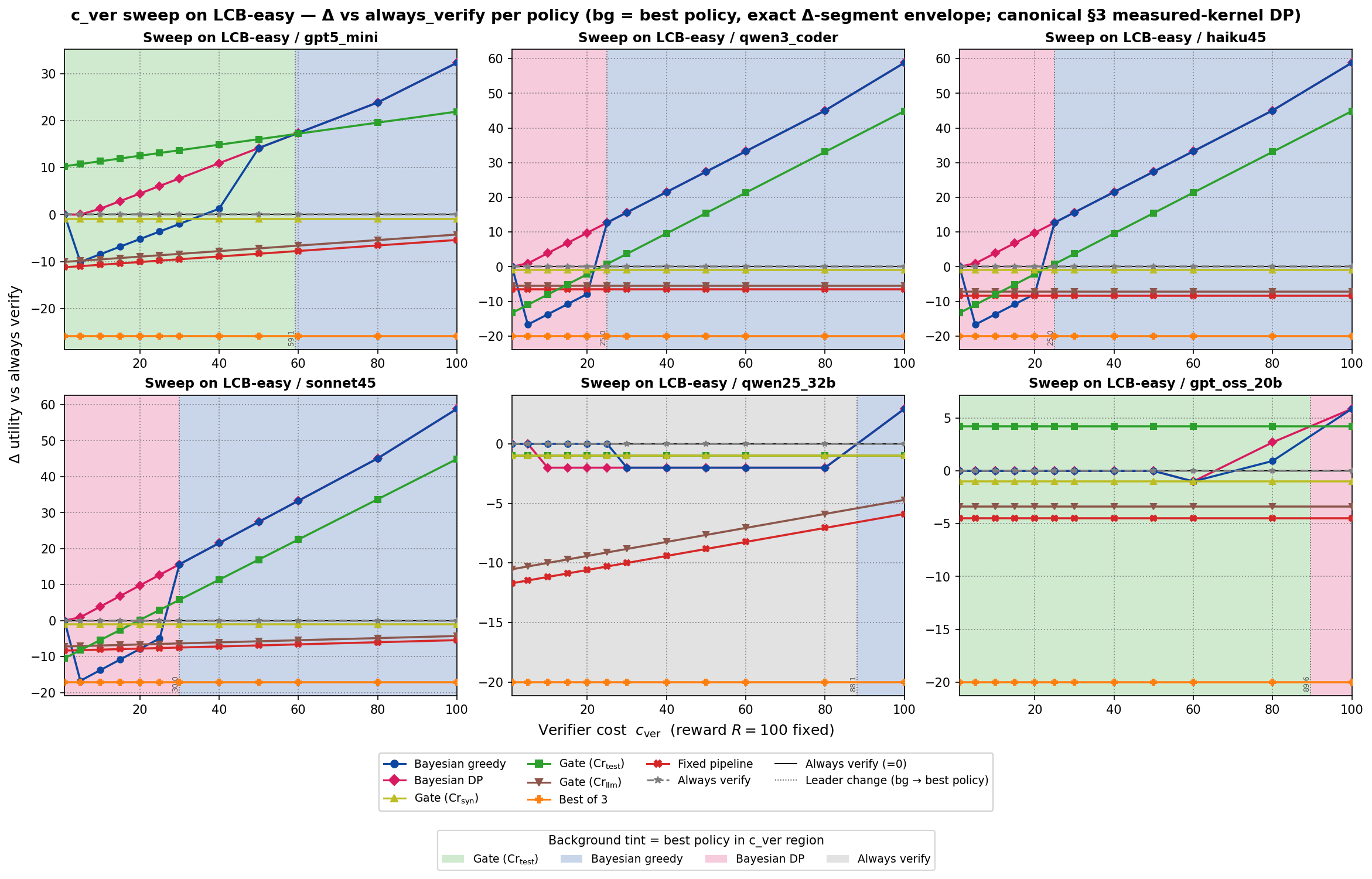}
  \caption{$C_{\mathrm{ver}}$ sweep on LCB-easy, one sub-panel per generator.}
  \label{fig:cver_sweep_lcb_easy}
\end{figure*}

\begin{figure*}[!htb]
  \centering
  \includegraphics[width=\linewidth]{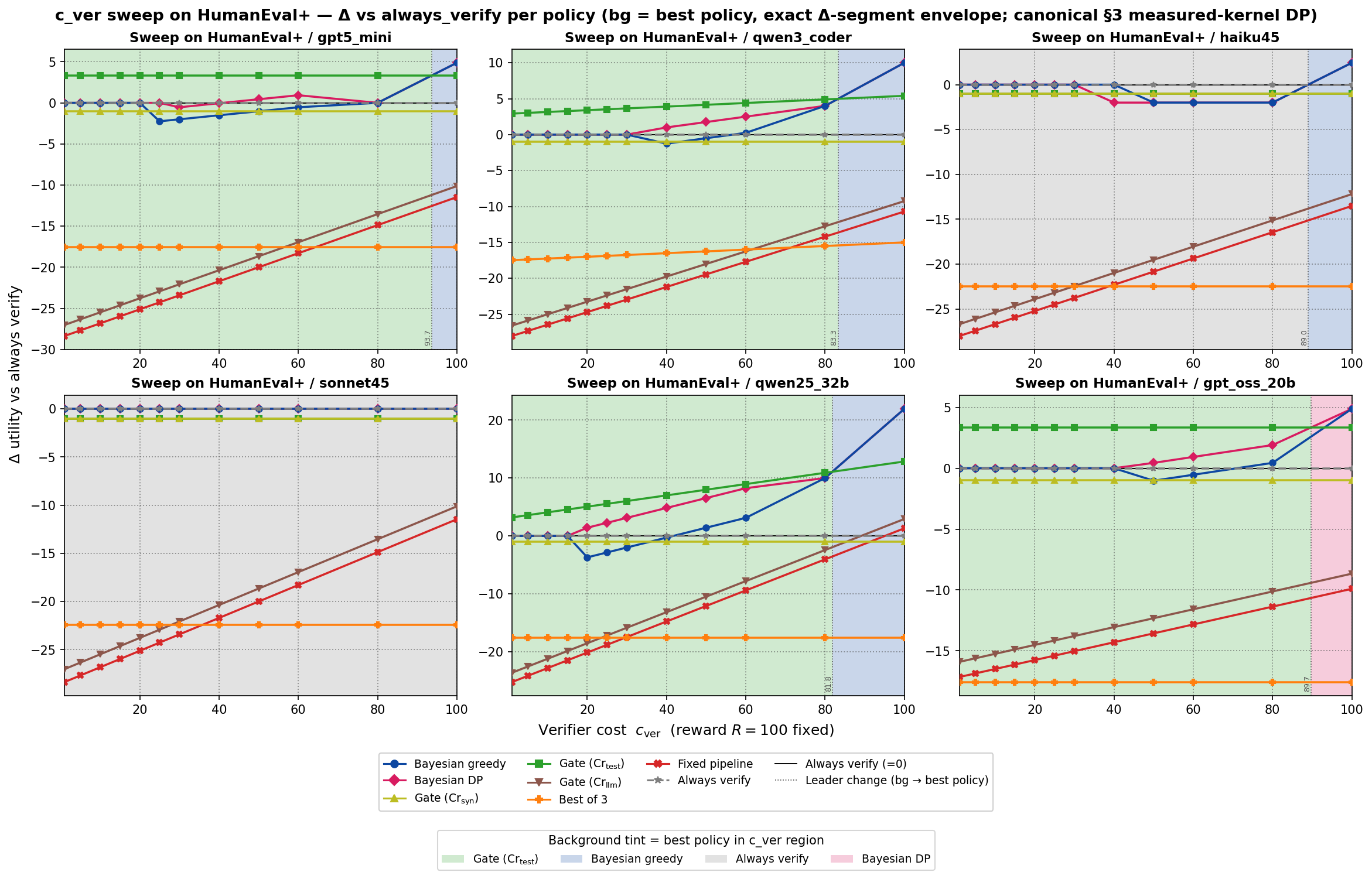}
  \caption{$C_{\mathrm{ver}}$ sweep on HumanEval+, one sub-panel per generator.}
  \label{fig:cver_sweep_humaneval}
\end{figure*}

\begin{figure*}[!htb]
  \centering
  \includegraphics[width=\linewidth]{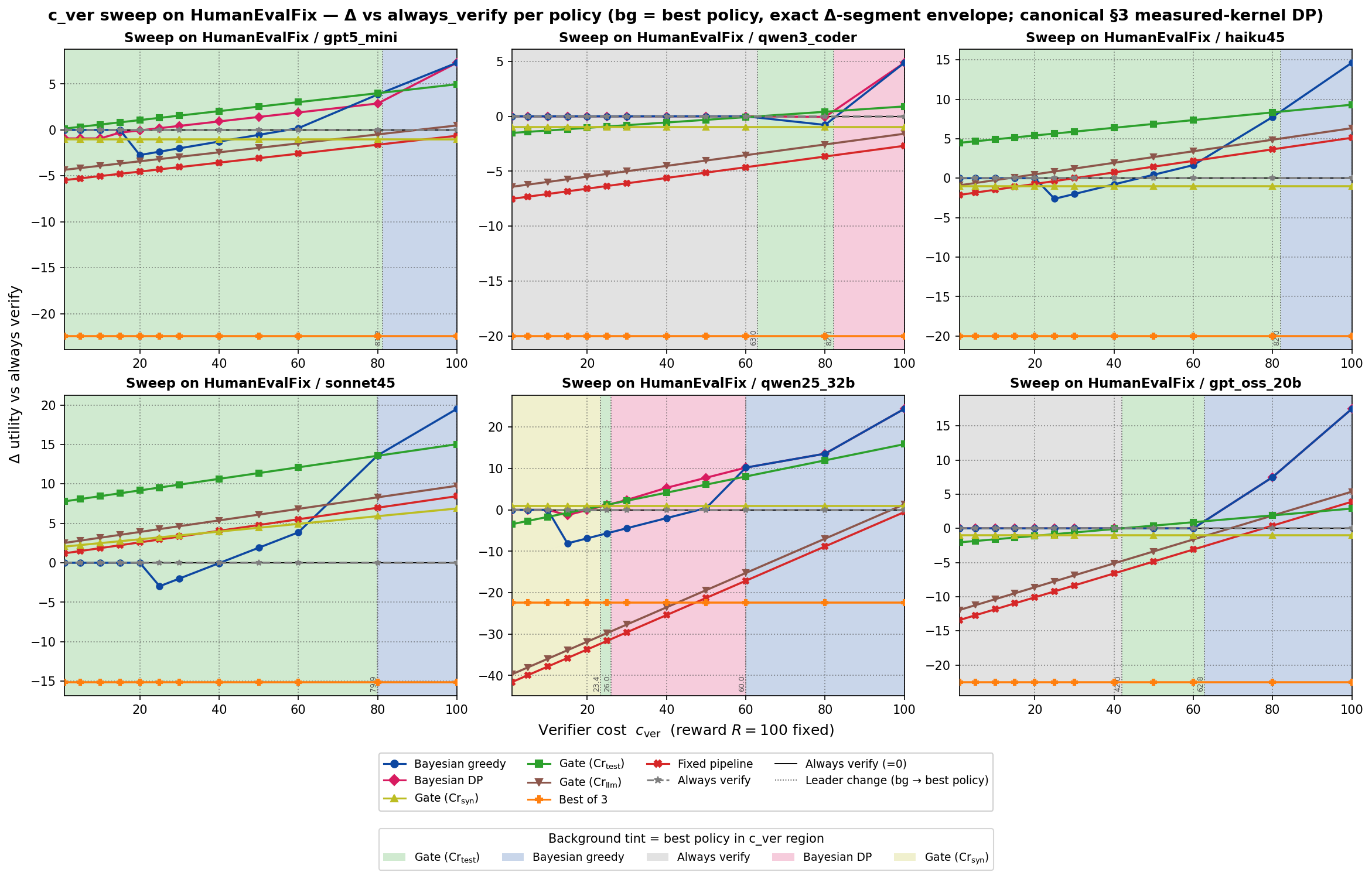}
  \caption{$C_{\mathrm{ver}}$ sweep on HumanEvalFix, one sub-panel
  per generator.}
  \label{fig:cver_sweep_humanevalfix}
\end{figure*}

\begin{figure*}[!htb]
  \centering
  \includegraphics[width=\linewidth]{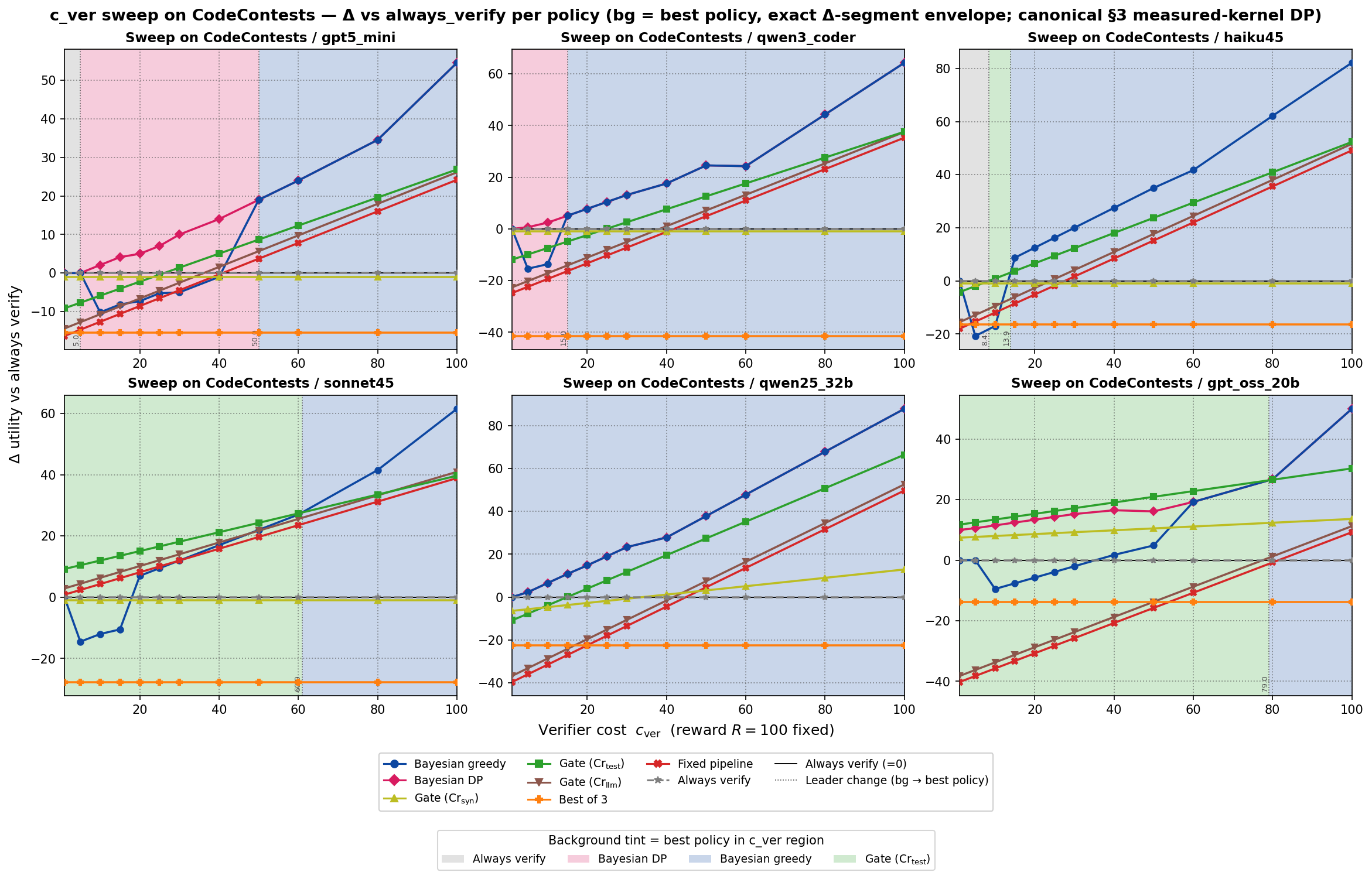}
  \caption{$C_{\mathrm{ver}}$ sweep on CodeContests, one sub-panel
  per generator.}
  \label{fig:cver_sweep_codecontests}
\end{figure*}

\begin{figure*}[!htb]
  \centering
  \includegraphics[width=\linewidth]{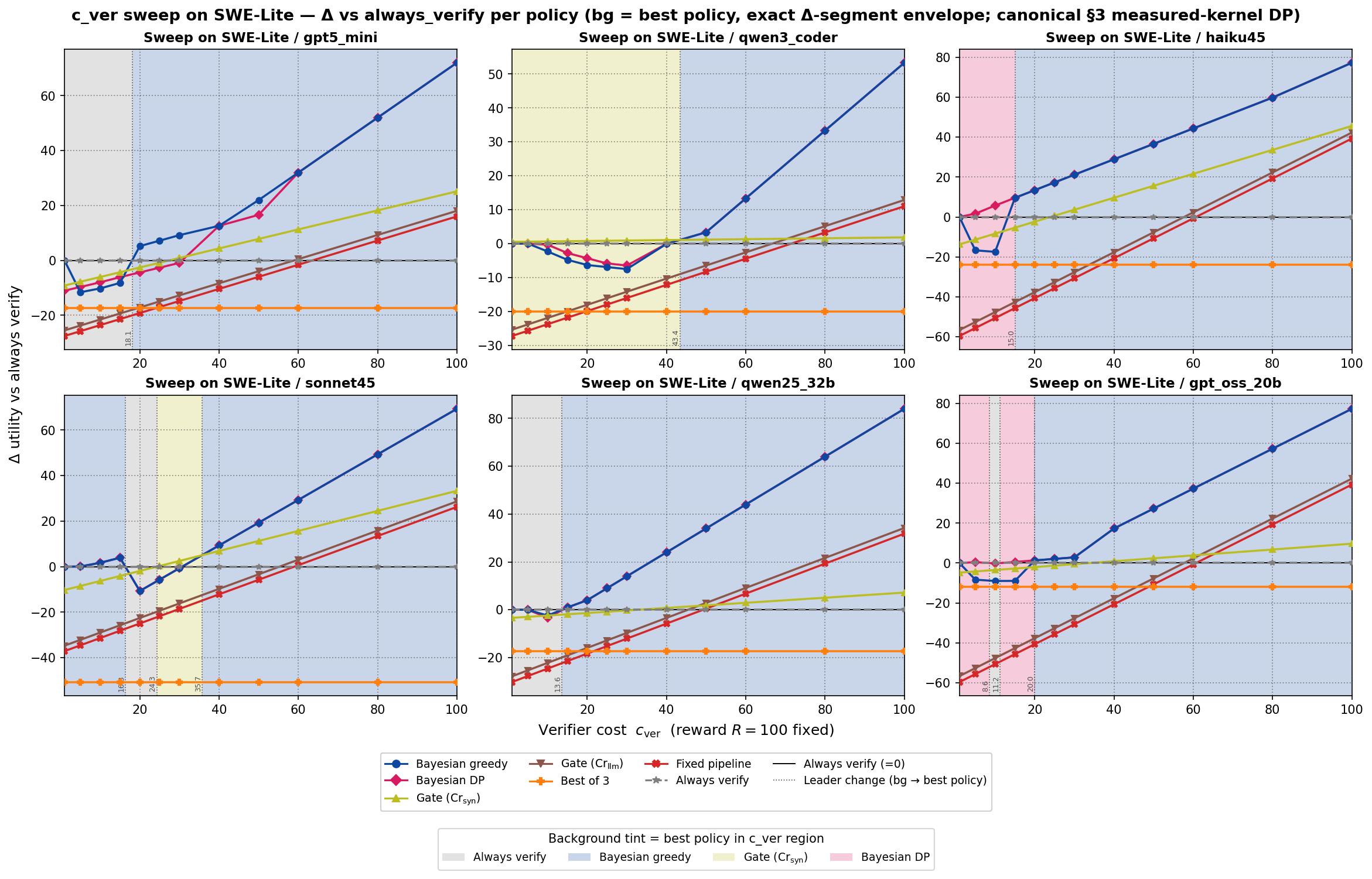}
  \caption{$C_{\mathrm{ver}}$ sweep on SWE-Bench Lite, one sub-panel
  per generator. 
  }
  \label{fig:cver_sweep_swe_lite}
\end{figure*}

\begin{figure*}[!htb]
  \centering
  \includegraphics[width=\linewidth]{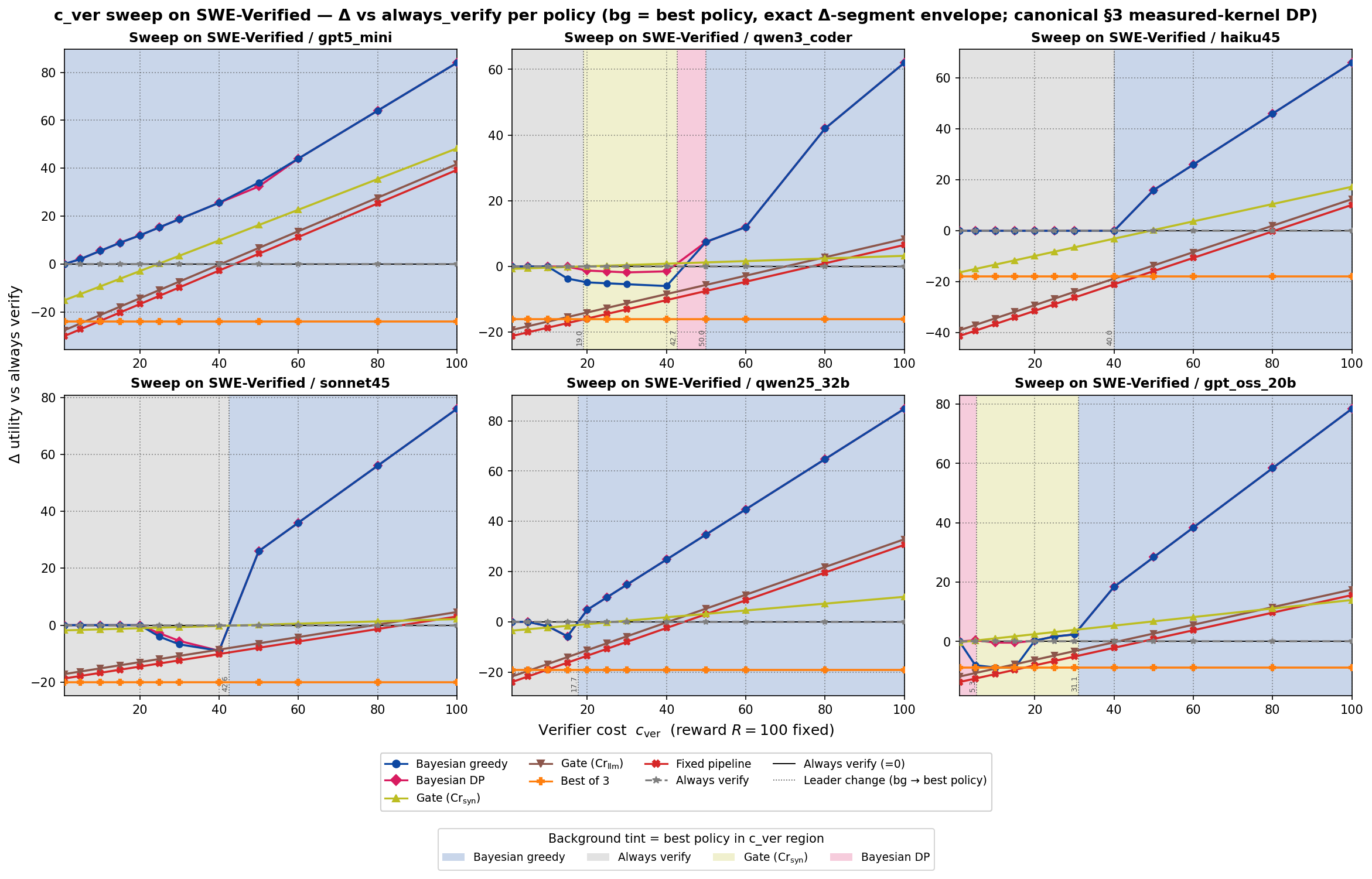}
  \caption{$C_{\mathrm{ver}}$ sweep on SWE-Bench Verified, one
  sub-panel per generator. 
  }
  \label{fig:cver_sweep_swe_verified}
\end{figure*}

Figures~\ref{fig:cver_sweep_mbpp}--\ref{fig:cver_sweep_swe_verified}
extend the headline $C_{\mathrm{ver}}$ sweep from
Figure~\ref{fig:cver_sweep_main} (MBPP+ / \texttt{claude-haiku-4.5})
to every benchmark in the panel. Each figure shows one benchmark with
all six generators on a single composite grid; same axes and policy
palette as the headline. Per-generator analytic crossover
$C_{\mathrm{ver}}^*/R$ tracks each cell's prior $P(Y{=}1)$ and is
visible as the empirical leader-change boundary in each sub-panel.

\FloatBarrier

\end{document}